\documentclass{article}
\newcommand{\R}{\mathbb{R}}
\newcommand{\N}{\mathbb{N}}
\newcommand{\nObservations}{T}
\newcommand{\nVariates}{V}

\newcommand{\nSegments}{M}
\newcommand{\segIndex}{m}
\newcommand{\nVariatePairs}{Q}
\newcommand{\nClusters}{K}
\newcommand{\clusterIndex}{k}
\newcommand{\timeseries}{\mathbf{X}}
\newcommand{\segment}[1][\segIndex]{\mathbf{S}_{#1}}
\newcommand{\segmentLength}{R}
\newcommand{\segmentEndIndex}[1][\segIndex]{\segmentation[#1]}
\newcommand{\segmentToCorrelation}{\mathrm{corr}}
\newcommand{\correlationMatrix}[1][\segIndex]{\mathbf{A}_{#1}}
\newcommand{\correlationMatrixElement}{a}
\newcommand{\correlationMatrixVector}[1][\segIndex]{\mathbf{\correlationMatrixElement}_{#1}^{U}}
\newcommand{\segmentation}[1][]{\eta_{#1}}
\newcommand{\partitionIndex}{i}
\newcommand{\segmentedClustering}[1][\partitionIndex]{\pi_{#1}}
\newcommand{\partition}[1][\partitionIndex]{\mathbf{\pi_{#1}}}
\newcommand{\clustering}[1][\partitionIndex]{\Phi_{#1}}
\newcommand{\segmentedClusteringOrderedPair}[1][\partitionIndex]{(\segmentation[#1], \clustering[#1])}

\newcommand{\clusterIndices}[1][\clusterIndex]{\phi_{#1}}
\newcommand{\cluster}[2][\clusterIndex]{C_{#1}(#2)}
\newcommand{\oldCluster}[1][]{\text{cluster check}}
\newcommand{\subsetOfSymmetricUnitDiagonalMatrices}[1][\nVariates]{\mathbb{A}_{#1}}
\newcommand{\setOfClusterings}[2]{\mathcal{K}(#1,#2)}
\newcommand{\setOfSegmentedClusterings}[2]{\mathcal{SC}(#1,#2)}
\newcommand{\nPatterns}{L}
\newcommand{\nRelaxedPatterns}{L'}
\newcommand{\patternIndex}{\ell}
\newcommand{\canonicalPattern}[1][\patternIndex]{\mathbf{P}_{#1}}
\newcommand{\canonicalPatternElement}{p}
\newcommand{\canonicalPatternVector}[1][\patternIndex]{\mathbf{\canonicalPatternElement}_{#1}^{U}}
\newcommand{\relaxedCanonicalPatternVector}[1][\patternIndex]{{\mathbf{\canonicalPatternElement}'}_{#1}^{U}}
\newcommand{\relaxedPattern}[1][\patternIndex]{\mathbf{P}'_{#1}}
\newcommand{\toleranceBands}{\mathcal{B}}
\newcommand{\patternMap}{\psi}
\newcommand{\argmin}{\arg\!\min}
\newcommand{\levelSetDistances}{\mathcal{D}_{L1}}
\newcommand{\levelSetIndex}{\delta}
\newcommand{\levelSet}[1][\levelSetIndex]{\mathfrak{L}_{#1}}
\newcommand{\avgLevelSetDistance}[1][\levelSetIndex]{\overline{\distanceValue^{\levelSet[#1]}}}
\newcommand{\rateOfIncreaseLevelSet}[2]{r^{\levelSet[#1,#2]}}
\newcommand{\avgRateOfIncreaseLevelSet}[2]{\overline{\rateOfIncreaseLevelSet{#1}{#2}}}
\newcommand{\collectionOfCorrelationMatrices}{\{\correlationMatrix^x\}_{\segIndex \in [\nSegments],\,x \in [\nRelaxedPatterns]}}

\newcommand{\distanceMeasure}[1][]{d_{#1}(\correlationMatrix, \relaxedPattern)}
\newcommand{\dLp}[1][p]{d_{L_{#1}}}
\newcommand{\dRefp}[1][p]{d_{L_{\text{ref}_{#1}}}}
\newcommand{\dDotp}[1][p]{d_{L_{\text{dot}_{#1}}}}
\newcommand{\dFoer}{d_{F}}
\newcommand{\dlogF}{d_{\log F}}
\newcommand{\distanceFunctionCorrxPatterny}[3][]{d_{#1}(\correlationMatrix^{#2}, \relaxedPattern[#3])}
\newcommand{\numberOfSelectedDistanceMeasures}{15}
\newcommand{\distanceValue}{\gamma}
\newcommand{\distancesMultiSet}{\mathfrak{D}}
\newcommand{\normedDistancesMultiSet}{\widetilde{\distancesMultiSet}}
\newcommand{\distancesMultiSetForLevelSet}[1][\levelSetIndex]{\normedDistancesMultiSet_{#1}}
\newcommand{\entropy}[1][\normedDistancesMultiSet]{H_{#1}}
\newcommand{\avgLevelSetEntropy}{\overline{H_{\levelSet[]}}}

\newcommand{\subjects}{N_D}
\newcommand{\nBadPartitions}{66}
\newcommand{\badPartitions}{N_{\pi}}
\newcommand{\silhouetteIndex}[1][\segIndex]{sil_{#1}}
\newcommand{\intraClusterDistance}[1][\segIndex]{ad_{#1}}
\newcommand{\interClusterDistance}[1][\segIndex]{bd_{#1}}

\newcommand{\clusterDistance}[1][\corrMatY]{d(\correlationMatrix, #1)}
\newcommand{\averageClusterDistance}[1][\clusterIndex]{\sigma_{#1}}
\newcommand{\clusterCentroid}[1][\clusterIndex]{\correlationMatrix[C_{#1}]}
\newcommand{\setOfSegmentations}[1]{
    \ifthenelse{\isempty{#1}}{
        \mathcal{G}
    }{
        \setOfSegmentations{}(#1)
    }
}

\usepackage{arxiv}
\usepackage[numbers]{natbib}

\usepackage[utf8]{inputenc} 
\usepackage[T1]{fontenc}    
\usepackage{hyperref}       
\usepackage{url}            
\usepackage{booktabs}       
\usepackage{amsfonts}       
\usepackage{nicefrac}       
\usepackage{microtype}      
\usepackage{lipsum}
\usepackage{enumitem}
\usepackage{graphicx}
\usepackage[dvipsnames]{xcolor}         
\usepackage{colortbl} 
\usepackage{placeins} 

\usepackage{amsmath}
\usepackage[]{amsthm}
\usepackage[]{amssymb}
\usepackage{dsfont}
\usepackage{graphicx}
\usepackage{subcaption}
\usepackage{wrapfig}
\usepackage{multirow}
\usepackage{authblk}
\usepackage{makecell}
\usepackage{ifthen}
\usepackage{xifthen}

\hypersetup{
    colorlinks=true,
    linkcolor=RoyalBlue,
    filecolor=RoyalBlue,
    urlcolor=RoyalBlue,
    citecolor=RoyalBlue
}

\title{Establishing Validity for Distance Functions and Internal Clustering Validity Indices in Correlation Space}

\author[1]{Isabella Degen}
\author[2]{Zahraa S Abdallah}
\author[3]{Kate Robson Brown}
\author[4]{Henry W J Reeve}
\affil[1]{School of Computer Science, University of Bristol}
\affil[2]{School of Engineering Mathematics and Technology, University of Bristol}
\affil[3]{College of Engineering and Architecture, University College Dublin}
\affil[4]{School of Artificial Intelligence, University of Nanjing}

\begin{document}
\maketitle
\begin{abstract} 
Internal clustering validity indices (ICVIs) assess clustering quality without ground truth labels. Comparative studies consistently find that no single ICVI outperforms others across datasets, leaving practitioners without principled ICVI selection. We argue that inconsistent ICVI performance arises because studies evaluate them based on matching human labels rather than measuring the quality of the discovered structure in the data, using datasets without formally quantifying the structure type and quality. Structure type refers to the mathematical organisation in data that clustering aims to discover. Validity theory requires a theoretical definition of clustering quality, which depends on structure type. We demonstrate this through the first validity assessment of clustering quality measures for correlation patterns, a structure type that arises from clustering time series by correlation relationships. We formalise 23 canonical correlation patterns as the theoretical optimal clustering and use synthetic data modelling this structure with controlled perturbations to evaluate validity across content, criterion, construct, and external validity. Our findings show that Silhouette Width Criterion (SWC) and Davies-Bouldin Index (DBI) are valid for correlation patterns, whilst Calinski-Harabasz (VRC) and Pakhira-Bandyopadhyay-Maulik (PBM) indices fail. Simple Lp norm distances achieve validity, whilst correlation-specific functions fail structural, criterion, and external validity. These results differ from previous studies where VRC and PBM performed well, demonstrating that validity depends on structure type. Our structure-type-specific validation method provides both practical guidance (quality thresholds SWC>0.9, DBI<0.15) and a methodological template for establishing validity for other structure types.
\end{abstract}


\section{Introduction}\label{sec:introduction}
Internal clustering validity indices (ICVIs) are widely used to assess the quality of a clustering result without requiring ground truth labels. Clustering is an established approach in data mining to discover hidden patterns in complex data that could lead to new discoveries \cite{Ezugwu2022}. Validating the quality of clustering outcomes is critical in high-stakes domains where spurious patterns could lead to incorrect scientific conclusions \cite{Xu2015, Marti2021, Gao2023, Jaeger2023}. ICVIs quantify the cohesion of objects within the clusters and the separation between the clusters to evaluate the structural quality of clustering results \cite{Fahad2014, Ezugwu2022, Xu2015, Gao2023, Omar2019}. In this context, \emph{structure} refers to the mathematical organisation (patterns) of the data that the clustering aims to discover. Comparative studies evaluating ICVIs use diverse datasets that contain various types of structures without quantifying what type of structure each dataset represents \cite{Milligan1985, Vendramin2010, Arbelaitz2013, Liu2013}. We define \emph{structure types} as different underlying mathematical formulations required to describe the organisation (e.g., patterns in point clouds, density, correlation, or topological structures). Whilst the field recognises that no single ICVI consistently outperforms others across different datasets, there is no principled way to select an ICVI \cite{ Jaeger2023, Todeschini2024}. 

The inconsistent performance of ICVIs stems from a fundamental mismatch between the mechanism of clustering and how ICVIs are evaluated. Clustering algorithms discover mathematical organisation in data, yet comparative studies evaluate ICVIs based on whether they match human-assigned labels or external indices, hence treating clustering like classification where label retrieval is the goal \cite{Milligan1985, Vendramin2010, Arbelaitz2013, Liu2013}. Without formalising what mathematical organisation constitutes optimal clustering for a given structure type, it becomes impossible to distinguish whether ICVIs degrade appropriately when structure is weak or whether they simply cannot measure that type of structure. The lack of quantifying the organisational quality in the datasets provides an explanation of why the same ICVI performs well on some datasets but poorly on others. These datasets contain different structure types at differing organisational quality.

We argue that ICVI validity depends on the structure type the clustering aims to discover. ICVIs were designed to quantify cohesion and separation by employing distance functions to assess dissimilarity, and meaningful dissimilarity depends on structure type. Consequently, an ICVI that is valid for one structure type may lack validity for others. Without clear guidance on which ICVIs are valid for which types of structures, selecting them to validate the quality of a clustering result remains arbitrary and leads to uncertainty about the efficacy of ICVIs \cite{Gagolewski2021, Yerbury2024}. 

We demonstrate the practical importance of structure-type-specific validation by establishing validity through empirical experiments for \emph{correlation patterns}, a structure type discovered through correlation-based clustering that has not been considered in previous comparative ICVI studies \cite{Milligan1985, Vendramin2010, Arbelaitz2013, Liu2013}. Correlation patterns describe different relationship regimes in sequential or time series data that can be used to identify market conditions in finance \cite{Marti2021}, detect operational states in industrial systems \cite{Iglesias2013}, or understand gene expression in biology \cite{Chandereng2020}. Our findings show that the valid distance functions and ICVIs for correlation patterns differ from those established in previous studies.

\begin{figure*}[t]
    \centerline{\includegraphics[width=0.9\textwidth]{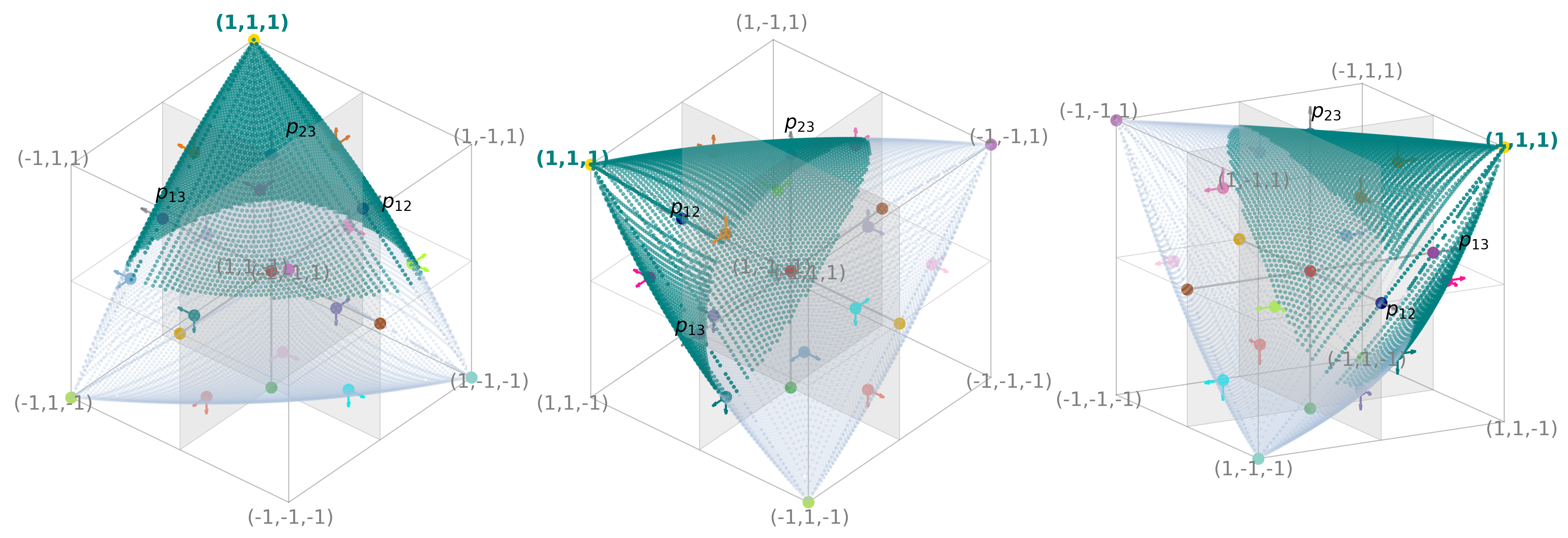}}
    \caption{Correlation elliptope for three variables from different perspectives. The 23 coloured points show the maximally distinct canonical correlation patterns. Arrows indicate patterns lying on coordinate planes through the origin.}
    \label{fig:correlation-elliptope}
\end{figure*}

Establishing that an ICVI is a valid measure of clustering quality involves a nomological network that theoretically describes these constructs for a specific structure type \cite{Salaudeen2025}. For correlation patterns, we mathematically formalise this structure type by defining canonical correlation patterns as geometrically maximally distinct correlation structures within the bounded space of valid correlation coefficients; see Figure\ref{fig:correlation-elliptope}. For three variables, this yields 23 valid canonical correlation patterns occupying the vertices and centre of the correlation elliptope. A valid ICVI identifies clusterings that separate canonical patterns as the highest clustering quality. We empirically validate this using synthetic data modelling canonical patterns with controlled perturbations and systematically degraded clustering quality.

Our approach  provides the methodological foundation for establishing ICVI validity by structure type, demonstrated here for correlation patterns, enabling principled ICVI selection rather than trial and error. Our contributions are:
\begin{enumerate}
    \item Formalisation of correlation patterns as a structure type for clustering. We define maximally distinct canonical correlation patterns that represent optimal clustering in the correlation space and enable the validity assessment of the cluster quality measure.
    \item First structure type specific validity evaluation of distance functions and ICVIs. Using psychometric validity theory, we empirically assess multiple validity aspects of 15 distance functions to quantify dissimilarity and four ICVIs to quantify clustering quality.
    \item Demonstration that the validity of distance functions to measure dissimilarity and ICVIs to measure clustering quality depends on the structure type. Our findings differ from previous comparative ICVI studies, showing that validity cannot be assumed across structure types. We provide practical guidance with quality thresholds for correlation-based clustering applications.
\end{enumerate}

The remainder of this paper is organised as follows. Section 2 reviews related work. Section 3 formalises canonical correlation patterns. Sections 4-5 present technical background and methodology. Sections 6-7 detail experimental setup and results. Section 8 discusses the findings, practical recommendations, and limitations. Section 9 concludes.

\section{Related Work}\label{sec:relatedwork}
ICVIs for evaluating clustering quality are numerous and extensively studied \cite{Omar2019, Hassan2024}. Comparative studies follow \citet{Milligan1985}'s seven-step framework for determining which ICVIs best identify correct cluster numbers. \citet{Vendramin2010} and \citet{Arbelaitz2013} extended this framework, evaluating performance beyond determining correct cluster numbers. Additionally, ICVIs were judged superior if correlating strongly with external indices using ground truth, thus requiring ICVIs to differentiate between various clustering quality. These comprehensive evaluations employed hundreds of synthetic and real datasets (e.g., UCI \cite{uci_ml_repo}) with diverse characteristics. Although some studies qualitatively describe these characteristics (e.g., convex point clouds, density-based clusters, complex shapes, or including noise) \cite{Liu2013}, they do not quantify structure types, which are the different underlying mathematical formulations that describe the organisation in the data that the clustering aims to discover. Comparative studies consistently conclude no single ICVI outperforms others across all datasets, aligning with Kleinberg's impossibility theorem, which states that no clustering function satisfies all desirable properties (scale invariance, richness, and consistency) simultaneously \cite{Kleinberg2002}. More recently, researchers have raised doubts about whether ICVIs can successfully optimise clustering outcomes \cite{Gagolewski2021, Yerbury2024}. \citet{Jaeger2023} describe clustering as "a bit of a quagmire", noting practitioners must develop customised approaches per application context. In short, the advice is "try various methods and see", which forms the basis of the debate of whether clustering is more "art than science" \cite{Guyon2009}.

Inconsistent ICVI performance and growing scepticism raise fundamental questions about ICVI validity for measuring clustering quality. Validity theory, established in psychometrics \cite{Cronbach1955, Campbell1959, Messick1994, ValidityStandards2014}, requires evaluating multiple validity aspects to establish whether measurements adequately capture abstract constructs. These aspects include content validity (capturing all construct aspects?), criterion validity (correlating with external standards?), construct validity (measuring theoretical construct?), and external validity (generalising across conditions?), each requiring separate evaluation \cite{Salaudeen2025}. For abstract constructs like clustering quality, validity assessment requires a nomological network that explicitly maps theoretical relationships to observable indicators \cite{Campbell1959, Salaudeen2025}.

Comparative ICVI studies primarily establish correct cluster number identification matching human judgement \cite{Milligan1985, Liu2013, Vendramin2010, Arbelaitz2013}. Although some studies examine ICVI correlation with external indices (additionally establishing criterion validity) \cite{Vendramin2010, Arbelaitz2013}, or robustness to noise and shape \cite{Liu2013}, they provide insufficient evidence on multiple validity aspects required by psychometric standards. Studies do not assess whether ICVIs adequately operationalise clustering quality as theoretical construct, nor whether distance functions are valid. Therefore, comparative studies evaluate ICVIs based matching human labels rather than optimal mathematical structure discovery. This treats clustering like classification, using human judgements or external indices as ground truth without formalising what mathematical organisation constitutes optimal clustering for different structure types. Whilst studies qualitatively describe dataset characteristics, they do not quantify structure types (convex point clouds, density variations, complex shapes) or assess whether ICVI validity depends on structure type. Systematic structure-type-specific assessment of multiple validity aspects remains absent from clustering validation literature.

We address these gaps by introducing multi-aspect validity assessment as validation goal, formalising correlation patterns as a structure type with canonical patterns enabling nomological network definition, and empirically demonstrating structure-type-dependent ICVI validity using a synthetic dataset that operationalises the structures we aim to discover.

\section{Formalising the Correlation Structure Type}\label{sec:formalisation}
\begin{figure*}[t]
    \centerline{\includegraphics[width=0.8\textwidth]{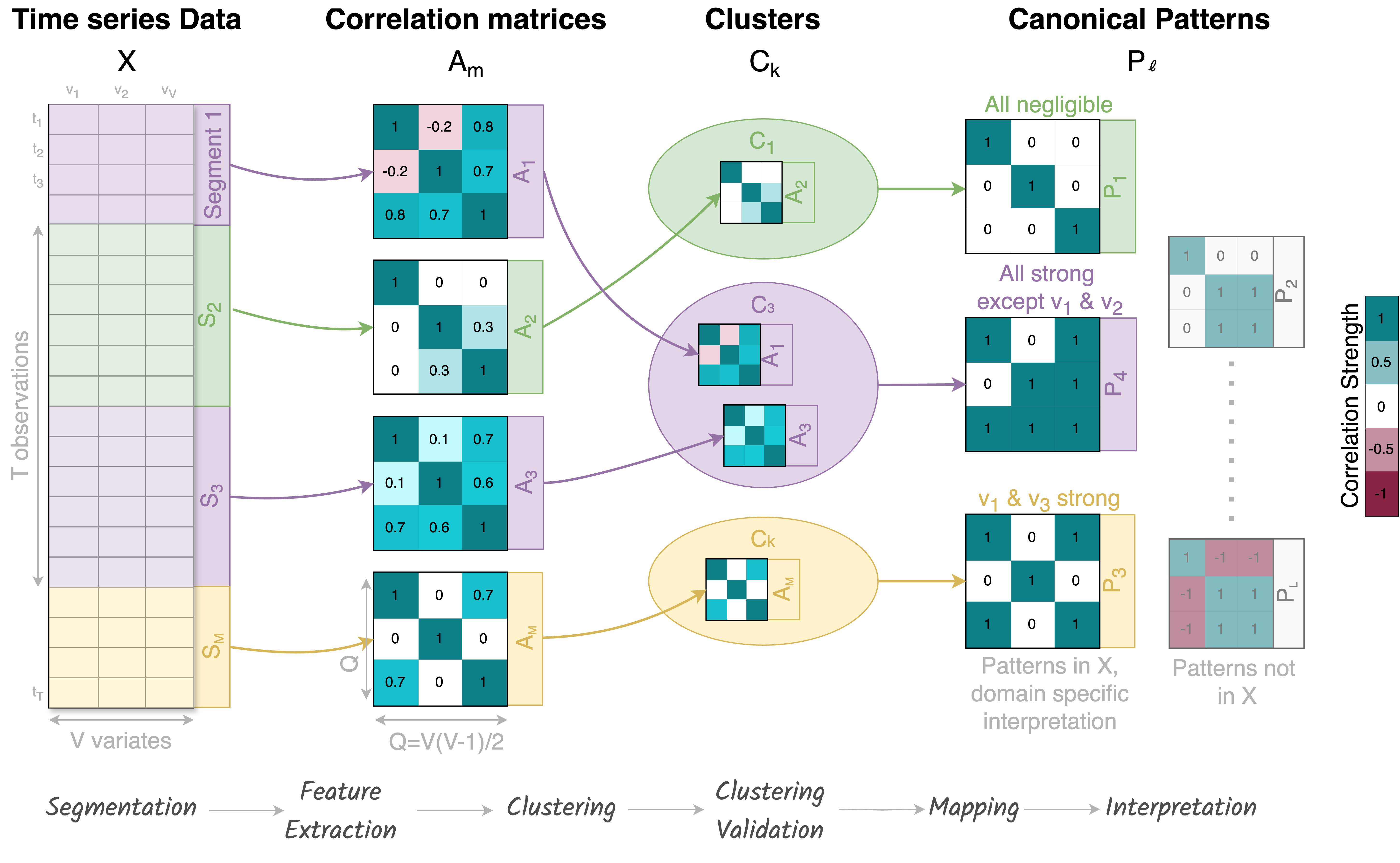}}
    \caption{Multivariate time series $\timeseries$ is segmented when correlation regimes change. Each segment's ($\segment$) correlation matrix  ($\correlationMatrix$) captures relationships between variate pairs ($\nVariatePairs$). Segments with similar patterns are clustered ($C_{\clusterIndex}$), then mapped to the most similar canonical patterns ($\canonicalPattern$) for validation and interpretation.}
  
    \label{fig:clustering-pipeline}
\end{figure*}

This section formalises correlation patterns as a structure type and establishes the mathematical framework that underpins our validity assessment of ICVIs. In the introduction, we define structure type as the underlying mathematical formulation required to describe the organisation in the data that clustering aims to discover. Figure~\ref{fig:clustering-pipeline} illustrates how correlation patterns as structure type are discovered in clustering. Multivariate time series are segmented when variable relationships change (Segmentation), each segment's correlation matrix is computed (Feature Extraction), and segments with similar patterns are clustered (Clustering). We introduce canonical correlation patterns as discrete validation targets representing maximally distinct correlation patterns, enabling both clustering quality validation and domain interpretation.

Fully formalising correlation patterns as a structure type requires defining: (1) how correlation patterns arise from clustering, (2) canonical patterns representing the theoretically optimal patterns, and (3) level sets as the theoretical description of similarity between canonical patterns.

Throughout, we shall let $\R$ denote the set of real numbers and $\N$ the set of positive integers. Given a natural number $N \in \N$, we let $[N]:=\{1,\ldots,N\}$.

\subsection{Correlation Patterns}\label{sec:clusters}
First, we formalise how correlation patterns arise as the objects being clustered. 

Let $\timeseries$ in $\R^{\nObservations \times \nVariates}$ be a matrix where $\nObservations \in \N$ is the number of observations and $\nVariates \in \N$ is the number of variates in the time series.  Given  $\nSegments \in \N$ we shall let $\setOfSegmentations{\nSegments}$ denote the collection of \emph{segmentations} into $\nSegments$ segments. Formally, for each $\nSegments \in \N$ the set $\setOfSegmentations{\nSegments}$ consists of all strictly increasing sequences $\segmentation =(\segmentEndIndex)_{\segIndex=0}^{\nSegments} \in [\nObservations]^{\nSegments+1}$ of length $\nSegments+1$ with $\segmentEndIndex[0]:= 0$ and $\segmentEndIndex[\nSegments]:= \nObservations$. For each $\segIndex\in [\nSegments]$, the values $\segmentEndIndex[\segIndex-1]+1$ and $\segmentEndIndex$ denote the start and end of the $\segIndex$-th index, respectively. We also let $\setOfSegmentations{}:=\bigcup_{\nSegments=1}^{\nObservations}\setOfSegmentations{\nSegments}$ denote the set of all possible segmentations of a time horizon $\nObservations$. For each $\nSegments \in \nObservations$, $\segmentation \in \setOfSegmentations{\nSegments}$ and $\segIndex \in [\nSegments]$ we let 
\begin{equation}
\segment \equiv \segment(\timeseries,\segmentation):=\timeseries(\segmentEndIndex[\segIndex-1]+1,\ldots,\segmentEndIndex[\segIndex]),
\label{eq:segments}
\end{equation}
which takes values in $\R^{|\segmentEndIndex[\segIndex]-\segmentEndIndex[\segIndex-1]|\times \nVariates}$, denote the $\segIndex$-th segment of $\timeseries$, containing sequential observations $t \in \{\segmentEndIndex[\segIndex-1]+1,\ldots, \segmentEndIndex[\segIndex]\}$. 

Then we shall let $\segmentToCorrelation: \bigcup_{\segmentLength \in \N, \, \segmentLength \geq \nVariates} \R^{\segmentLength \times \nVariates} \rightarrow \R^{\nVariates\times \nVariates}$ denote the Spearman rank correlation \cite{Puth2015} mapping from $R \times \nVariates$ matrices $\tilde{X}$ to the corresponding $\nVariates\times \nVariates$ correlation matrix $\correlationMatrix[]$. The tied values are assigned average ranks, and the Spearman correlation is calculated using the standard tie-corrected formula. For our application of $\segmentToCorrelation$, we define the correlation coefficient $\correlationMatrixElement_{ij}:=0$ for the cases of $\segment$ when exactly one of the variates $i$ or $j$ is constant, and $\correlationMatrixElement_{ij}:=1$ when both are constant. The image of $\segmentToCorrelation$ is contained within the subspace $\subsetOfSymmetricUnitDiagonalMatrices \subseteq \R^{\nVariates\times \nVariates}$ consisting of symmetric positive semi-definite matrices with unit diagonal and elements $\correlationMatrixElement_{ij} \in [-1,1]$ for $i,j\in V$ and $i\neq j$. Note that $\subsetOfSymmetricUnitDiagonalMatrices$ forms a $\nVariatePairs$-dimensional subspace where $\nVariatePairs:={\nVariates(\nVariates-1)}/{2}$ is the number of unique pairs of variates. As such, we sometimes represent elements of $\subsetOfSymmetricUnitDiagonalMatrices$ by vectors in $\R^{\nVariatePairs}$, where the correlation coefficients $\correlationMatrixElement_{ij}$ are ordered lexicographically $\correlationMatrixVector[]:=[\correlationMatrixElement_{12}, \correlationMatrixElement_{13}, ..., \correlationMatrixElement_{(\nVariates-1)\nVariates}]$. For each segment $\segment$ where $m\in[M]$, we write 
\begin{equation}
\correlationMatrix \equiv \correlationMatrix(\timeseries,\segmentation) := \segmentToCorrelation(\segment),
\label{eq:correlation-matrix}
\end{equation}
for its correlation matrix.

Next, we write $\setOfClusterings{\nSegments}{\nClusters}$ for the set of all \emph{clusterings} $\clustering[]:=(\clusterIndices[1],\ldots,\clusterIndices[\nClusters])$ which are sequences of $\nClusters$ disjoint non-empty subsets of $[\nSegments]$ whose union is $\bigcup \clustering[]=\clusterIndices[1]\cup\ldots\cup\clusterIndices[\nClusters]=[\nSegments]$.

Additionally, we let $\setOfSegmentedClusterings{\nSegments}{\nClusters}$ denote the set of all \emph{segmented clusterings} which are tuples of the form 
\begin{equation}
    \segmentedClustering[] := \segmentedClusteringOrderedPair[],
    \label{eq:segmentedClustering}
\end{equation}
where $\segmentation[] \in
\setOfSegmentations{\nSegments}$ and $\clustering[] \in \setOfClusterings{\nSegments}{\nClusters}$ is a clustering of these segments.

Furthermore, given $\segmentedClustering[] \in \setOfSegmentedClusterings{\nSegments}{\nClusters}$ and $\segIndex \in [\nSegments]$ and $\clusterIndex \in [\nClusters]$ we let 
\begin{equation}
    \cluster{\segmentedClustering[]}:=\{\correlationMatrix\,:\, \segIndex \in \clusterIndices[\clusterIndex]\},
    \label{eq:a-cluster-ck}
\end{equation}
denote a \emph{cluster} where $\clusterIndices \in \clustering[]$.

Finally, to compare how similar two correlation matrices are, we shall leverage a non-negative function $
d: \subsetOfSymmetricUnitDiagonalMatrices \times \subsetOfSymmetricUnitDiagonalMatrices \to \R_{\geq 0}$ on the space $\subsetOfSymmetricUnitDiagonalMatrices$ as \emph{distance function}. Given correlation matrices $\correlationMatrix[x]$, $\correlationMatrix[y] \in \subsetOfSymmetricUnitDiagonalMatrices$ we view $d(\correlationMatrix[x], \correlationMatrix[y])$ as quantifying the degree of dissimilarity between $\correlationMatrix[x]$ and $\correlationMatrix[y]$. In practice many examples of $d$ will be metrics, yet we require no further properties beyond non-negativity. We provide the definitions for the various distance functions employed in Appendix~\ref{app:distance-functions}.

\subsection{Canonical Patterns}\label{sec:canonical-patterns}
Next, we formalise canonical correlation patterns as the theoretical optimal clustering for this structure type. Canonical correlation patterns represent maximally distinct correlation patterns as the theoretical boundaries of structural differentiation (maximal separation) in the correlation space.

Let $\canonicalPattern[] \in \subsetOfSymmetricUnitDiagonalMatrices$ denote a \emph{canonical correlation pattern} with elements $\canonicalPatternElement_{ij}\in\{-1,0,1\}$ for $i,j \in [\nVariates]$ and $i\neq j$. Each canonical pattern $\canonicalPattern$ with $\patternIndex \in [\nPatterns]$ represents a unique idealised correlation pattern capturing strong positive correlation, no correlation, or strong negative correlation between the $\nVariatePairs$ unique variate pairs. The total number of different canonical patterns is $\nPatterns:=3^\nVariatePairs$.

A canonical pattern $\canonicalPattern$ must be positive semi-definite to be a valid correlation matrix. To fulfil this requirement, we define a relaxed discretisation criterion where $\relaxedPattern$ is considered equivalent to its canonical form $\canonicalPattern$, if $\relaxedPattern$ is positive semi-definite after adjusting its correlation coefficients within tolerance bands $\toleranceBands:=\{[-1,-0.7],[-0.2,0.2],[0.7,1]\}$. Canonical patterns $\canonicalPattern$ that cannot be made positive semi-definite after this relaxation within $\toleranceBands$ are invalid. The tolerance bands are geometrically derived from the correlation elliptope representing all valid correlation coefficients, see Figure~\ref{fig:correlation-elliptope}. For $\nVariates=3$, the positive semi-definite constraints that define the correlation elliptope form unit circles when one correlation coefficient is zero $\canonicalPatternElement_{ij}=0$, for $i \neq j$. Therefore, patterns $\canonicalPattern$ that are positioned in the corners of these planes through the origin of the coordinate system (e.g., $[1,1,0]$) can be at most on the unit circle to remain valid correlation patterns. Hence, the maximal coordinates for these corner patterns are $\cos(45^\circ) = \sin(45^\circ) = \sqrt{2}/2=0.707$. The tolerance bands $\toleranceBands$ represent practical thresholds that continue to clearly distinguish between strong negative, negligible, and strong positive correlation coefficients, while respecting the geometry of the correlation space $\subsetOfSymmetricUnitDiagonalMatrices$. This relaxation process determines which of the $\nPatterns$ theoretical patterns yield valid correlation matrices, resulting in $\nRelaxedPatterns <\nPatterns$ relaxed canonical patterns $\relaxedPattern$ where $\patternIndex \in [\nRelaxedPatterns]$. To facilitate human interpretation, we use the idealised $\canonicalPattern$ to represent a pattern. For calculations, we use the corresponding relaxed form $\relaxedPattern$, see Appendix~\ref{app:canonical-patterns} for a list of the 23 valid canonical patterns out of the 27 possible and their relaxed form for three time series variates. To illustrate this relaxation process, consider a few examples (for $\nVariates=3$): pattern $\canonicalPatternVector[13] = [1,1,1]$ (all strong positive correlations) requires no adjustment and remains $\relaxedCanonicalPatternVector[13] = [1,1,1]$, pattern $\canonicalPatternVector[4] = [0,1,1]$ becomes $\relaxedCanonicalPatternVector[4] = [0,0.71,0.7]$ after relaxation within $\toleranceBands$, and pattern $\canonicalPatternVector[14] = [1,1,-1]$ is invalid since it cannot achieve positive semi-definiteness even after relaxation it continues to be positioned outside the correlation elliptope.

Hence, canonical correlation patterns define what perfect correlation-based clustering looks like. Time series data that perfectly separate into canonical patterns achieve maximum possible clustering quality, as no greater structural contrast can exist within the mathematical constraints of valid correlation matrices.

To associate an empirical correlation matrix $\correlationMatrix$ with its closest relaxed canonical pattern, let $\patternMap: \subsetOfSymmetricUnitDiagonalMatrices \mapsto \{\relaxedPattern : \patternIndex \in [\nRelaxedPatterns]\}$ denote the \emph{canonical pattern mapping} 
\begin{equation}
    \patternMap(\correlationMatrix) := \min\,\underset{\patternIndex \in [\nRelaxedPatterns]}\argmin \left\{ d(\correlationMatrix, \relaxedPattern) \right\}.
    \label{eq:canonical-pattern-mapping}
\end{equation}
Here, `$\min$' ensures that ties are broken by selecting the smallest pattern index $\patternIndex$.

Finally, we write $\correlationMatrix^\patternIndex$, for the empirical correlation matrix of a segment with index $\segIndex \in [\nSegments]$ that was generated from the relaxed canonical pattern $\relaxedPattern$, $\patternIndex \in [\nRelaxedPatterns]$. The superscript $\patternIndex$ indicates the ground truth pattern index for the correlation matrix.

\subsection{Theoretical Pattern Similarity}\label{sec:level-sets}
Having defined how correlation patterns arise and ideal canonical correlation patterns, we now define the theoretically expected similarity construct between canonical patterns. Intuitively, two canonical patterns should be considered more similar if they differ in fewer correlation coefficients. We formalise this intuition by defining level sets that group pairs of canonical patterns by this intuitive similarity based on how many coefficients change by how much.

Let $d_{L1}: \subsetOfSymmetricUnitDiagonalMatrices \times \subsetOfSymmetricUnitDiagonalMatrices \to \R_{\geq 0}$ denote the L1 distance between canonical patterns so that 
\begin{equation}
d_{L1}(\canonicalPattern[x], \canonicalPattern[y]) := \sum_{i=1}^{\nVariates}\sum_{j>i}^{\nVariates}|\canonicalPatternElement_{x_{ij}}-\canonicalPatternElement_{y_{ij}}|
\label{eq:L1-distance},
\end{equation}
for pattern indices $x,y \in [\nRelaxedPatterns]$.

Let $\levelSetDistances := \{\levelSetIndex \in \R : \exists x,y \in [\nRelaxedPatterns],\, d_{L1}(\canonicalPattern[x],\canonicalPattern[y]) = \levelSetIndex \}$ denote the ordered sequence of achievable L1 distances between valid canonical patterns. The theoretical maximum $\levelSetIndex_{\text{max}}$ is bounded by $\nVariates(\nVariates-1)$, although this $\levelSetIndex_{\text{max}}$ may not be achievable due to the constraints on valid correlation matrices. To give an example, with three time series variates ($\nVariates=3$) the sequence becomes $\levelSetDistances = \{0,1,2,3,4,5\}$, where a distance of 1 represents one coefficient changing by $\pm1$, a distance of 2 represents either two coefficients changing by $\pm1$ or one coefficient flipping sign, and so forth. The theoretical maximum distance of $6$ for $\nVariates=3$ cannot be achieved while only retaining the canonical patterns $\canonicalPattern$ that can satisfy the tolerance bands $\toleranceBands$ and the relaxed pattern $\relaxedPattern$ is a valid correlation matrix.

For each potential distance value $\levelSetIndex \in \levelSetDistances$, we define the corresponding \emph{level set} $\levelSet$ as

\begin{equation}
\levelSet := \{(x,y) : x,y \in [\nRelaxedPatterns], d_{L1}(\canonicalPattern[x],\canonicalPattern[y]) = \levelSetIndex\}.
\label{eq:level-set}
\end{equation}

The collection $\{\levelSet\}_{\levelSetIndex \in \levelSetDistances}$ groups pattern index pairs $(x,y)$ by their L1 distance, with $\levelSet[0]$ containing indices of identical patterns, $\levelSet[1]$ containing indices of patterns where one correlation coefficient differs by $\pm1$, and so forth. This creates a natural ordering of dissimilarity between the level sets, while pattern pairs within each $\levelSet$ are theoretically similar.

\subsection{Theoretical Dissimilarity Characterisation}\label{sec:distance-characterisation}
To complete the formalisation, we now provide the definitions for quantities that characterise the degree of dissimilarity measured by a distance function $d$ that are consistent with theoretical similarity established by the level sets. 

Given a correlation matrix  $\correlationMatrix^x$ with a known ground truth canonical pattern with index $x \in [\nRelaxedPatterns]$ and a relaxed canonical pattern $\relaxedPattern[y]$ with $y \in [\nRelaxedPatterns]$, we write $\distanceValue_{m,y}^{x} \in \R_{\geq 0}$ for the computed distance value $\distanceFunctionCorrxPatterny{x}{y}$ using a specific distance function $d$. 

Then given a collection of empirical correlation matrices $\collectionOfCorrelationMatrices$ with known ground truth, $\avgLevelSetDistance \in \R_{\geq 0}$ denotes the \emph{average level set distance} for the level set $\levelSet$ formed from the collection of correlation matrices and canonical patterns in that level set and is defined as

\begin{equation}
\avgLevelSetDistance := \frac{1}{|\levelSet|}\sum_{x,y \in \levelSet, x \neq y} \distanceValue_{m,y}^{x},
\label{eq:avg-level-set-distance}
\end{equation}
for $\levelSetIndex \in \levelSetDistances$.

Next, $\rateOfIncreaseLevelSet{i}{j} \in \R_{\geq 0}$ denotes the \emph{rate of increase} in average level set distance between two level sets $\levelSet[i]$ and $\levelSet[j]$ defined as

\begin{equation}
\rateOfIncreaseLevelSet{i}{j}:= \text{abs}(\avgLevelSetDistance[i]-\avgLevelSetDistance[j])
\label{eq:rate-of-increase-level-set}
\end{equation}
for $i, j \in \levelSetDistances$.

Next, $\distancesMultiSet$ denotes the collection of distance values and is defined as $\distancesMultiSet := \{\distanceValue_{m,y}^{x} : x,y \in [\nRelaxedPatterns], m \in [\nSegments]\}$ and $\normedDistancesMultiSet$ denotes the finite and normalised version of $\distancesMultiSet$ that is defined as 
$$\normedDistancesMultiSet := \left\{\frac{\gamma - \min(\distancesMultiSet \cap \mathbb{R})}{\max(\distancesMultiSet \cap \mathbb{R}) - \min(\distancesMultiSet \cap \R)} : \gamma \in \distancesMultiSet \cap \R\right\}.$$

Then, $\entropy$ denotes the \emph{overall entropy of distances} quantifying the variability of distance values in $\normedDistancesMultiSet$ defined as Shannon's entropy \cite{Shannon1948} 

\begin{equation}
     \entropy := \sum_{i \in [k']} p_i \log_2(1/p_i),
    \label{eq:shannon-entropy}
\end{equation}

where $k$ is the number of equal-width bins over $[0,1]$, $k' \leq k$ is the number of non-empty bins, and $p_i$ is the proportion of distances falling in bin $i$ defined as $p_i := n_i/|\normedDistancesMultiSet|$ with $n_i$ being the count of distances in non-empty bin $i \in [k']$. 

Finally, we write $\distancesMultiSetForLevelSet$ for the subset of finite, normalised distances $\distancesMultiSetForLevelSet \subset \mathfrak{D}$ falling into level set $\levelSet$ defined as $\distancesMultiSetForLevelSet := \{\distanceValue_{m,y}^{x} : x,y \in [\nRelaxedPatterns], m \in [\nSegments], d_{L1}(\canonicalPattern[x],\canonicalPattern[y])= \levelSetIndex\}$. Then, $\entropy[\distancesMultiSetForLevelSet]$ denotes the \emph{level set entropy} quantifying the variability of distance values in $\distancesMultiSetForLevelSet$ defined using Equation \ref{eq:shannon-entropy}.

Then the \emph{average entropy} across all level sets is defined as
\begin{equation}
      \avgLevelSetEntropy := \frac{1}{|\levelSetDistances|}\sum_{\levelSetIndex\in\levelSetDistances}\entropy[\distancesMultiSetForLevelSet].
    \label{eq:average-entropy-level-sets}
\end{equation}

Together, canonical patterns (optimal clustering) and level sets (intuitive similarity relationships) provide the theoretical foundation required for validity assessment. In the methodology section, we describe how these theoretical constructs enable systematic evaluation of distance functions and ICVIs.

\section{Quantifying Clustering Quality}\label{sec:evaluation-reference-formulas}
This section presents the external and internal clustering validation indices (ICVIs) examined in this study, with formulas adapted to our notation. In our notation, the clustering result is a segmented clustering $\segmentedClustering$, and the objects to be clustered are correlation matrices $\correlationMatrix$.

\subsection{External Clustering Validity Index}\label{sec:external-validity-index}
External validation requires a ground truth segmented clustering $\partition[G]$ \cite{Omar2019}\cite{Omar2019}.

The Jaccard index is defined as
\begin{equation}
    J_O := \frac{\partition[G]\cap \partition}{\partition[G] \cup \partition}.
    \label{eq:jaccard-general}
\end{equation}
Eq.~\ref{eq:jaccard-general} is commonly adjusted to fit either time series clustering or segmentation domains, and defined as \cite{Vendramin2010}
\begin{equation}
J_S: = \frac{TP}{TP+FN+FP}.
\label{eq:standard-jaccard-index}
\end{equation}

For clustering without segmentation, $TP$ is the number of object pairs in the same cluster in both $\clustering[G]$ and $\clustering$; $FN$ is pairs together in $\clustering[G]$ but separated in $\clustering$; $FP$ is pairs separated in $\clustering[G]$ but together in $\clustering$. The Jaccard coefficient does not exclude true negatives ($TN$) hence measuring the accuracy of grouping objects that should be clustered together. This exclusion prevents the index from being dominated by the large number of objects that are correctly kept in separate clusters, making it attractive for clustering since in clustering problems, the majority of object pairs should remain ungrouped \cite{Vendramin2010}.

For segmentation without clustering, $TP$ is change points in $\segmentation[\partitionIndex]$ within tolerance zone $sz$ around change points in $\segmentation[G]$ (counting only the first per zone); $FP$ is additional change points within $sz$ plus those outside all zones; $FN$ is zones in $\segmentation[G]$ without detected change points \cite{Burg2020, Gensler2014}.

For our domain that combines segmentation and clustering (Eq.~\ref{eq:segmentedClustering}), we define the Jaccard index as
\begin{equation}
    J :=\frac{TP}{T},
    \label{eq:jaccard-index}
\end{equation}
where $TP$ is observations $t$ in the same cluster in both $\partition[G]$ and $\partition$, and $T$ is total observations. Since $FP+FN$ counts all observations that are not in the same cluster in $\partition[G]$ and $\partition$ by segmentation or clustering errors it follows that $TP+FN+FP \equiv T$.

\subsection{Internal Clustering Validity Indices - ICVI}\label{sec:internal-validity-indices}
Internal clustering validity indices (ICVIs) assess clustering quality by quantifying intra-cluster cohesion and inter-cluster separation without ground truth information \cite{Omar2019}.

Some ICVIs make use of a cluster and data centroids. We define the \emph{cluster centroid} for our correlation-based domain as 
\begin{equation}
    \clusterCentroid := \segmentToCorrelation(\timeseries(\bigcup_{\segIndex \in \clusterIndices} {\segmentEndIndex[\segIndex-1]+1,\ldots,\segmentEndIndex[\segIndex]})),
    \label{eq:cluster-controid}
\end{equation}
where $\clusterIndices \in \clustering$. 

Similarly, we define the \emph{data centroid} as
\begin{equation}
    \correlationMatrix[\timeseries] := \segmentToCorrelation(\timeseries).
    \label{eq:data-controid}
\end{equation}

\paragraph{Silhouette Width Criterion (SWC):}
SWC assesses similarity of an object to its own cluster versus other clusters \cite{Rousseeuw1987}. Bounded by $[-1,1]$, higher values indicate better cohesion and separation; negative values indicate poor clustering. The silhouette of the $m$-th segment $\silhouetteIndex$ is defined as
\[\silhouetteIndex := \frac{\interClusterDistance -\intraClusterDistance}{\max(\intraClusterDistance, \interClusterDistance)},\] where $\intraClusterDistance$ is the average distance average distance to other correlation matrices in cluster $\cluster{\segmentedClustering[i]}$ that is defined as \[\intraClusterDistance:=\frac{1}{|\clusterIndices| - 1} \sum_{y \in \clusterIndices, y \neq \segIndex} \clusterDistance,\] and $\interClusterDistance$ is the average distance to all correlation matrices in the nearest other cluster $\cluster[o]{\segmentedClustering[i]}$ where $o\in\clustering$ and $o\neq k$, which is defined as \[\interClusterDistance:=\min \left( \frac{1}{|\clusterIndices[o]|} \sum_{y \in \clusterIndices[o]} \clusterDistance\right).\] SWC is defined as
\begin{equation}
    SWC := \frac{1}{\nSegments}\sum_{\segIndex\in[\nSegments]}\silhouetteIndex.
    \label{eq:scw}
\end{equation}

\paragraph{Davies Bouldin Index (DBI):}
DBI quantifies the ratio of within-cluster scatter to between-cluster separation. Bounded by $[0, \infty)$, lower values indicate better quality \cite{Davies1979}. For cluster $\cluster{\segmentedClustering[i]}$, the average distance $\averageClusterDistance$ to the cluster centroid $\clusterCentroid$ is defined as \[\averageClusterDistance:=\frac{1}{|\clusterIndices|}\sum_{m \in \clusterIndices} \clusterDistance[\clusterCentroid].\] DBI is defined as
\begin{equation}
    DBI := \frac{1}{\nClusters}\sum_{\clusterIndex\in[\nClusters]}\underset{y\in[\nClusters],y \neq \clusterIndex}{\max}\left(\frac{\averageClusterDistance+\averageClusterDistance[y]}{d(\clusterCentroid, \clusterCentroid[y])}\right)
    \label{eq:dbi}
\end{equation}

\paragraph{Calinski-Harabasz Index or Variance Ratio Criterion (VRC):}
VRC quantifies the ratio of between-cluster to within-cluster dispersion \cite{Calinski1974}. It takes values in $[0, \infty)$; higher values indicate better quality. Values $<100$ typically indicate poor clustering.
The between-cluster variance $BCV$ is defined as \[BCV:=\frac{1}{\nClusters-1}\sum_{k \in [\nClusters]} |\clusterIndices| d(\clusterCentroid,\correlationMatrix[\timeseries])^2,\] where $\clusterIndices\in \clustering$. The within-cluster variance $WCV$ is defined as \[WCV:=\frac{1}{\nSegments-\nClusters}\sum_{k \in [\nClusters]}\sum_{\segIndex \in \clusterIndices}\clusterDistance[\clusterCentroid]^2.\] VRC is then defined as
\begin{equation}
    VRC = \frac{BCV}{WCV}.
    \label{eq:vrc}
\end{equation}

\paragraph{Pakhira-Bandyopadhyay-Maulik Index (PBM):}
The PBM index assesses compactness and separation whilst favouring fewer clusters \cite{Pakhira2004}. It takes values in $[0, \infty)$; higher values indicate better quality.
Compactness is measured by \[e_C:=\sum_{\clusterIndex \in [\nClusters]}\sum_{\segIndex \in \clusterIndices} \clusterDistance[\clusterCentroid],\] where $\clusterIndices\in\clustering$. The inter-cluster separation is measured by \[D_C:=\underset{\clusterIndex,y\in[\nClusters],y \neq \clusterIndex}{\max}(d(\clusterCentroid[k],\clusterCentroid[y])).\] To make the ICVI comparable across different datasets it uses a normalisation factor \[e_1:=\sum_{ \segIndex\in\nSegments}d(\correlationMatrix,\correlationMatrix[\timeseries]).\] PBM index is defined as
\begin{equation}
PBM=\left(\frac{1}{\nClusters}\frac{e_1}{e_C} D_C\right)^2.
\label{eq:pmb}
\end{equation}

\section{Methodology}\label{sec:method}
In this section, we present the method to establish ICVI validity to measure clustering quality. Since clustering quality assess how coherent and well separated clusters are \cite{Kaufman1990}, we must also establish distance function validity for measuring correlation matrix dissimilarity. We follow \citet{Salaudeen2025}'s validity theory framework for AI applications, closest to correlation-based clustering validation. We evaluate the validity of distance functions on their own and as a variable in ICVIs.

\citet{Salaudeen2025}'s framework requires defining claims and determining whether their objects are \emph{constructs} (abstract concepts without direct measurement) or \emph{criteria} (directly measurable concepts). Our claims:
\begin{enumerate}
    \item \textbf{Claim 1:} A distance function $d$ measures dissimilarity between correlation matrices and enables assessing whether a correlation matrix $\correlationMatrix[a]$ is more different from $\correlationMatrix[b]$ than $\correlationMatrix[c]$. \textbf{Object 1:} Dissimilarity of correlation matrices.
    \item \textbf{Claim 2:} An ICVI $I$ measures the quality of a correlation-based clustering outcome and enables assessing whether coherence within clusters and separation between clusters are higher for a segmented clustering $\segmentedClustering[a]$ than $\segmentedClustering[b]$ (see Eq. \ref{eq:segmentedClustering}). \textbf{Object 2:} Correlation-Based clustering quality.
\end{enumerate}

Since both objects of our claims (dissimilarity and clustering quality) are abstract constructs (not directly measurable), we require an explicit nomological network that maps the theoretical relationships between these constructs and their observable indicators \cite{Salaudeen2025}. Our nomological network comprises three interconnected components formalised in Section~\ref{sec:formalisation}: 
\begin{itemize}
    \item \textbf{Canonical Correlation Patterns} $\{\relaxedPattern\}_{\patternIndex \in [\nRelaxedPatterns]}$ as theoretical ideals representing maximally distinct (separated) correlation structures.
    \item \textbf{Level Sets} $\{\levelSet\}_{\levelSetIndex \in \levelSetDistances}$ as the theoretical similarity structure that intuitively groups pairs of correlation patterns by the L1 distance of their correlation coefficients.
    \item \textbf{Canonical Pattern Mapping} $\patternMap(\correlationMatrix)$ as the mechanism linking empirical correlation matrices to their most similar relaxed canonical pattern.    
\end{itemize}

Canonical patterns define optimal structure, level sets operationalise similarity relationships, and pattern mapping links empirical data to theory. Together, these components establish testable theoretical predictions that correlation patterns within the same level set should be more similar than patterns across different level sets, and that clustering outcomes that better separate into distinct canonical patterns should exhibit higher quality. 

The CSTS benchmark dataset serves as \emph{measurement instrument} to collect observations for evaluation \cite{degen2025csts}. CSTS operationalises our nomological network by modelling segmented clusterings of canonical correlation patterns under different conditions (distribution shifts, sparsification, downsampling). The dataset provides validated ground truth for ideal clustering outcomes in the correlation space (clustered by the canonical pattern used to generate the data), as well as controlled lower quality clustering outcomes with errors introduced through segmentation errors and cluster assignment errors. CSTS provides two equivalent, statistically independent datasets (exploratory and confirmatory), each with 30 subjects.

We evaluate construct validity (structural, convergent, discriminant), criterion validity (predictive and concurrent), content validity, and external validity. These evaluate \cite{Cronbach1955, Campbell1959, Messick1994, Salaudeen2025}: 
\begin{itemize}
    \item \textbf{Content Validity}: Does the measurement (distance functions, clustering indices) cover all relevant aspects of the construct (dissimilarity, clustering quality)?
    \item \textbf{Criterion Validity}: Does the measurement correlate with a known validated standard?
    \begin{itemize}
        \item \textit{Predictive Validity}: Can the measurement predict downstream outcomes?
        \item \textit{Concurrent Validity}: Does the measurement agree with a standard measurement that uses ground truth?
    \end{itemize}  
    \item \textbf{Construct Validity}: Does the measurement actually measure the theoretical construct?
    \begin{itemize}
        \item \textit{Structural Validity}: Does the measurement match our theoretical predictions following our nomological network?
        \item \textit{Convergent Validity}: Do different measurements of the same construct agree?
        \item \textit{Discriminant Validity}: Does the measurement of distinct constructs diverge?
    \end{itemize}
    \item \textbf{External Validity}: Do the findings generalise to different conditions?
\end{itemize}

We do not assess consequential validity (practical implications for downstream decisions), as our evaluation focuses on fundamental properties of distance functions and ICVIs generally, not specific real-world applications.

We categorise the evaluation result into \emph{valid} (satisfies thresholds for the various aspects of validity), \emph{invalid} (fails 
the thresholds of validity), and \emph{optimal} (statistically significantly outperforms other measurements). To establish optimal measurements, we use two-phase testing (exploratory and confirmatory) to reduce false discoveries \citet{Nosek2018}. In the exploratory phase, we establish and preregister hypotheses using CSTS's exploratory dataset, ordered by effect size per data variant. In the confirmatory phase, we retest preregistered hypotheses using CSTS's confirmatory dataset, stopping at first failure to control family-wise error. We organise these tests into the following six families to control Type I error rates due to multiplicity:

\begin{enumerate}
    \item \textbf{Family 1 - Distance Function}: Tests whether an optimal distance function for canonical pattern mapping exists.
    \item \textbf{Family 2 - ICVI}: Tests whether there is an optimal distance function to maximise raw ICVI values on ground-truth clusterings.
    \item \textbf{Family 3 - ICVI}: Tests whether an optimal distance function exists to maximise correlation between each ICVI and the Jaccard index.
    \item \textbf{Family 4 - ICVI}: Tests whether an optimal ICVI exists measured by correlation with the Jaccard index when using each ICVI's optimal distance function.
    \item \textbf{Family 5 - ICVI}: Tests whether ICVI values do not significantly change depending on the number of clusters.
    \item \textbf{Family 6 - ICVI}: Tests whether ICVI values do not significantly change depending on the number of segments.
\end{enumerate}

A measurement can be called optimal within a family (not across families) when it passes confirmatory testing. If it fails the confirmatory testing, it remains valid, but no claim of statistical superiority can be made.

The following subsections detail the specific validity assessments for distance functions and ICVIs and specify the validity thresholds for a measurement.

\subsection{Establishing Validity for Distance Functions}\label{sec:method-distance-function-assessment}
We now describe our method to establish distance function validity for measuring correlation matrix dissimilarity.

\begin{table}[ht!]
\caption{Evaluations run to establish the various aspects of validity of a distance function as a measurement of dissimilarity between correlation matrices according to our nomological network. The threshold column defines when a distance function is valid for an evaluation.}
\label{tab:distance-function-validity-framework}
\centering
\small
\setlength{\tabcolsep}{4pt}
\begin{tabular*}{\columnwidth}{@{\extracolsep{\fill}}l p{0.28\columnwidth} p{0.4\columnwidth} c}
\toprule
\textbf{Validity Type} & \textbf{Evaluation} & \textbf{Establishes} & \textbf{Threshold} \\
\midrule
\textbf{Content} & Theoretical reasoning of tested aspects & Validity evaluations cover all relevant aspects of dissimilarity between correlation matrices. & N/A \\
\midrule
\textbf{Criterion} & & & \\[0.5em]
\textbf{Predictive} & Macro $F_1$ score of 1NN classifier & Distance function enables accurate downstream classification of correlation matrices to their ground truth canonical pattern. & $F_1>0.98$ \\[0.5em]
\midrule
\textbf{Construct} & & & \\[0.5em]
1. \textbf{Structural} & $\distanceFunctionCorrxPatterny{x}{x} \approx 0$, for all $\segIndex$ in CSTS and $x \in [\nRelaxedPatterns]$ & Identity preservation: Empirical correlation matrices have near-zero dissimilarity to their ground truth relaxed canonical pattern for which they were generated. & $d\leq0.1$ \\[0.5em]
2. \textbf{Structural} & $\avgLevelSetDistance[i] <^* \avgLevelSetDistance[j]$ for $i,j \in \levelSetDistances, \,j=i+1$, and $^* p<0.05$ & Ordinal structure preservation: Dissimilarity increases significantly and monotonically the more the coefficients differ between adjacent level sets. & Pass/Fail \\[0.5em]
3. \textbf{Structural} & high $\avgRateOfIncreaseLevelSet{i}{j}$ for $i,j \in \levelSetDistances$ and $j=i+1$ & Average Sensitivity: the larger the increases in dissimilarity between level sets, the more strongly the distance function distinguishes between different correlation matrices. & $\avgRateOfIncreaseLevelSet{i}{j} > 0.7$ \\[0.5em]
4. \textbf{Structural} & high overall entropy $\entropy$ & Discrimination: the greater the variation in distances across all pairwise comparisons of correlation matrices, the more distinct measurements the distance function produces. & $\entropy>4$ \\[0.5em]
5. \textbf{Structural} & low average level set entropy $\avgLevelSetEntropy$ & Consistency: the more similar the distances within a level set are, the more uniformly the distance function treats equivalent structural differences between different correlation matrices. & $\avgLevelSetEntropy<3$ \\[0.5em]
\textbf{Convergent} & Multiple distance functions pass structural tests & Different measurements of the dissimilarity construct agree. & Pass/Fail \\[0.5em]
\textbf{Discriminant} & $d$ is invalid for structural tests 1-5 for raw data variants; and validity degrades for downsampled data variants & Measurement fails when the structure is absent or distorted. Canonical patterns are absent from raw data (apart from $\relaxedPattern[0]$). Downsampling distorts correlation patterns (MAE increases from 0.02 (SD 0.02) to 0.13 (SD 0.08)). & Pass/Fail \\
\midrule
\textbf{External} & Generalisation across data variants & Findings hold across normal sparsified (70\% \& 10\%), non-normal (100\% \& 10\%) conditions. & Pass/Fail \\
\bottomrule
\end{tabular*}
\end{table}

We evaluate the following  $\numberOfSelectedDistanceMeasures$ distance functions: $L_p$ norms ($\dLp$, see Eq.~\ref{eq:lp_norm}) and $L_p$ norms with a reference vector ($\dRefp$, see Eq.~\ref{eq:lp_norm_with_ref}) with $p=[1, 2, 3, 5, inf]$; $L_p$ norms dot transformed with $p=[1, 2, inf]$ ($\dDotp$, see Eq.~\ref{eq:lp_norm_dot_transform}); Log Frobenius ($\dlogF$, see Eq.~\ref{eq:log-cov-frobenius-distance}) and Förstner $d_F$ ($\dFoer$, see Eq.~\ref{eq:stable-foerstner-distance}). We select vector-based functions for directly quantifying coefficient differences and matrix-based functions for capturing matrix geometry. We focus on distance functions with direct geometric interpretation that are applicable to correlation matrices without data-dependent parameter tuning. We hypothesise some selected functions measure dissimilarity in ways supporting our nomological network's predictions.

Table \ref{tab:distance-function-validity-framework} shows evaluations for establishing construct, criterion, and external validity, with thresholds per evaluation. We argue that these evaluations together cover all relevant aspects of dissimilarity between correlation matrices as defined by our nomological network and hence establish content validity. We do not assess concurrent validity for criterion validity since there is no gold standard external measure available for measuring dissimilarity. Distance functions achieving thresholds for four of five structural, convergent, and discriminant evaluations show construct validity. For predictive criterion validity and external validity, they also need to pass the threshold for these evaluations.

Validity thresholds derive from correlation elliptope geometry explained in Section~\ref{sec:canonical-patterns}. Constructing valid strong correlation matrices requires adjusting coefficients from $\pm1$ to $\pm0.7$ to maintain positive semi-definiteness, establishing $0.7$ as the geometric boundary for valid strong correlations. We therefore expect the distances of valid distance functions to increase by at least $0.7$ per correlation coefficient change from 0 to 1, which is equivalent to a change in level set. For comparing empirical matrices to their ground truth patterns, we expect the difference over all coefficients to be $\leq 0.1$. Entropy thresholds ($\entropy>4$, $\avgLevelSetEntropy<3$) were empirically set to distinguish distance functions that passed discriminant validity tests from those that failed, ensuring that structural criteria identify measures robust to distortion. The $F_1 > 0.98$ threshold ensures a near-perfect classification that should be achievable for ideal patterns.

We determine whether a distance function is statistically optimal by ranking all valid functions on the structural validity tests 1-5 and the criterion validity test for each subject in the CSTS dataset per data variant. We compare average ranks using Wilcoxon signed rank tests following a step-down procedure \cite{Wilcoxon1945}, comparing best-ranked functions with sequentially lower-ranked until finding significant exploratory differences per data variant. Following our two-phase testing approach, we preregister the hypotheses and confirm whether they hold in the confirmatory phase. These tests correspond to Family 1, which establishes whether there is an optimal distance function for canonical pattern mapping per data variant. We also performed power analysis to provide details about the statistical robustness of our findings. 

For predictive criterion validity (Table~\ref{tab:distance-function-validity-framework}), we assess whether the distance functions correctly map empirical correlation matrices to their ground truth canonical patterns using 1-nearest neighbour (1NN) classification \cite{Cover1967NN, Dau2019, Grecki20241NN}. Relaxed canonical patterns $\nRelaxedPatterns$ serve as hardcoded neighbours. Each empirical matrix $\correlationMatrix$ in the CSTS dataset is assigned to its nearest canonical pattern $\relaxedPattern$ under distance function $d$, operationalising the pattern mapping capability (see Eq. \ref{eq:canonical-pattern-mapping}). We use the$ F_{1\text{macro}}$ score to assess accuracy (Appendix~\ref{app:macro-f1-score}). Since this directly tests whether the geometry of the distance function aligns with the canonical pattern structure, the near-perfect classification ($F_1 > 0.98$) is achievable and validates the nomological network independently of the structural tests. Distance functions achieving near perfect classification successfully recover the ground truth canonical patterns without relying on the level set framework.

\subsection{Establishing Validity for ICVIs}\label{sec:method-internal-index-assessment}
Having established valid distance functions, we now describe establishing ICVI validity for measuring correlation-based clustering quality. Specifically, we evaluate whether different ICVIs reliably quantify the quality of segmented clusterings (see Equation~\ref{eq:segmentedClustering}) and identify which of the valid distance functions optimise their performance.

\begin{table}[ht!]
\caption{Evaluations run to establish the various aspects of validity of an ICVI as a measurement of clustering quality for correlation-based segmented clusterings according to our nomological network. The threshold column defines when a ICVI is valid for an evaluation.}
\label{tab:icvi-validity-framework}
\centering
\small
\setlength{\tabcolsep}{4pt}
\begin{tabular*}{\columnwidth}{@{\extracolsep{\fill}}l p{0.28\columnwidth} p{0.32\columnwidth} p{0.2\columnwidth}}
\toprule
\textbf{Validity Type} & \textbf{Evaluation} & \textbf{Establishes} & \textbf{Threshold} \\
\midrule
\textbf{Content} & Theoretical reasoning of tested aspects & Validity evaluations cover all relevant aspects of clustering quality for correlation-based clusterings. & N/A \\
\midrule
\textbf{Criterion} & & & \\[0.5em]
\textbf{Concurrent} & Strong correlation with Jaccard index (external validation using ground truth labels) & Internal clustering validation (without labels) agrees with external clustering validation (with ground truth). & $|r| > 0.5$ \\[0.5em]
\midrule
\textbf{Construct} & & & \\[0.5em]
1. \textbf{Structural} & Optimal ICVI scores for perfect ground-truth clustering & ICVI achieves theoretically optimal values when clusters perfectly separate canonical patterns with maximal within-cluster coherence. & $\text{SWC}>0.9$, $\text{DBI}<0.15$, $\text{VRC}>1000$, $\text{PBM}>10$ \\[0.5em]
2. \textbf{Structural} & Clustering with highest MAE achieves low quality score & ICVI achieves low quality socre for low quality clustering & $\text{SWC} <0$, $\text{DBI} >2$, $\text{VRC}<100$, $\text{PBM}<10$ \\[0.5em]
3. \textbf{Structural} & ICVI value is consistent for $\nClusters$ 23, 11, or 6 & Sensitivity to the number of clusters: ICVI performance remains consistent when number of clusters changes. &$\text{SWC} > 0.9$, $\text{DBI}<0.15$, $\text{VRC}>1000$, $\text{PBM}>10$ \\[0.5em]
4. \textbf{Structural} & ICVI value is consistent for $\nSegments$ 100, 50, or 25 & Sensitivity Object count: ICVI performance remains consistent when number of segments changes. &$\text{SWC}>0.9$, $\text{DBI}<0.15$, $\text{VRC}>1000$, $\text{PBM}>10$ \\[0.5em]
\textbf{Convergent} & Multiple ICVIs pass structural tests & Different measurements of the clustering quality construct agree. & Pass/Fail \\[0.5em]
\textbf{Discriminant} & ICVI validity degrades for raw and downsampled data variants & Measurement fails when canonical pattern structure is absent or distorted. & For raw $\text{SWC}<0$,  $\text{DBI}>100$, $\text{VRC}<1$, and $\text{PBM}<1$ \\
\midrule
\textbf{External} & Generalisation across data variants & Findings hold across normal sparsified (70\% \& 10\%), non-normal (100\% \& 10\%) conditions. & Pass/Fail \\
\bottomrule
\end{tabular*}
\end{table}

We evaluate the following ICVIs: Silhouette Width Criterion (SWC; see Eq. \ref{eq:scw}), Davies-Bouldin Index (DBI; see Eq. \ref{eq:dbi}), Calinksi-Harabasz index (VRC, see Eq. \ref{eq:vrc}), Pakhira-Bandyopadhyay-Maulik index (PBM, see Eq. \ref{eq:pmb}). We select indices based on consistent strong performance in comprehensive studiess \cite{Vendramin2010, Arbelaitz2013} and widespread adoption. Critically, we prioritise indices whose performance should theoretically remain stable regardless of the number of clusters $\nClusters$ or segments $\nSegments$ (objects in other studies), since not every canonical pattern necessarily appears in every dataset despite equally high clustering quality. We restrict ICVIs to valid distance functions identified in Section~\ref{sec:method-distance-function-assessment}.

Table \ref{tab:icvi-validity-framework} shows evaluations for establishing criterion, construct, and external validity. We argue that ICVI-distance function combinations that pass criterion, construct (Structural evaluations 1 and 2, and either 3 or 4), and external evaluations for all data conditions cover all relevant aspects of clustering quality as defined by our nomological network, and hence establish content validity and overall validity. We require passing all external conditions because the patterns remain equally well modelled across normal and non-normal, complete and sparse conditions. This was verified in CSTS by measuring MAE between ground truth and modelled correlation patterns \cite{degen2025csts} where MAE remained 0.02 (SD 0.02) for both normal and non-normal 100\% data and increased only minimally to 0.03 (SD 0.03) at 10\% sparsification. Since pattern quality is preserved, valid ICVIs should consistently recognise ground-truth clustering quality across these conditions.

Since, our dataset operationalises optimal clustering, ICVI-specific thresholds for structural validity tests ($\text{SWC}>0.9$, $\text{DBI}<0.15$, $\text{VRC}>1000$, and $\text{PBM}>10$) were empirically established from data variants where canonical patterns are well-modelled (MAE$<0.02$).

We determine optimal valid ICVI-distance combinations using Families 2-6: Family 2 for structural validity evaluation 1, Family 3 for criterion validity, Family 4 across structural construct and criterion validity, Family 5 for structural evaluation 2, Family 6 for structural evaluation 3. We use Wilcoxon signed rank tests \cite{Wilcoxon1945}. During the exploratory phase for Family 2 and 3, we rank distance functions by their performance for each ICVI per data variant, comparing the best-ranked distance function with sequentially lower-ranked functions until finding significant differences. For Family 4, we directly compare SWC and DBI using their optimal distance functions. For Families 5 and 6, we compare ICVI values for 100\%, 50\%, and 25\% of cluster, respectively, segment counts. Following our two-phase testing approach, we preregister the hypotheses and confirm whether they hold in the confirmatory phase for each data variant and each Family.

\section{Experimental Setup}\label{sec:eperimental-setup}
In this section, we describe the dataset, experimental configurations for establishing validity, and implementation details.

\subsection{Synthetic Benchmark Dataset}\label{sec:synthetic-data}
We used the open source CSTS benchmark dataset \cite{degen2025csts}, a correlation structure-specific synthetic benchmark that operationalises canonical patterns and provides controlled ground truth for validation. CSTS consists of exploratory and confirmatory parts, each comprising $\subjects=30$ different subjects. The exploratory and confirmatory parts are statistically equivalent but independent. For each subject, there are 12 different data variants of three-variate regime-switching time series data. 

\begin{figure}[h]
\centerline{\includegraphics[width=\columnwidth]{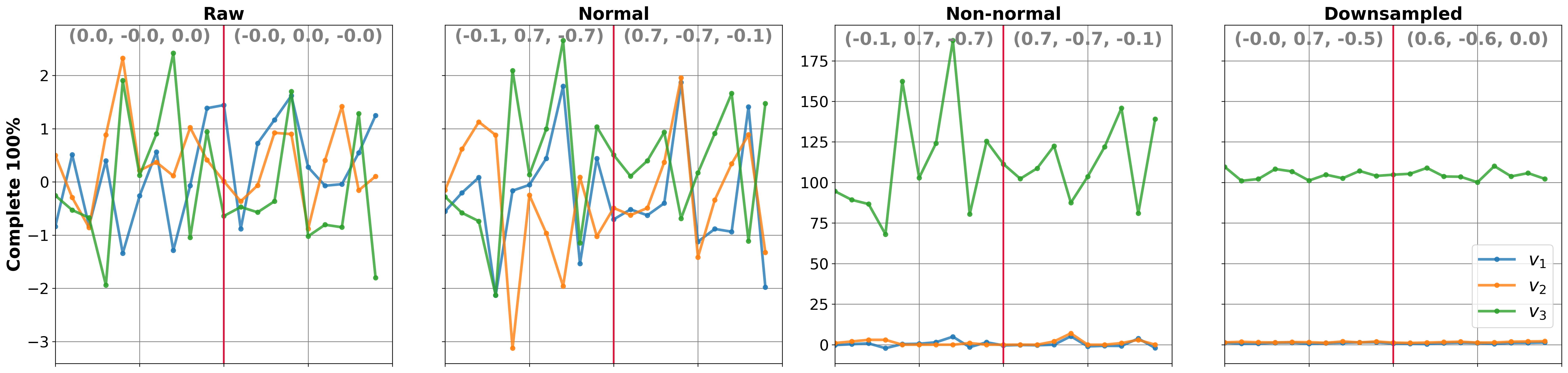}}
\caption{Example regime change (red line) between segments from canonical pattern 5 $[0,1,-1]$ to 15 $[1,-1,0]$ for the four generation stages: raw = uncorrelated data, normal = canonical correlation patterns, non-normal = distribution-shifted data, downsampled = temporally aggregated data. Empirical correlation coefficients ($a_{12}, a_{13}, a_{23}$) shown above each segment.}
\label{fig:timeseiers-for-complete}
\end{figure} 

The 12 data variants are made up of four generation stages (raw, normal, non-normal, downsampled) with each three completeness levels (complete, partial, sparse). The raw generation stage is normally distributed random data; in the normal stage each segment follows a different relaxed canonical pattern $\relaxedPattern$, ensuring each pattern is used $4-5$ times per subject; in the non-normal stage, the data undergo distribution shifts (extreme value and negative binomial distributions); and in the downsampled stage, the data are downsampled from 1-second to 1-minute resolution using mean aggregation. The three completeness levels for each stage represent all observations (complete 100\%), deleting 30\% of observations at random (partial 70\%), and deleting 90\% of observations at random (sparse 10\%). Figure~\ref{fig:timeseiers-for-complete} illustrates a regime change between two adjacent segments for the four generation stages. Each subject contains 100 segments of various lengths from 900 to 36,000 observations, resulting in approximately 1.26 million observations per subject before sparsification or downsampling. 

\begin{figure}[h]
\centerline{\includegraphics[width=0.6\columnwidth]{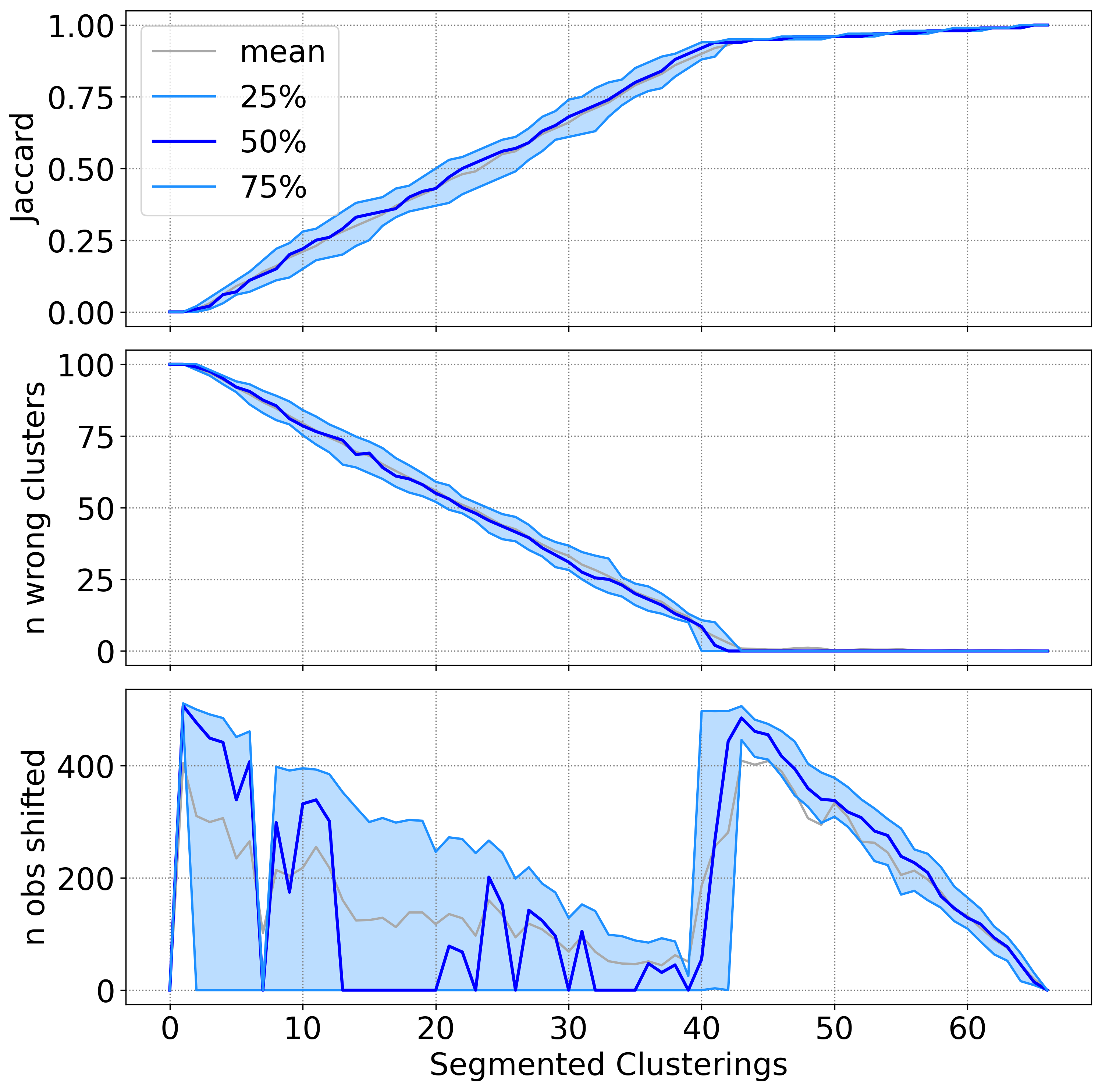}}
\caption{Descriptive statistics of Jaccard Index, number of segments assigned to wrong clusters, and number of observations shifted for 66 lower quality clusterings and the ground truth across the 30 exploratory subjects.}
\label{fig:jaccard-index-partitions}
\end{figure}

For each subject, CSTS contains $\badPartitions=\nBadPartitions$ lower quality segmented clusterings that span the entire range of the Jaccard index ($[0,1]$). The quality was degraded using the following three strategies: (1) shifting the segment end index forward by a randomly selected $n \in [1,800]$ observations, (2) randomly assigning $n \in [1,100]$ segments to incorrect cluster, and (3) combining (1) and (2). Figure~\ref{fig:jaccard-index-partitions} demonstrates that strategy (1) (last 22 lower quality segmented clusterings) has a lower impact on the Jaccard index than assigning whole segments to the wrong cluster.

Finally, CSTS provides versions with clusters reduced from 23 to 11 to 6, and segments reduced from 100 to 50 to 25.

\subsection{Establishing Validity for Distance Functions}\label{sec:exp-distance-functions-experiments-setup}  
For the tests in Table~\ref{tab:distance-function-validity-framework}, we calculated all distances $\distanceValue_{m,y}^{x} \in \R_{\geq 0}$ using each of the distance functions $\distanceFunctionCorrxPatterny{x}{y}$ for the collection of correlation matrices $\collectionOfCorrelationMatrices$ in the CSTS dataset for each data variant. Log Frobenius and Förstner distance functions required stabilisation to handle singular matrices common in canonical patterns (see Eq.~\ref{eq:log-cov-frobenius-distance} and Eq.~\ref{eq:stable-foerstner-distance}).

For the structural validity evaluation 2, statistical significance was assessed using 99\% confidence intervals with Bonferroni correction to account for comparing 5 level sets, maintaining 95\% overall confidence.    

For entropy calculations (structural evaluation 4-5), the number of bins $k$ to calculate $p_i$ for normalised distances was $50$ ($\sqrt{\nSegments*\nRelaxedPatterns}$). This resulted in a bin width of $1/k=0.02$. Empirical tests confirmed that for $k\geq50$, the relative ranking of distance functions remained stable across our entropy criteria. The number of distance values for overall entropy was $|\normedDistancesMultiSet|=2500$ distances, while the number of distance values for level set entropy ranged from $99$ distances ($\levelSet[5]$) to $662$ distances ($\levelSet[3]$).

To identify potentially optimal distance functions per data variant, we ranked the raw values from the structural validity evaluations 1-5 and the predictive criterion validation from 1 (best) to 6 (number of valid distance functions) per subject. Tied ranks were preserved. The binary pass/fail evaluations were coded as $1$ for pass and $0$ for failed. An average rank of all six tests was calculated per distance function per subject, weighting each evaluation equally. Wilcoxon signed rank tests were calculated for the average ranks, excluding pairs with differences $<0.001$ (SciPy exact p-value implementation). During the exploratory phase, we used two-sided tests with $\alpha=0.05$. Table~\ref{tab:preregistrated-family-1} shows the preregistered hypotheses from the exploratory phase and specifies the testing hierarchy for the confirmatory phase (ordered by largest effect size first). For the confirmatory phase, we used one-sided tests with $\alpha=0.05$. 

For power analysis and preregistration, we calculated the effect size $e$ as
\begin{equation}
    e := \frac{Z}{\sqrt{N_{nz}}},
    \label{eq:wilcox_effect_size}
\end{equation}
where $N_{nz}$ is the number of pairs with a difference $>0.001$, $Z$ is the z-score, derived from the p-value using the inverse normal cumulative distribution function. For exploratory two-sided tests we used $1-p/2$, for the confirmatory one-sided tests $1-p$. The achieved power is calculated as
\begin{equation}
    \text{Power} := \mathcal{N}((e\sqrt{N_{nz}}) - z_{\alpha}),
    \label{eq:wilcox_power}
\end{equation}
where $\mathcal{N}$ is the cumulative standard normal distribution function and $z_{\alpha}$ denotes the critical value for $\alpha$. The formula shows the confirmatory one-sided tests, for the exploratory two-sided tests we used $z_{\alpha/2}$.

\subsection{Establishing Validity for ICVIs}\label{sec:exp-internal-indices-experiments-setup}
For the tests in Table~\ref{tab:icvi-validity-framework}, we calculated all four ICVIs (SWC, DBI, VRC, PBM) for all valid distance functions and 67 segmented clusterings (66 lower quality and ground truth) per subject. For concurrent criterion validity, we calculated Pearson correlations between each ICVI and the Jaccard index.

For computational stability, division-by-zero cases in DBI were addressed by introducing minimal cluster centroid distance $d_{min}(\clusterCentroid, \clusterCentroid[y])=10^{-15}$ when identical centroids occurred. For data variants producing constant ICVI values, correlation coefficients were set to $0$ and p-values to $1$ to enable batch processing whilst indicating absence of correlation.

For criterion validity, sample size was calculated using \cite{Looney2020}
\begin{equation}
    \badPartitions := b + c^2(\frac{z_{\alpha} + z_{1-\beta}}{z(\rho_1)-z(\rho_0)})^2,
    \label{eq:correlation_sample_size}
\end{equation}
with $b=3$, $c^2=1$, $\rho_0=0.5$, $\rho_1=0.7$, $\alpha=0.05$, and $\beta=0.8$, where $z(\bullet)$ denotes the Fisher transform and $z_{\alpha}=1.645$, $z_{1-\beta}=0.84$ denote the Gaussian critical values. A sample size of $\geq 65$ lower quality clusterings ensures a power $\geq 80\%$ for combinations achieving $|r| \geq 0.5$.

Exploratory analysis for Family 2-6 used two-sided Wilcoxon signed rank tests with $\alpha=0.05$ (SciPy exact p-value implementation). The hypotheses are preregistered in Tables \ref{tab:preregistrated-family-2} to \ref{tab:preregistrated-family-6}. For Family 2, we excluded pairs with differences $<0.0001$. For SWC, higher values indicate better quality; for DBI, lower values indicate better quality, with hypotheses written to reflect this directionality. For Family 3, we excluded pairs with correlation differences $<0.001$. Since DBI exhibits negative correlation with the Jaccard index, whilst SWC shows positive correlation, the correlation coefficients of DBI were multiplied by $-1$ to enable comparison. For Family 4-6, we used the optimal distance function established in Family 2 and 3. For Family 4, we excluded pairs with differences $<0.001$ and multiplied DBI correlation coefficients by $-1$ for strength comparison. For Families 5 and 6, we excluded pairs with differences $<0.0001$.

For confirmatory testing across all families, we used one-sided Wilcoxon signed rank tests with $\alpha=0.05$, testing preregistered hypotheses in order of decreasing effect sizes from exploratory analysis and stopping upon the first non-significant result \cite{Nosek2018}. Power analysis used effect size and achieved power calculations following Equations~\ref{eq:wilcox_effect_size} and~\ref{eq:wilcox_power}.

\subsection{Implementation details}
Experiments were implemented in Python~3.9 using NumPy~1.26.4 for numerical computations, SciPy~1.11.3 for distance function calculations, statistical analysis, correlation calculations, and distributions, Pandas~2.2.2 for descriptive statistics and results aggregation, and Scikit-learn~1.4.2 for kNN, F1 score, and SWC calculations. The experiments were run on a MacBook Pro with an Apple M1 chip (8 cores: 4 performance, 4 efficiency) and 16GB RAM running macOS 15.x. The longest calculations were distance functions for all data variants ($\sim 5 hours$), and ICVIs for all 24,120 segmented clusterings ($\sim 8 hours$). All code is available
in our GitHub repository \url{https://github.com/isabelladegen/corrclust-validation} to
enable reproduction of our results.

\section{Results}\label{sec:results}
We present the validity assessment results in two parts. First, we assess the validity of the 15 distance functions. Second, we assess the validity of the four ICVIs using the valid distance functions identified in the first part.

\subsection{Distance Function Validity}
We assessed 15 distance functions across our criterion, construct, discriminant, and external validity criteria. Table \ref{tab:results-distance-validity} summarises the validity assessment across all evaluations and conditions. Six ($\dLp[1]$, $\dLp[2]$, $\dLp[3]$, $\dLp[5]$, $\dDotp[1]$, and $\dDotp[2]$) of the 15 distance functions achieved overall validity by passing the thresholds defined in Table~\ref{tab:distance-function-validity-framework} for all aspects. Correlation-specific functions ($\dlogF$, $\dFoer$) failed construct, criterion and external validity despite theoretical appeal. $\dLp[\inf]$ and reference-based functions ($\dRefp$) failed structural and external validity.

\begin{table}[ht!]
\caption{Validity assessment for distance functions across the different validity aspects: overall, criterion validity (Crit), construct validity (including structural (Str), convergent (Conv), discriminant for raw and downsampled (ds) 100\% conditions), and external validity (including normal 70\% and 10\% sparsification, and non-normal 100\% and 10\% sparsification conditions).}
\label{tab:results-distance-validity}
\centering
\small
\setlength{\tabcolsep}{3pt}
\begin{tabular*}{\columnwidth}{@{\extracolsep{\fill}}l c c c c c c c c c c}
\toprule
& & & \multicolumn{4}{c}{\textbf{Construct}} & \multicolumn{4}{c}{\textbf{External}} \\
\cmidrule(lr){4-7} \cmidrule(lr){8-11}
& \textbf{Overall} & \textbf{Crit} & \textbf{Str} & \textbf{Conv} & \multicolumn{2}{c}{\textbf{Discriminant}} & \multicolumn{2}{c}{\textbf{Normal}} & \multicolumn{2}{c}{\textbf{Non-normal}} \\
\cmidrule(lr){6-7} \cmidrule(lr){8-9} \cmidrule(lr){10-11}
\textbf{d} & & & & & raw & ds & 70\% & 10\% & 100\% & 10\%\\
\midrule
$\dLp[1]$ & \multicolumn{1}{c|}{\checkmark} & \multicolumn{1}{c|}{\checkmark} & \checkmark & \checkmark & \checkmark & \multicolumn{1}{c|}{\checkmark} & \checkmark & \checkmark & \checkmark & \checkmark \\
$\dLp[2]$ & \multicolumn{1}{c|}{\checkmark} & \multicolumn{1}{c|}{\checkmark} & \checkmark & \checkmark & \checkmark & \multicolumn{1}{c|}{\checkmark} & \checkmark & \checkmark & \checkmark & \checkmark \\
$\dLp[3]$ & \multicolumn{1}{c|}{\checkmark} & \multicolumn{1}{c|}{\checkmark} & \checkmark & \checkmark & \checkmark & \multicolumn{1}{c|}{\checkmark} & \checkmark & \checkmark & \checkmark & \checkmark \\
$\dLp[5]$ & \multicolumn{1}{c|}{\checkmark} & \multicolumn{1}{c|}{\checkmark} & \checkmark & \checkmark & \checkmark & \multicolumn{1}{c|}{\checkmark} & \checkmark & \checkmark & \checkmark & \checkmark \\
$\dLp[\infty]$ & \multicolumn{1}{c|}{\texttimes} & \multicolumn{1}{c|}{\checkmark} & \texttimes & \checkmark & \checkmark & \multicolumn{1}{c|}{\checkmark} & \texttimes & \checkmark & \texttimes & \checkmark \\
$\dRefp[1]$ & \multicolumn{1}{c|}{\texttimes} & \multicolumn{1}{c|}{\checkmark} & \texttimes & \checkmark & \checkmark & \multicolumn{1}{c|}{\checkmark} & \checkmark & \texttimes & \checkmark & \texttimes \\
$\dRefp[2]$ & \multicolumn{1}{c|}{\texttimes} & \multicolumn{1}{c|}{\checkmark} & \texttimes & \checkmark & \checkmark & \multicolumn{1}{c|}{\checkmark} & \checkmark & \texttimes & \checkmark & \texttimes \\
$\dRefp[3]$& \multicolumn{1}{c|}{\texttimes} & \multicolumn{1}{c|}{\checkmark} & \texttimes & \checkmark & \checkmark & \multicolumn{1}{c|}{\checkmark} & \texttimes & \texttimes & \texttimes & \texttimes \\
$\dRefp[5]$ & \multicolumn{1}{c|}{\texttimes} & \multicolumn{1}{c|}{\checkmark} & \texttimes & \checkmark & \checkmark & \multicolumn{1}{c|}{\checkmark} & \texttimes & \texttimes & \checkmark & \texttimes \\
$\dRefp[\infty]$ & \multicolumn{1}{c|}{\texttimes} & \multicolumn{1}{c|}{\checkmark} & \texttimes & \checkmark & \checkmark & \multicolumn{1}{c|}{\checkmark} & \checkmark & \texttimes & \checkmark & \texttimes \\
$\dDotp[1]$ & \multicolumn{1}{c|}{\checkmark} & \multicolumn{1}{c|}{\checkmark} & \checkmark & \checkmark & \checkmark & \multicolumn{1}{c|}{\checkmark} & \checkmark & \checkmark & \checkmark & \checkmark \\
$\dDotp[2]$ & \multicolumn{1}{c|}{\checkmark} & \multicolumn{1}{c|}{\checkmark} & \checkmark & \checkmark & \checkmark & \multicolumn{1}{c|}{\checkmark} & \checkmark & \checkmark & \checkmark & \checkmark \\
$\dDotp[\infty]$ & \multicolumn{1}{c|}{\texttimes} & \multicolumn{1}{c|}{\checkmark} & \texttimes & \checkmark & \checkmark & \multicolumn{1}{c|}{\checkmark} & \texttimes & \checkmark & \texttimes & \checkmark \\
$\dlogF$ & \multicolumn{1}{c|}{\texttimes} & \multicolumn{1}{c|}{\texttimes} & \texttimes & \checkmark & \checkmark & \multicolumn{1}{c|}{\checkmark} & \texttimes & \texttimes & \texttimes & \texttimes \\
$\dFoer$ & \multicolumn{1}{c|}{\texttimes} & \multicolumn{1}{c|}{\texttimes} & \texttimes & \checkmark & \checkmark & \multicolumn{1}{c|}{\checkmark} & \texttimes & \texttimes & \texttimes & \texttimes \\
\bottomrule
\end{tabular*}
\end{table}

\begin{figure*}[ht]
    \centerline{\includegraphics[width=\textwidth]{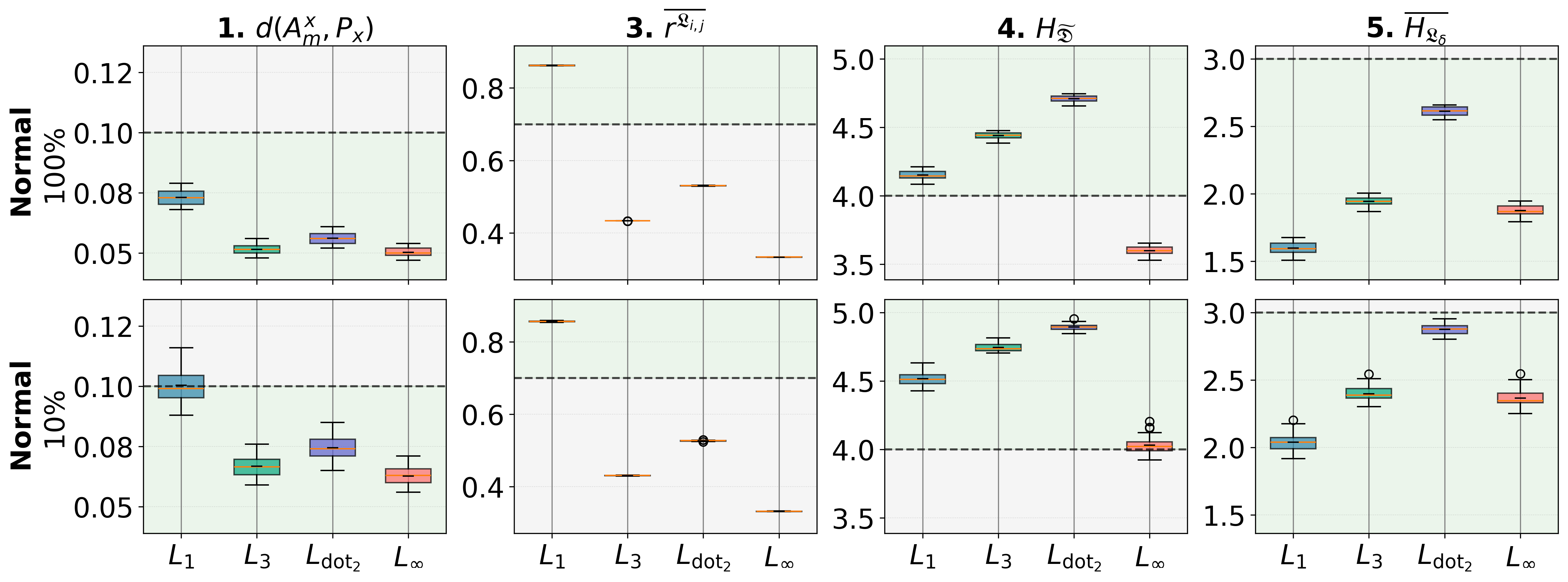}}
    \caption{Structural validity performance four valid ($\dLp[1]$, $\dLp[3]$, $\dDotp[2]$) and the invalid $\dLp[\infty]$ distance functions for tests 1, and 3-5 (columns) for the normal complete and sparse data variants. Dashed lines indicate validity thresholds. Green shading = valid region, grey = invalid. Box plots show median, interquartile range, and outliers for the 30 exploratory subjects.}
    \label{fig:distance-structural-validity}
\end{figure*}

Figure~\ref{fig:distance-structural-validity} illustrates the structural performance for valid ($\dLp[1]$), optimal in some conditions ($\dLp[3]$, $\dDotp[2]$), and invalid ($\dLp[\infty]$) distance functions for the normal and non normal complete (100\%) and sparse (10\%) data conditions. $\dLp[1]$ uniquely passed all five structural criteria in all conditions. Other valid functions failed one of the five structural criteria: $\dLp[3]$, $\dLp[\infty]$, and the dot-transformed functions failed the threshold for the average rate of increase between level sets ($\avgRateOfIncreaseLevelSet{i}{j}$, Column 2), achieving mean values of 0.33 ($\dLp[\infty]$) to 0.53 ($\dDotp[2]$) rather than the expected $>0.7$. The invalid function $\dLp[\infty]$ additionally struggled with overall entropy ($\entropy$, Column 4) particularly for complete data variants achieving mean values of 3.6 instead of the 4.0 threshold. All Lp norm based distance functions passed structural test 2 that requires monotonic increase of average distances between level sets. They further achieved a perfect $F_1$ score for criterion validity classifying empirical correlation matrices correctly to their canonical pattern in all data conditions, whilst appropriately degrading for downsampled data to 0.84-0.91 and raw data (no patterns) to 0. See Appendix Tables~\ref{tab:construct-mean-sd-normal-100} to~\ref{tab:external-mean-sd-non-normal-10} for mean and standard deviation for the 5 structural evaluations for construct validity and the predictive evaluation for criterion validity.detailed test results (mean and SD per test, per condition).

\begin{figure*}[t]
    \centerline{\includegraphics[width=\textwidth]{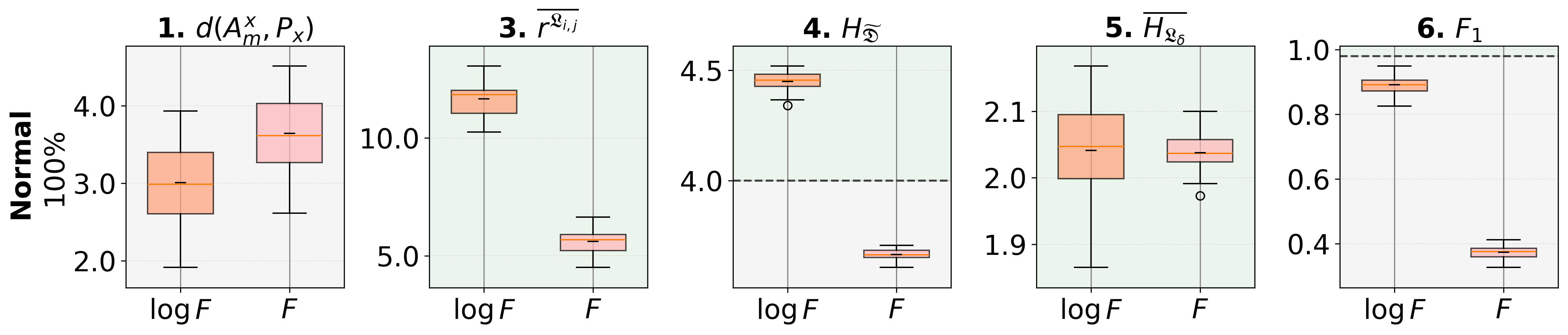}}
    \caption{Structural validity performance for the invalid correlation-specific distance functions ($\dlogF$ and $\dFoer$) on the normal complete (100\%) data for evaluations 1 and 3-5 as well as the criterion validity ($F_1$ score). Dashed lines indicate validity thresholds. Green shading = valid region, grey = invalid. Box plots show median, interquartile range, and outliers for the 30 exploratory subjects.}
    \label{fig:correlation-specific-failures}
\end{figure*}

Figure~\ref{fig:correlation-specific-failures} shows the results for the structural validity evaluations of the invalid correlation-specific distance functions $\dlogF$ and $\dFoer$ for the normal, complete data. Although designed specifically for correlation matrices, both correlation-specific distance functions fail multiple structural, and external validity criteria.

$\dlogF$ achieves a strong average rate of increase in distances between level sets ($\avgRateOfIncreaseLevelSet{i}{j}=11.65\pm0.77$), far exceeding the 0.7 threshold and appropriate overall and level set entropy characteristics. However, it fails identity preservation for which empirical correlation matrices should be similar to their canonical ground truth pattern ($\distanceFunctionCorrxPatterny{x}{x}=3.01\pm0.56$ vs threshold $<0.1$), ordinal validity (distances do not monotonically increase between level sets), and criterion validity ($F_1=0.89\pm0.03$ vs threshold $>0.98$). The results indicate that the measure captures some differences between patterns, but cannot establish near-zero dissimilarity when comparing a matrix to its own ground truth pattern, does not capture fewer coefficient changes as more similar (level set structure), and struggles with downstream applications. $\dFoer$ shows similar discrimination properties but more severe failures. It also fails identity preservation ($\distanceFunctionCorrxPatterny{x}{x}=3.65\pm0.53$ vs threshold $<0.1$), ordinal validity, and criterion validity ($F_1=0.37\pm0.02$ vs threshold $>0.98$). In addition to $\dlogF$, $\dFoer$ also fails overall entropy ($\entropy=3.67\pm$ versus threshold $>4$). The low $F_1$ score prevents accurate pattern mapping.

For the six valid distance functions, we also evaluated whether an optimal distance function exists for specific data conditions (Family 1). The range in mean ranks across the distance functions within each condition was narrow (0.20–0.30), indicating similar performance, see Appendix Table~\ref{tab:distance-measure-avg-ranks}. However, these small differences for some  data conditions are consistent enough to be statistically significant. $\dLp[3]$ was confirmed optimal for normal 100\% and 70\% and non-normal 100\% data conditions. $\dDotp[2]$ was confirmed optimal for the non-normal 70\% data, whilst the confirmatory test failed for the sparse 10\% data. See Appendix Tables~ \ref{tab:preregistrated-family-1} and~\ref{tab:family1_statistical_validation_results} for details of the statistical results.

\subsection{ICVIs Validity}
We assessed four ICVIs, each with the six valid distance functions ($\dLp[1]$, $\dLp[2]$, $\dLp[3]$, $\dLp[5]$, $\dDotp[1]$, and $\dDotp[2]$), across our criterion, construct, discriminant, and external validity criteria. Table \ref{tab:results-indices-validity} summarises the validity assessment across all evaluations and conditions. SWC achieves overall validity for all valid distance functions other than $\dLp[1]$ and $\dDotp[1]$ that do not pass external validity (failing structural evaluation 1) for the non-normal, sparse data variant. DBI achieves overall validity for distance functions $\dLp[3]$ and $\dLp[5]$. DBI fails external validity for the sparse normal and non-normal data variants for distance functions $\dLp[1]$, and $\dDotp[1]$. For the non-normal, sparse data variant only distance functions $\dLp[3]$ and $\dLp[5]$ achieve construct validity. VRC and PBM fail criterion validity for their weak correlations with the Jaccard index (mean correlations 0.21 to 0.34), while SWC and DBI are strongly correlated to the Jaccard index (mean correlations 0.89 to 0.92). PBM also fails structural validity for all distance functions other than $\dLp[1]$, and $\dDotp[1]$ for the normal, complete data variant, as well as external validity.

\begin{table}[ht!]
\caption{Validity assessment for ICVIs using valid distance functions across the different validity aspects: construct validity (including structural (Str), convergent (Conv), discriminant for raw and downsampled (ds) 100\% conditions), criterion validity (Crit), and external validity (including normal 70\% and 10\% sparsification, and non-normal 100\% and 10\% sparsification conditions).}
\label{tab:results-indices-validity}
\centering
\small
\setlength{\tabcolsep}{3pt}
\begin{tabular*}{\columnwidth}{@{\extracolsep{\fill}}l c c c c c c c c c c}
\toprule
& & & \multicolumn{4}{c}{\textbf{Construct}} & \multicolumn{4}{c}{\textbf{External}} \\
\cmidrule(lr){4-7} \cmidrule(lr){8-11}
& \textbf{Overall} & \textbf{Crit} & \textbf{Str} & \textbf{Conv} & \multicolumn{2}{c}{\textbf{Discriminant}} & \multicolumn{2}{c}{\textbf{Normal}} & \multicolumn{2}{c}{\textbf{Non-normal}} \\
\cmidrule(lr){6-7} \cmidrule(lr){8-9} \cmidrule(lr){10-11}
\textbf{d} & & & & & raw & ds & 70\% & 10\% & 100\% & 10\%\\
\midrule
\textbf{SWC} & & & & & & & & & & \\
$\dLp[1]$ & \multicolumn{1}{c|}{\texttimes} & \multicolumn{1}{c|}{\checkmark} & \checkmark & \checkmark & \checkmark & \multicolumn{1}{c|}{\checkmark} & \checkmark & \checkmark & \checkmark & \texttimes \\
$\dLp[2]$ & \multicolumn{1}{c|}{\checkmark} & \multicolumn{1}{c|}{\checkmark} & \checkmark & \checkmark & \checkmark & \multicolumn{1}{c|}{\checkmark} & \checkmark & \checkmark & \checkmark & \checkmark \\
$\dLp[3]$ & \multicolumn{1}{c|}{\checkmark} & \multicolumn{1}{c|}{\checkmark} & \checkmark & \checkmark & \checkmark & \multicolumn{1}{c|}{\checkmark} & \checkmark & \checkmark & \checkmark & \checkmark \\
$\dLp[5]$ & \multicolumn{1}{c|}{\checkmark} & \multicolumn{1}{c|}{\checkmark} & \checkmark & \checkmark & \checkmark & \multicolumn{1}{c|}{\checkmark} & \checkmark & \checkmark & \checkmark & \checkmark \\
$\dDotp[1]$ & \multicolumn{1}{c|}{\texttimes} & \multicolumn{1}{c|}{\checkmark} & \checkmark & \checkmark & \checkmark & \multicolumn{1}{c|}{\checkmark} & \checkmark & \checkmark & \checkmark & \texttimes \\
$\dDotp[2]$ & \multicolumn{1}{c|}{\checkmark} & \multicolumn{1}{c|}{\checkmark} & \checkmark & \checkmark & \checkmark & \multicolumn{1}{c|}{\checkmark} & \checkmark & \checkmark & \checkmark & \checkmark \\
\midrule
\textbf{DBI} & & & & & & & & & & \\
$\dLp[1]$ & \multicolumn{1}{c|}{\texttimes} & \multicolumn{1}{c|}{\checkmark} & \checkmark & \checkmark & \checkmark & \multicolumn{1}{c|}{\checkmark} & \checkmark & \texttimes & \checkmark & \texttimes \\
$\dLp[2]$ & \multicolumn{1}{c|}{\texttimes} & \multicolumn{1}{c|}{\checkmark} & \checkmark & \checkmark & \checkmark & \multicolumn{1}{c|}{\checkmark} & \checkmark & \checkmark & \checkmark & \texttimes \\
$\dLp[3]$ & \multicolumn{1}{c|}{\checkmark} & \multicolumn{1}{c|}{\checkmark} & \checkmark & \checkmark & \checkmark & \multicolumn{1}{c|}{\checkmark} & \checkmark & \checkmark & \checkmark & \checkmark \\
$\dLp[5]$ & \multicolumn{1}{c|}{\checkmark} & \multicolumn{1}{c|}{\checkmark} & \checkmark & \checkmark & \checkmark & \multicolumn{1}{c|}{\checkmark} & \checkmark & \checkmark & \checkmark & \checkmark \\
$\dDotp[1]$ & \multicolumn{1}{c|}{\texttimes} & \multicolumn{1}{c|}{\checkmark} & \checkmark & \checkmark & \checkmark & \multicolumn{1}{c|}{\checkmark} & \checkmark & \texttimes & \checkmark & \texttimes \\
$\dDotp[2]$ & \multicolumn{1}{c|}{\texttimes} & \multicolumn{1}{c|}{\checkmark} & \checkmark & \checkmark & \checkmark & \multicolumn{1}{c|}{\checkmark} & \checkmark & \checkmark & \checkmark & \texttimes \\
\midrule
\textbf{VRC} & & & & & & & & & & \\
$\dLp[1]$ & \multicolumn{1}{c|}{\texttimes} & \multicolumn{1}{c|}{\texttimes} & \checkmark & \checkmark & \checkmark & \multicolumn{1}{c|}{\checkmark} & \checkmark & \checkmark & \checkmark & \checkmark \\
$\dLp[2]$ & \multicolumn{1}{c|}{\texttimes} & \multicolumn{1}{c|}{\texttimes} & \checkmark & \checkmark & \checkmark & \multicolumn{1}{c|}{\checkmark} & \checkmark & \checkmark & \checkmark & \checkmark \\
$\dLp[3]$ & \multicolumn{1}{c|}{\texttimes} & \multicolumn{1}{c|}{\texttimes} & \checkmark & \checkmark & \checkmark & \multicolumn{1}{c|}{\checkmark} & \checkmark & \checkmark & \checkmark & \checkmark \\
$\dLp[5]$ & \multicolumn{1}{c|}{\texttimes} & \multicolumn{1}{c|}{\texttimes} & \checkmark & \checkmark & \checkmark & \multicolumn{1}{c|}{\checkmark} & \checkmark & \checkmark & \checkmark & \checkmark \\
$\dDotp[1]$ & \multicolumn{1}{c|}{\texttimes} & \multicolumn{1}{c|}{\texttimes} & \checkmark & \checkmark & \checkmark & \multicolumn{1}{c|}{\checkmark} & \checkmark & \checkmark & \checkmark & \checkmark \\
$\dDotp[2]$ & \multicolumn{1}{c|}{\texttimes} & \multicolumn{1}{c|}{\texttimes} & \checkmark & \checkmark & \checkmark & \multicolumn{1}{c|}{\checkmark} & \checkmark & \checkmark & \checkmark & \checkmark \\
\midrule
\textbf{PBM} & & & & & & & & & & \\
$\dLp[1]$ & \multicolumn{1}{c|}{\texttimes} & \multicolumn{1}{c|}{\texttimes} & \checkmark & \texttimes & \checkmark & \multicolumn{1}{c|}{\checkmark} & \checkmark & \texttimes & \checkmark & \texttimes \\
$\dLp[2]$ & \multicolumn{1}{c|}{\texttimes} & \multicolumn{1}{c|}{\texttimes} & \texttimes & \texttimes & \checkmark & \multicolumn{1}{c|}{\checkmark} & \texttimes & \texttimes & \texttimes & \texttimes \\
$\dLp[3]$ & \multicolumn{1}{c|}{\texttimes} & \multicolumn{1}{c|}{\texttimes} & \texttimes & \texttimes & \checkmark & \multicolumn{1}{c|}{\checkmark} & \texttimes & \texttimes & \texttimes & \texttimes \\
$\dLp[5]$ & \multicolumn{1}{c|}{\texttimes} & \multicolumn{1}{c|}{\texttimes} & \texttimes & \texttimes & \checkmark & \multicolumn{1}{c|}{\checkmark} & \texttimes & \texttimes & \texttimes & \texttimes \\
$\dDotp[1]$ & \multicolumn{1}{c|}{\texttimes} & \multicolumn{1}{c|}{\texttimes} & \checkmark & \texttimes & \checkmark & \multicolumn{1}{c|}{\checkmark} & \checkmark & \texttimes & \checkmark & \texttimes \\
$\dDotp[2]$ & \multicolumn{1}{c|}{\texttimes} & \multicolumn{1}{c|}{\texttimes} & \texttimes & \texttimes & \checkmark & \multicolumn{1}{c|}{\checkmark} & \texttimes & \texttimes & \texttimes & \texttimes \\
\bottomrule
\end{tabular*}
\end{table}

\begin{figure}[h!]
\centering
\begin{subfigure}[b]{0.49\textwidth}
    \centering
    \includegraphics[width=\textwidth]{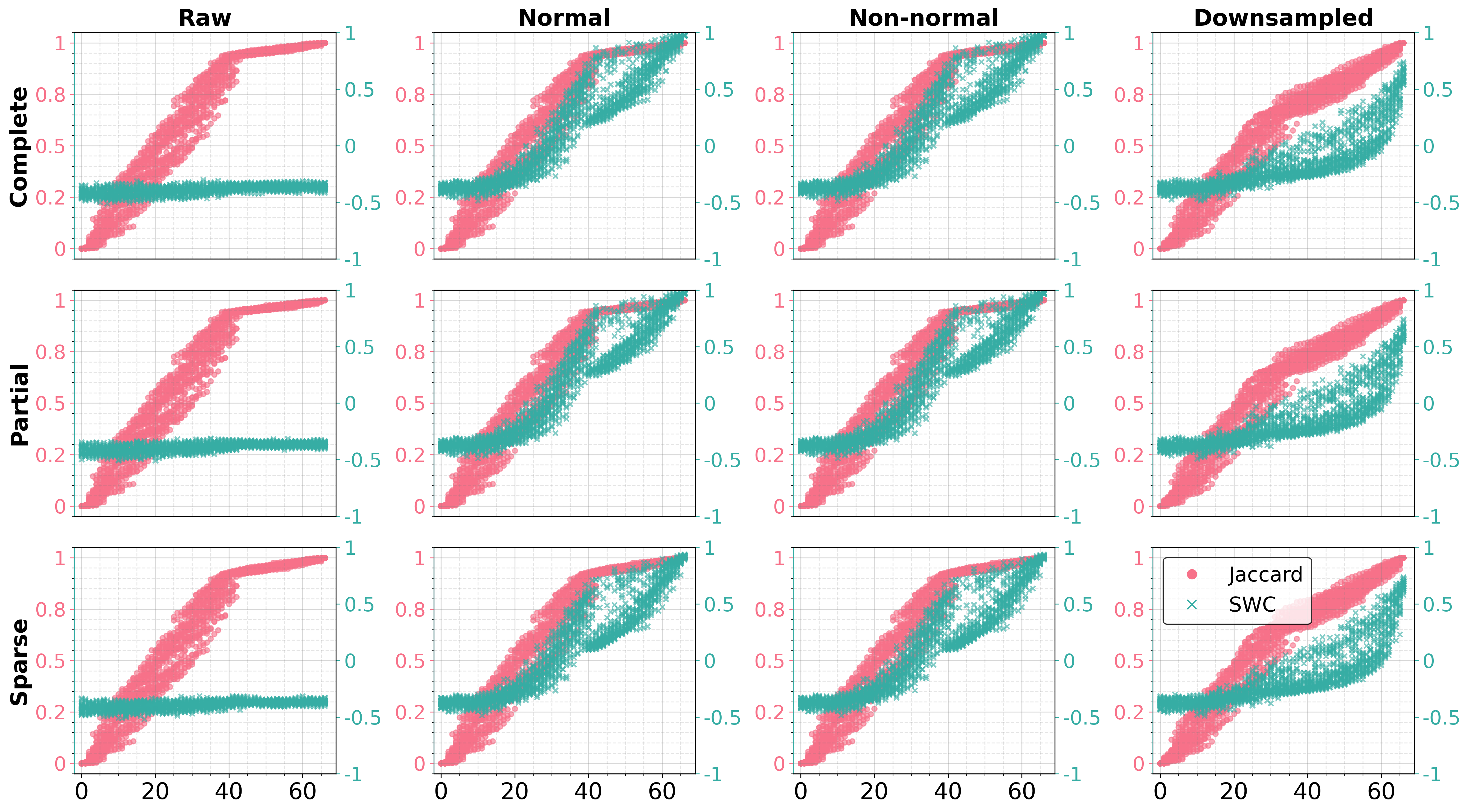}
    \caption{SWC}
    \label{fig:scatter-plot-scw}
\end{subfigure}
\hfill
\begin{subfigure}[b]{0.49\textwidth}
    \centering
    \includegraphics[width=\textwidth]{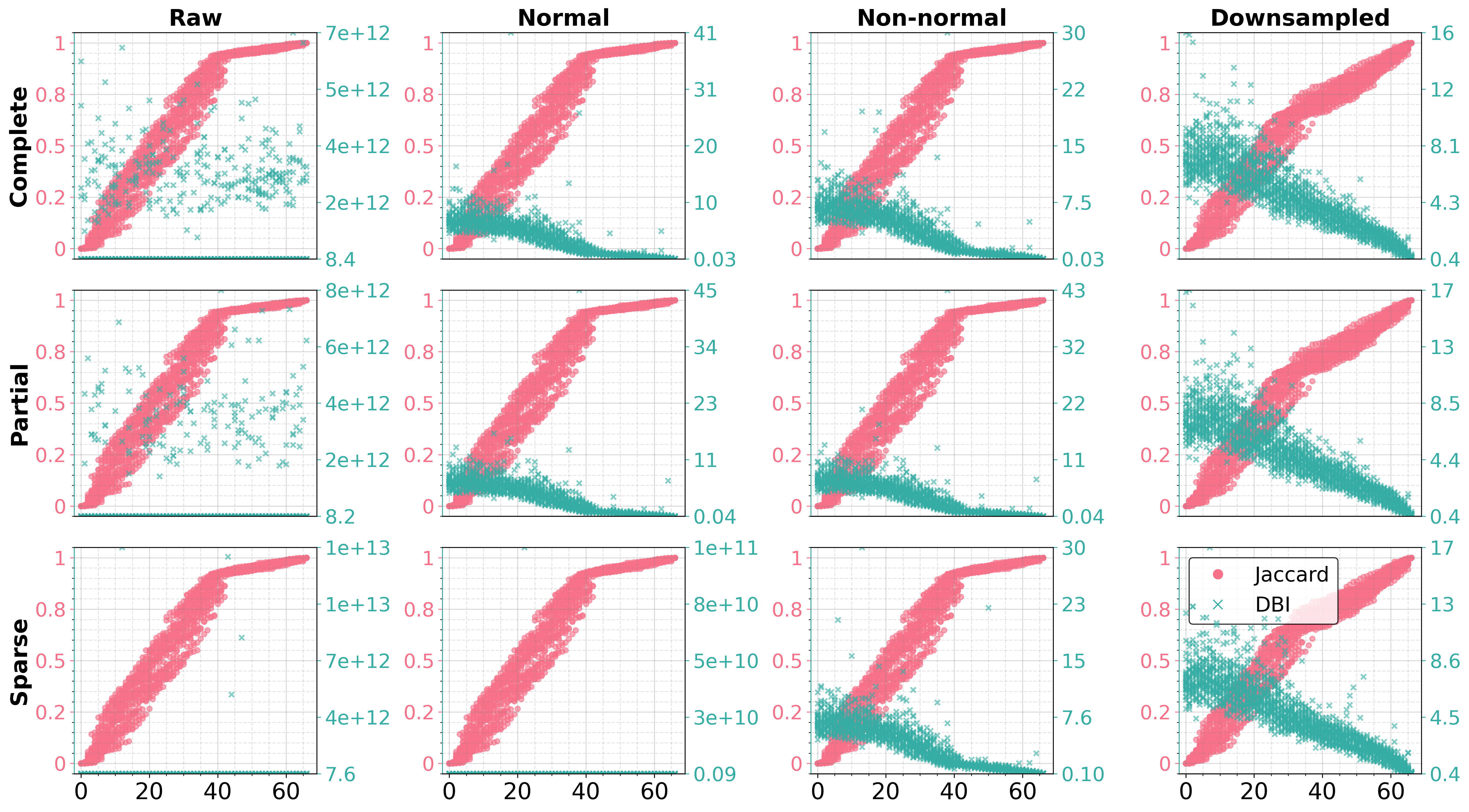}
    \caption{DBI}
    \label{fig:scatter-plot-dbi}
\end{subfigure}
\caption{Jaccard index (pink, left y-axes) and (a) SWC, (b) DBI (green, right y-axes) across data variants. Normal and Non-normal data conditions are use to establish construct and external validity. Raw variants lack correlation patterns, while downsampling data distorts patterns establishing discriminant validity. Each subplot shows 67 clusterings of varying quality for the 30 exploratory subject.}
\label{fig:combined-scatter_plots}
\end{figure}

Figure~\ref{fig:combined-scatter_plots} illustrates the strong correlation between SWC and DBI with the Jaccard index across data conditions that supports their criterion validity, as well as their discriminant validity through degradation for raw and downsampled variants.

\begin{figure*}[ht]
    \centerline{\includegraphics[width=\textwidth]{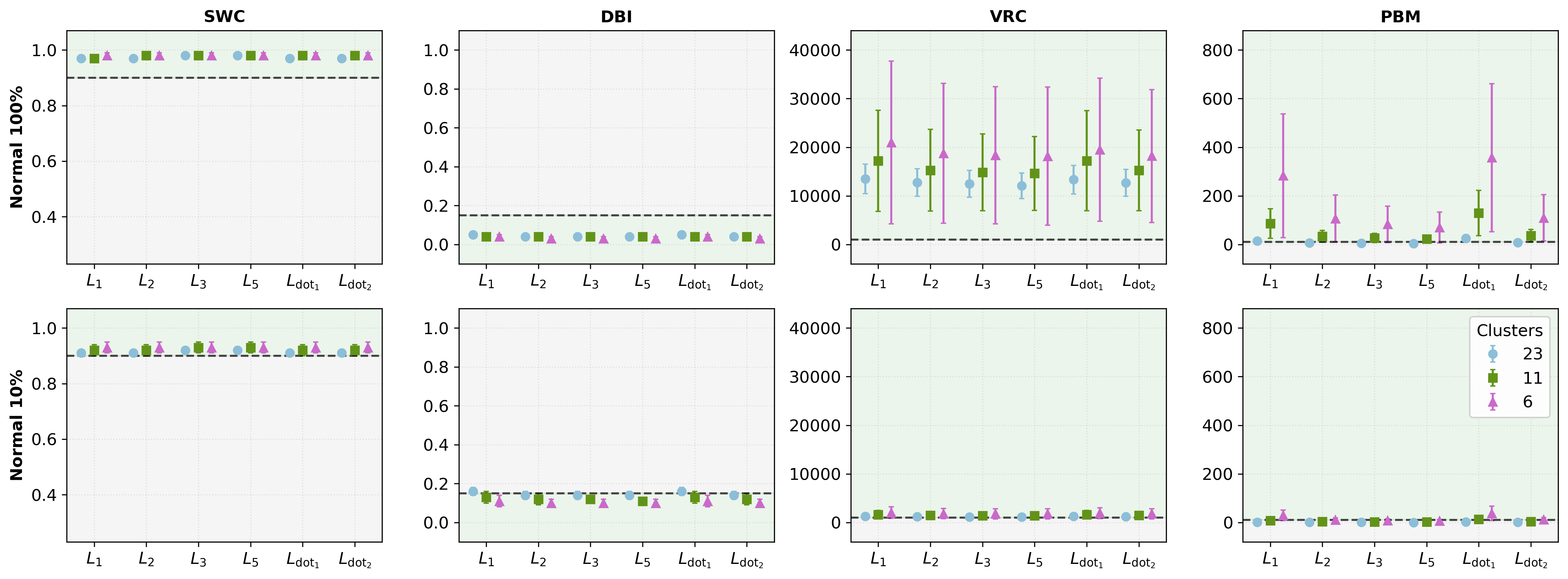}}
    \caption{Structural validity for ICVIs for the six valid distance functions and varying numbers of clusters. First row shows normal complete data, second row shows sparse data. Dashed lines indicate validity thresholds. Green shading = valid region, grey = invalid. Sparse data variants move results closer to the validity threshold. Error bars show mean and standard deviations for the 30 exploratory subjects.}
    \label{fig:icvi-structural-validity-1-3}
\end{figure*}

Figure~\ref{fig:icvi-structural-validity-1-3} illustrates the mean results of the structural performance evaluations for the ICVIs under different data conditions and for the full and reduced cluster count versions of the dataset (structural evaluations 1 and 3). The figure illustrates that SWC and DBI consistently achieve optimal scores for ground truth clustering with little variation between the 30 subjects. Both indices struggle more with the sparse data variant, but are robust to distribution changes. VRC and PBM, on the other hand, vary strongly across the different subjects becoming larger for the non-normal data conditions, while also struggling more with the sparse data variant.

\begin{figure*}[ht]
    \centerline{\includegraphics[width=\textwidth]{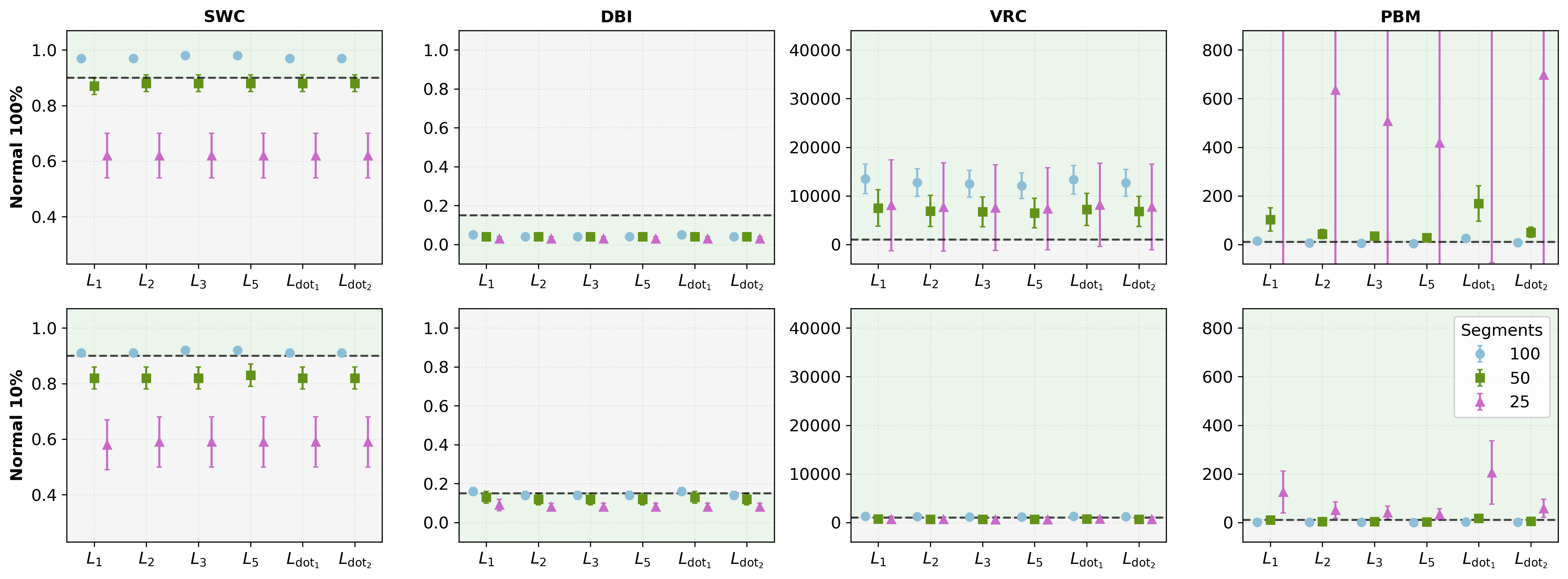}}
    \caption{Structural validity for ICVIs for the six valid distance functions and varying numbers of segments. First row shows normal complete data, second row shows sparse data. SWC, VRC and PBM fail evaluation 4. Dashed lines indicate validity thresholds. Green shading = valid region, grey = invalid. Error bars show mean and standard deviations for the 30 exploratory subjects.}
    \label{fig:icvi-structural-validity-1-4}
\end{figure*}

Figure~\ref{fig:icvi-structural-validity-1-4} illustrates the mean results of the structural performance evaluations for the ICVIs under different data conditions and for the full and reduced segment count versions of the dataset (structural evaluations 1 and 4). In addition to the findings for the varying number of clusters, SWC struggles with segment reduction, while DBI is more robust to fewer objects per cluster. VRC's and PBM's standard deviation spans a large range making it difficult to provide guidance as to what values should be expected in practice for high quality clustering. Note that reducing the segment count reduces the number of segments per cluster to about 2.3 for 50 segments and 1.6 for 25 segments (from 4 to 5 for 100 segments per subject), while also completely removing some clusters (for 50 segments 1.13, for 25 segments 6.83) \cite{degen2025csts}. Reducing the cluster count does not have the same effect as the retained clusters keep all 4 to 5 segments. See Appendix Tables~\ref{tab:icvi-mean-sd-normal-100} to~\ref{tab:discriminant-mean-sd-100} for mean and standard deviation results for the various evaluations to establish overall validity.

To establish whether one of the distance functions is optimal to calculate the valid ICVIs, we tested whether the top ranked distance function leads to significantly more optimal ICVI values (Family 2 tests), as well as whether it leads to significantly stronger correlations with the Jaccard index (Family 3 tests). We found that $\dLp[5]$ for both SWC and DBI leads to both more optimal values and a stronger correlation with the Jaccard index, compared to the second ranked $\dLp[3]$. We found that $\dLp[5]$ is the optimal distance function to use to calculate both SWC and DBI in all data conditions. For the Family 2 confirmatory tests, the p-values were 0.0002 or smaller, with a power of 97\% or more for all conditions and both indices. For the Family 3 tests, this held true for SWC. For DBI, only the normal, partial data condition leads to significantly stronger correlation (p=0.004, power 83.4\%) than $\dLp[3]$. For all other conditions, we could not find a significant difference in correlation strength between $\dLp[5]$ and $\dLp[3]$ for DBI. 

We further investigated whether DBI or SWC was the optimal ICVI to be used, established by a stronger correlation with the Jaccard index (our criterion validity test, Family 4). We found that the two indices were not significantly different under all conditions, except for sparse data variants, where DBI was significantly stronger correlated to the Jaccard index (p=0.002, power 97\%).

Next, we investigated whether reducing cluster count from 23 to 11 to 6 significantly impacts ICVI values (Family 5). Reducing the cluster count from 23 to 6 significantly improved SWC and DBI for all data conditions. This effect was stronger for DBI (p<0.0001, power>99.9\%) than for SWC (p ranging from 0.0006 to 0.01, power from 74.9\% to 94.4\%) with stronger effects coming from the sparse data conditions. Finally, we established whether reducing segments from 100 to 50 to 25 significantly impacts ICVI values (Family 6). Reducing the segment count (number of objects per cluster) from 100 to 50 significantly worsened SWC under all data conditions (p<0.0001, power>99.9\%). For DBI a reduction from 100 to 50 segments significantly improved the ICVI only for the normal 100\% data variant (p=0.024, power 63.2\%). However, the reduction from 50 to 25 segments significantly improved DBI for all data conditions (p<0.0001, power >99.9\%).

For Family 2-4 and 5 all preregistered hypotheses passed, for Family 5 six failed confirmation. See Appendix Tables~\ref{tab:preregistrated-family-2} to~\ref{tab:family6_statistical_validation_results} for all preregistered hypotheses and detailed statistical validation results.

\section{Discussion}\label{sec:discussion}
In this section, we summarise our primary findings, compare them with previous comparative ICVI studies, provide practical guidance, and discuss limitations and future directions.

\subsection{Primary Results}
Our validity evaluation identified six valid distance functions and two valid ICVIs for correlation patterns. Multi-aspect assessment distinguished valid from invalid measurements and revealed which aspects each measurement struggles with.

The six valid distance functions are $\dLp[1]$, $\dLp[2]$, $\dLp[3]$, $\dLp[5]$, $\dDotp[1]$, and $\dDotp[2]$. Most data variants showed no significant differences between these functions. For normal complete and partial and non-normal complete variants, $\dLp[3]$ was statistically optimal. For the non normal partial data condition, $\dDotp[2]$ was optimal. $L_p$ based distance functions succeed through direct geometric interpretation of coefficient dissimilarity, aligning naturally with level set structure where patterns differing by fewer coefficients are more similar. Correlation-specific distance functions $\dlogF$ and $\dFoer$ failed construct, criterion, and external validity. These functions struggled with identity preservation (failing to return near-zero distances for empirical-canonical matrix comparisons), failed to recognise that dissimilarity should increase as more coefficients differ, and achieved suboptimal classification accuracy ($\dlogF$ achieved $F_1=0.89$ vs. $L_p$ functions $F_1=1$; $\dFoer$ failed at $F_1=0,37$). Designed for positive definite matrix geometry, these functions struggle with singularities common in canonical patterns, and their design misaligns with coefficient-based semantic similarity.

Distance function choice affects ICVI validity. SWC and DBI achieved validity depending on the distance function used to calculate them. SWC achieved validity with four distance functions ($\dLp[2]$, $\dLp[3]$, $\dLp[5]$, $\dDotp[2]$); DBI only with $\dLp[3]$ and $\dLp[5]$. $\dLp[5]$ was confirmed as optimal for both indices in all data conditions, producing significantly higher ICVI values (see Family 2). $\dLp[5]$ also led to significantly stronger correlations with the Jaccard index (see Family 3) for SWC under all data conditions and for DBI for the normal complete and partial data variants. This is the first empirical demonstration that distance function selection affects not only ICVI values but validity itself.

\begin{figure}[t]
\centerline{\includegraphics[width=\columnwidth]{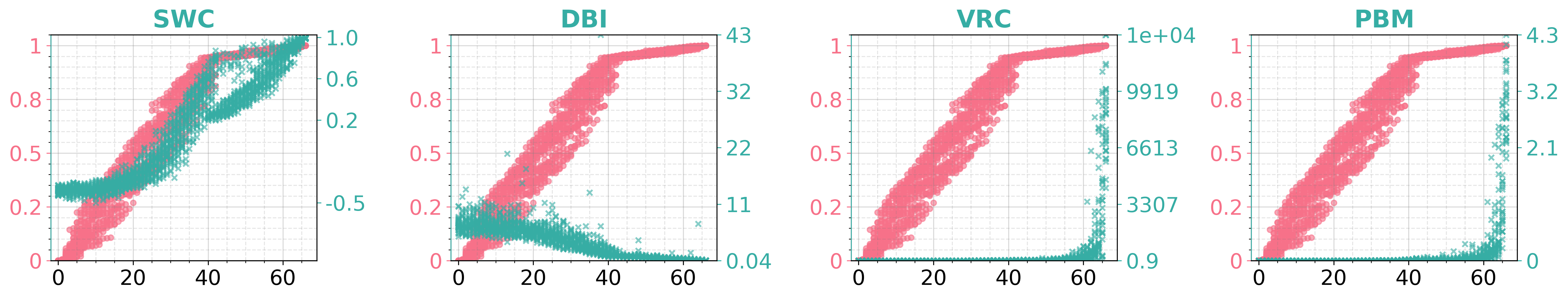}}
\caption{Scatter plots for valid (SWC, DBI) and invalid (VRC, PBM) ICVIs (green, x, right y-axes) and Jaccard index (pink, o, left y-axes) for segmented clusterings of various quality.}
\label{fig:L5_all_measures_nn_70_scatter_plots}
\end{figure}

DBI and SWC performed equivalently under most conditions, but DBI showed significantly stronger Jaccard correlation for sparse variants (Family 4), whilst SWC showed higher sensitivity to segmentation errors than both DBI and Jaccard; see Figure~\ref{fig:L5_all_measures_nn_70_scatter_plots}.

Both indices achieved less optimal values for the sparse data variants, despite MAE not increasing. Both ICVIs showed undesirable sensitivity to cluster counts. Reducing cluster count from 23 to 6 significantly improved both SWC and DBI values (see Family 5 results). Practically, the bias towards fewer clusters is unlikely to affect clustering validation with SWC improving marginally from 0.97 to 0.98 and DBI from 0.04 to 0.03. For segment count reduction (number of objects per cluster), SWC and DBI had opposite effects. SWC worsened significantly to the point of failing structural validity for this test, whilst DBI improved significantly (see Family 6 results). Reducing segment count to 50 or 25 per subject leaves fewer segments per cluster (2-3 vs. 4-5) whilst removing some clusters entirely \cite{degen2025csts}. This indicates that SWC requires more than two objects per cluster to be able to assess within-cluster cohesion.

VRC and PBM failed criterion validity; PBM also failed structural validity for most distance functions. Both showed weak correlations with the Jaccard index (0.21 to 0.34) compared to SWC and DBI (0.89 to 0.92), which is apparent in Figure~\ref{fig:L5_all_measures_nn_70_scatter_plots}. Furthermore, both indices showed high variability in index values between subjects and conditions that prevents establishing reliable quality thresholds, see Figures~\ref{fig:icvi-structural-validity-1-3} and~\ref{fig:icvi-structural-validity-1-4}. VRC and PBM rely on geometric concepts like overall dataset centroid, which translate poorly to correlation space. DBI uses only cluster centroids; SWC relies solely on object-to-object dissimilarity, remaining meaningful across object types. 

\subsection{Comparison with Previous Work}
Previous comprehensive ICVI comparisons established that no single ICVI outperforms others across datasets \cite{Milligan1985, Vendramin2010, Arbelaitz2013, Liu2013, Todeschini2024}. SWC performed strongly in all studies, DBI performed in the top ICVIs in \cite{Arbelaitz2013, Liu2013, Todeschini2024}, VRC in \cite{Vendramin2010, Arbelaitz2013, Todeschini2024} and PBM in \cite{Vendramin2010, Todeschini2024}. Our work introduces psychometric validity theory to establish ICVI validity before comparison. This enforces four methodological requirements  and explains why previous studies observed inconsistent ICVI performance.

First, \textbf{validity theory requires explicit claim definition}. We claim: ICVIs measure clustering quality as cohesion within clusters and separation between clusters. This treats cluster quality as an abstract construct requiring formal operationalisation \cite{Salaudeen2025}. In hindsight, previous studies defined different claims: "finding the correct number of clusters" \cite{Milligan1985, Liu2013, Todeschini2024}, "identifying the best partition" \cite{Vendramin2010, Arbelaitz2013}, or "selecting the optimal similarity paradigm" \cite{Yerbury2024}. Claim definition determines whether a nomological network is required. Previous studies treated their claim objects (correct cluster number, best partition, optimal similarity paradigm) as directly measurable criteria via human judgement or external indices rather than abstract constructs, thus not recognising the need for nomological networks.

Second, \textbf{treating clustering quality as an abstract construct requires a nomological network}. Formalising a nomological network requires defining theoretical optimal organisation which depends on structure type. We define a nomological network built on canonical correlation patterns as maximally distinct patterns, establishing theoretical predictions that clusterings separating these patterns represent optimal quality in the correlation space. Since measuring cohesion and separation relies on dissimilarity (itself an abstract construct), we define level sets as a nomological network to operationalise dissimilarity in correlation space. This formalisation explicitly maps theoretical relationships between clustering quality and dissimilarity to observable indicators (ICVIs and distance functions). Treating dissimilarity as an abstract construct makes selecting similarity paradigms using ICVIs inappropriate, supporting \cite{Yerbury2024}.

Third, \textbf{measurement instruments must operationalise the nomological network}. Here, the dataset is the instrument \cite{Salaudeen2025}.The CSTS benchmark dataset models canonical correlation patterns as our nomological network foundation, providing controlled quality degradation through segmentation and assignment errors for evaluation against known quality boundaries \cite{degen2025csts}. Perturbations (distribution shifts, sparsification, downsampling) enable external validity evaluation by verifying canonical pattern preservation, with optimal performance expected when patterns are preserved. This controlled approach is necessary because clustering algorithms limit assessment to their outputs rather than theoretical quality boundaries. Previous studies used clustering algorithms to generate varying quality clusterings, confounding measure validity with algorithm competency. Whilst previous studies examined factors like overlap, noise, shape, and density \cite{Vendramin2010, Arbelaitz2013, Liu2013, Todeschini2024}, they did not formalise what mathematical organisation constitutes optimal clustering for those structure types nor specify how their datasets operationalised this structure. Without formalising structure type through a nomological network, validation risks optimising measures to specific datasets and human labels rather than establishing whether measures capture the theoretical construct. Notably, \citet{Rousseeuw1987} constructed theoretical examples demonstrating optimal cohesion and separation when proposing SWC, potentially explaining its consistent strong performance. \citet{Liu2013} found SWC performs poorly on arbitrary shapes, but this is expected since arbitrary shapes do not maximise separation, making poor performance a structure mismatch rather than a validity failure.

Fourth, \textbf{multiple validity aspects must be evaluated and require meeting explicit thresholds}. We evaluated content, criterion, construct, and external validity with thresholds from our nomological framework. This examined whether ICVIs capture all construct aspects (content), align with external standards (criterion) and theoretical predictions (structural construct), converge across measures (convergent), fail when structure is absent (discriminant), and generalise across conditions (external). Previous studies assessed some validity aspects. \citet{Vendramin2010} and \citet{Arbelaitz2013} examined criterion validity through external index correlation; they, \citet{Liu2013}, and \citet{Todeschini2024} examined robustness to noise and shape variations without demonstrating dataset operationalisation of nomological networks. \citet{Yerbury2024} evaluated whether ICVIs can be used to select similarity paradigms and suggested external measures are more appropriate. Validity theory requires independent validation of similarity paradigms against nomological networks operationalised through datasets. Evaluation of a similarity paradigm with an external measure establishes criterion validity, not the other aspects of validity. No previous study established validity across all aspects, optimising alignment with labelled datasets rather than theoretical understanding of optimal clustering.

Applying validity theory reveals that ICVI validity depends on structure type enabling nomological network formalisation. For correlation patterns, only SWC and DBI achieved overall validity. VRC and PBM failed criterion validity, contrasting with previous findings in which they performed well for other structures and claims \cite{Vendramin2010, Arbelaitz2013, Todeschini2024}. ICVI validity also depends on distance function selection. Establishing distance function validity separately showed six $L_p$-based distance functions achieved validity with $\dLp[3]$ and $\dDotp[2]$ optimal, whilst correlation-specific distance functions designed for positive definite matrix geometry failed structural, criterion, and external validity. SWC achieved validity with four distance functions, whilst DBI achieved validity with only two, for both $\dLp[5]$ was optimal. 

These differences demonstrate that structure-type-specific formalisation is necessary for validity assessment, explaining the longstanding observation that no single ICVI consistently outperforms others across all datasets. Different labelled datasets represent different structure types or varying quality levels within the same structure type. Different structure types require different valid ICVIs and distance functions. Our structure-type-specific validation provides a template for establishing clustering measure validity, advancing from comparative ranking to rigorous validity assessment.

\subsection{Practical Considerations and Recommendations}
In this section, we translate our findings into practical guidelines for correlation-based clustering validation. Having formally defined and operationalised optimal clustering for correlation patterns, we provide SWC and DBI quality thresholds for assessing discovered pattern strength.

Core recommendations:
\begin{enumerate}
    \item \textbf{Use SWC and DBI together} as complementary ICVIs, avoid VRC and PBM
    \item \textbf{Use quality thresholds} SWC > 0.9 and DBI < 0.15 to indicate patterns as distinct as canonical patterns
    \item \textbf{Use $\dLp[3]$} to map empirical correlation matrices to canonical patterns
    \item \textbf{Use $\dLp[5]$} to calculate SWC and DBI
    \item \textbf{Avoid downsampling}; use highest frequency sampling, including irregularly sampled data
\end{enumerate}

ICVIs of 0.6 < SWC < 0.9 and 0.15 < DBI < 0.4 indicate moderate quality, with patterns weaker than canonical patterns. This may reflect segmentation or clustering errors, or more moderate patterns like downsampled CSTS variants \cite{degen2025csts}. ICVIs of SWC < 0.6 and DBI > 0.4 represent poor results; Table~\ref{tab:thresholds-for-SWC-DBI} shows means for canonical patterns in the different data conditions.

For interpretation of high-quality results, map each cluster's pattern to the closest of 23 canonical patterns by calculating empirical correlation across all cluster segments (Eq. \ref{eq:cluster-controid}). Canonical patterns provide a finite interpretable set that represents conceptual correlation regimes regardless of exact empirical correlations. For example, pattern $[0,0,0]$ indicates times when all three variates have negligible correlations, while pattern $[1,0,0]$ indicates a strong positive correlation between the first two variates but negligible correlations with the third. Experts can assess alignment with theoretical expectations and investigate unexpected patterns for knowledge gaps or novel insights.

\begin{table}[!ht]
\caption{Mean SWC and DBI per data condition. Raw data contain no patterns; downsampled data contain moderate patterns. ICVIs use $\dLp[5]$.}
\label{tab:thresholds-for-SWC-DBI}
\small
\setlength{\tabcolsep}{4pt}
\begin{tabular*}{\columnwidth}{@{\extracolsep{\fill}}l c c c c c}
\toprule 
& \multirow{2}{*}{\textbf{ICVI}} & \multirow{2}{*}{\textbf{Raw}} & \multirow{2}{*}{\textbf{Normal}} & \multirow{2}{*}{\textbf{Non-normal}} & \textbf{Down-}\\ 
& & & & & \textbf{sampled}\\ 
\midrule 
\multirow{2}{*}{\textbf{Complete} (100\%)} & SWC & -0.37 & 0.98 & 0.97 & 0.64\\
& DBI & 16.57 & 0.05 &  0.05 & 0.5 \\ [0.3em]
\multirow{2}{*}{\textbf{Partial} (70\%)} & SWC & -0.37 & 0.97 & 0.97 & 0.63\\
& DBI & 17 & 0.05 & 0.05 & 0.49\\ [0.3em]
\multirow{2}{*}{\textbf{Sparse} (10\%)} & SWC & -0.36 & 0.92  & 0.92 & 0.67\\
& DBI & 15.56 & 0.14 & 0.13 & 0.44\\
\bottomrule 
\end{tabular*}
\end{table}

The recommended methods scale well on standard hardware. Our experiments ran on a MacBook Pro M1. DBI ($O(V(M+K^2))$, where $V$ is the number of time series variates, $M$ is the number of segments, and $K$ is the number of clusters \cite{Vendramin2010}) has lower computational complexity than SWC ($O(VM^2)$, \cite{Vendramin2010}). For large datasets or when efficiency is critical, assess quality first with DBI, then verify promising results with SWC. 

\subsection{Limitations and Future Directions}
Our study has limitations that we would like to acknowledge to contextualise our results and guide future research directions.

Establishing ICVI validity requires formalising the structure type to mathematically define optimal clustering quality independent of domain interpretation. Whether other structure types permit such formalisation remains an open question. Structure types without bounded parameter spaces or clear geometric foundations may require alternative approaches to defining optimal clustering.

The CSTS benchmark dataset models specific perturbations that include specific distributions and sparsity levels \cite{degen2025csts}. Many perturbation types (e.g., temporal dependencies such as autocorrelation and trends) remain uninvestigated. We have not quantified threshold points for maximum sparsity or distribution parameters beyond which validity fails. Our findings are limited to the conditions tested. Defining such thresholds would help practitioners assess whether their data quality is sufficient for reliable correlation-based clustering validation.

Our empirical evaluation uses three variates ($\nVariates=3$), resulting in 27 possible patterns of which 23 satisfy positive semi-definiteness for correlation matrices. While the framework generalises mathematically to $\nVariates>3$, the number of canonical patterns grows exponentially as $3^{{\nVariates(\nVariates-1)}/{2}}$, making $\nVariates=4$ (729 possible patterns) computationally intractable for visualisation, interpretation, and memory requirements. Until tractable approaches for $\nVariates>3$ exist, practitioners should analyse correlation patterns using systematically varied triplets.

Our study examines four indices and fifteen distance functions. Whilst sufficient to demonstrate that correlation patterns differ from euclidean point clouds, we cannot provide guidance for untested combinations. However, we have made the code and dataset available to allow practitioners to test further combinations as required.

\section{Conclusion}\label{sec:conclusions}
Our work demonstrates that ICVI validity is a property of structure type, not of individual datasets, because validity must be assessed against theoretical optimal clustering that real-world data cannot guarantee to manifest. This finding explains four decades of research that no single ICVI consistently outperforms others across all datasets. Datasets contain different, even multiple, structure types with varying organisational quality. Each structure type requires a distinct mathematical definition of optimal clustering quality that serves as the validity benchmark. To determine clustering quality (high within-cluster cohesion and between-cluster separation), we must first validate whether an ICVI and the distance functions it relies on are valid for the particular structure type we are seeking to discover.

Establishing validity requires mathematically defining optimal clustering quality for the structure type. This formalisation serves as the nomological network that operationalises theoretical similarity relationships required for assessing cohesion and separation. We demonstrated this method for correlation patterns, a structure type that has received no attention regarding ICVI performance. Our canonical correlation patterns represent maximally distinct patterns, and hence theoretical optimal clustering, whilst level sets characterise expected similarity between correlation patterns. Formalisation shifts clustering quality from a property of the dataset to a property of the mathematical organisation, enabling domain-independent validity assessment against theoretical optima.

Validity assessment revealed that for correlation patterns, SWC and DBI achieve validity, whilst VRC and PBM fail criterion validity. These results differ from previous studies where VRC and PBM performed well on Euclidean point cloud-like datasets, empirically confirming that validity depends on structure type. Furthermore, distance function selection directly affects ICVI validity, not merely index values. Simple $\dLp[]$ distance functions achieved validity for correlation patterns, whilst correlation-specific distance functions designed for positive definite matrix geometry failed structural, criterion, and external validity.

Structure-type-specific validation using synthetic datasets that operationalise theoretical constructs enables practical advances previously unattainable. Valid ICVIs provide domain-independent objective measures of organisation strength. Quality thresholds SWC > 0.9 and DBI < 0.15 indicate discovered patterns achieve canonical pattern strength (maximum structural contrast in correlation space). Practitioners can use valid ICVIs to assess whether a structure type is present and its quality. ICVI selection becomes principled and based on the structure type to be discovered.

We call on the clustering research community to organise methods by structure type rather than algorithm class. Only through systematic validation across structure types can we build a comprehensive understanding of when and why clustering quality measures succeed or fail, providing principled method selection. Existing indices require validation across structure types to establish their scope of validity. New indices should be motivated by existing indices failing for specific structure types, with formal definition and quantification of clustering quality. This organisation moves the field from "try and see" toward principled selection. Domain-specific interpretation remains necessary to understand what finding or failing to find a specific structure type means for a particular domain.

\section*{Acknowledgments}
We thank UK Research and Innovation (UKRI) which was funding Isabella's PhD research through the UKRI Doctoral Training in Interactive Artificial Intelligence under grant EP/S022937/1 and EPSRC which is funding her EPSRC Doctoral Impact Fellowship. 

We are grateful to everybody involved in the Interactive AI CDT at Bristol University for their support and guidance.

We used Claude Sonnet 4.5 to help with writing and code for graphics as well as a research dialogue tool. Our use adheres to ethical guidelines for use and acknowledgement of generative AI in academic research. Each author has made a substantial contribution to the work, which has been thoroughly vetted for accuracy, and assumes responsibility for the integrity of their contributions.

\bibliographystyle{unsrtnat}  
\bibliography{main} 

@article{Fahad2014,
   author = {Adil Fahad and Najlaa Alshatri and Zahir Tari and Abdullah Alamri and Ibrahim Khalil and Albert Y. Zomaya and Sebti Foufou and Abdelaziz Bouras},
   doi = {10.1109/TETC.2014.2330519},
   issn = {21686750},
   issue = {3},
   journal = {IEEE Transactions on Emerging Topics in Computing},
   month = {9},
   pages = {267-279},
   publisher = {IEEE Computer Society},
   title = {A survey of clustering algorithms for big data: Taxonomy and empirical analysis},
   volume = {2},
   year = {2014},
   url = {https://ieeexplore.ieee.org/document/6832486}
}

@article{Jaeger2023,
   author = {Adam Jaeger and David Banks},
   doi = {10.1002/WICS.1597},
   issn = {1939-0068},
   issue = {3},
   journal = {Wiley Interdisciplinary Reviews: Computational Statistics},
   month = {5},
   pages = {e1597},
   publisher = {John Wiley & Sons, Ltd},
   title = {Cluster analysis: A modern statistical review},
   volume = {15},
   url = {https://onlinelibrary.wiley.com/doi/full/10.1002/wics.1597},
   year = {2023}
}

@article{Marti2021,
   author = {Gautier Marti and Frank Nielsen and Mikołaj Bińkowski and Philippe Donnat},
   doi = {10.1007/978-3-030-65459-7_10},
   isbn = {978-3-030-65459-7},
   issn = {1860-4870},
   journal = {Signals and Communication Technology},
   pages = {245-274},
   publisher = {Springer, Cham},
   title = {A Review of Two Decades of Correlations, Hierarchies, Networks and Clustering in Financial Markets},
   url = {https://link.springer.com/chapter/10.1007/978-3-030-65459-7_10},
   year = {2021}
}

@article{Xu2015,
   author = {Dongkuan Xu and Yingjie Tian},
   doi = {10.1007/S40745-015-0040-1},
   issn = {2198-5812},
   issue = {2},
   journal = {Annals of Data Science 2015 2:2},
   month = {8},
   pages = {165-193},
   publisher = {Springer},
   title = {A Comprehensive Survey of Clustering Algorithms},
   volume = {2},
   url = {https://link.springer.com/article/10.1007/s40745-015-0040-1},
   year = {2015}
}

@article{Ezugwu2022,
   author = {Absalom E. Ezugwu and Abiodun M. Ikotun and Olaide O. Oyelade and Laith Abualigah and Jeffery O. Agushaka and Christopher I. Eke and Andronicus A. Akinyelu},
   doi = {10.1016/J.ENGAPPAI.2022.104743},
   issn = {0952-1976},
   journal = {Engineering Applications of Artificial Intelligence},
   month = {4},
   pages = {104743},
   publisher = {Pergamon},
   title = {A comprehensive survey of clustering algorithms: State-of-the-art machine learning applications, taxonomy, challenges, and future research prospects},
   volume = {110},
   year = {2022},
   url = {https://www.sciencedirect.com/science/article/pii/S095219762200046X}
}

@article{Gao2023,
   doi = {10.1016/J.PSYCHRES.2023.115265},
   author = {Caroline X. Gao and Dominic Dwyer and Ye Zhu and Catherine L. Smith and Lan Du and Kate M. Filia and Johanna Bayer and Jana M. Menssink and Teresa Wang and Christoph Bergmeir and Stephen Wood and Sue M. Cotton},
   issn = {0165-1781},
   journal = {Psychiatry Research},
   month = {9},
   pages = {115265},
   pmid = {37348404},
   publisher = {Elsevier},
   title = {An overview of clustering methods with guidelines for application in mental health research},
   volume = {327},
   year = {2023},
   url = {https://www.sciencedirect.com/science/article/pii/S0165178123002159}
}

@article{Milligan1985,
   author = {Glenn W. Milligan and Martha C. Cooper},
   doi = {10.1007/BF02294245/METRICS},
   issn = {00333123},
   issue = {2},
   journal = {Psychometrika},
   month = {6},
   pages = {159-179},
   publisher = {Springer-Verlag},
   title = {An examination of procedures for determining the number of clusters in a data set},
   volume = {50},
   url = {https://link.springer.com/article/10.1007/BF02294245},
   year = {1985},
}

@article{Kleinberg2002,
   author = {Jon Kleinberg},
   journal = {NeurIPS 02},
   pages = {463–470},
   title = {An Impossibility Theorem for Clustering},
   year = {2002},
}

@misc{Omar2019,
      title={Evaluation Metrics for Unsupervised Learning Algorithms}, 
      author={Julio-Omar Palacio-Niño and Fernando Berzal},
      year={2019},
      eprint={1905.05667},
      archivePrefix={arXiv},
      primaryClass={cs.LG},
      url={https://arxiv.org/abs/1905.05667}, 
}

@article{Vendramin2010,
   author = {Lucas Vendramin and Ricardo J.G.B. Campello and Eduardo R. Hruschka},
   doi = {10.1002/SAM.10080},
   issn = {1932-1872},
   issue = {4},
   journal = {Statistical Analysis and Data Mining: The ASA Data Science Journal},
   month = {8},
   pages = {209-235},
   publisher = {John Wiley & Sons, Ltd},
   title = {Relative clustering validity criteria: A comparative overview},
   volume = {3},
   url = {https://onlinelibrary.wiley.com/doi/full/10.1002/sam.10080},
   year = {2010},
}

@article{Gensler2014,
   author = {André Gensler and B. Sick},
   journal = {LWA},
   title = {Novel Criteria to Measure Performance of Time Series Segmentation Techniques},
   year = {2014},
}

@article{Burg2020,
   author = {Gerrit J J Van Den Burg and Christopher K I Williams},
   journal = {arXiv.org},
   month = {3},
   title = {An Evaluation of Change Point Detection Algorithms},
   url = {https://github.com/alan-turing-institute/TCPD.},
   year = {2020},
}

@article{Liu2013,
   author = {Yanchi Liu and Zhongmou Li and Hui Xiong and Xuedong Gao and Junjie Wu and Sen Wu},
   doi = {10.1109/TSMCB.2012.2220543},
   issn = {2168-2275},
   issue = {3},
   journal = {IEEE transactions on cybernetics},
   month = {6},
   pages = {982-994},
   pmid = {23193245},
   publisher = {IEEE Trans Cybern},
   title = {Understanding and enhancement of internal clustering validation measures},
   volume = {43},
   url = {https://pubmed.ncbi.nlm.nih.gov/23193245/},
   year = {2013},
}

@book{Kaufman1990,
   author = {Leonard Kaufman and Peter J. Rousseeuw},
   doi = {10.1002/9780470316801},
   isbn = {9780471878766},
   month = {3},
   publisher = {Wiley},
   title = {Finding Groups in Data},
   url = {https://onlinelibrary.wiley.com/doi/book/10.1002/9780470316801},
   year = {1990}
}

@article{Todeschini2024,
   author = {Roberto Todeschini and Davide Ballabio and Veronica Termopoli and Viviana Consonni},
   doi = {10.1016/J.CHEMOLAB.2024.105117},
   issn = {0169-7439},
   journal = {Chemometrics and Intelligent Laboratory Systems},
   month = {8},
   pages = {105117},
   publisher = {Elsevier},
   title = {Extended multivariate comparison of 68 cluster validity indices. A review},
   volume = {251},
   year = {2024}
}

@article{Yerbury2024,
   author = {Luke W. Yerbury and Ricardo J. G. B. Campello and G. C. Livingston, Jr. and Mark Goldsworthy and Lachlan O’Neil},
   doi = {10.1145/3748726},
   issn = {1556-4681},
   issue = {8},
   journal = {ACM Transactions on Knowledge Discovery from Data},
   month = {9},
   pages = {1-53},
   publisher = {Association for Computing Machinery (ACM)},
   title = {On the Use of Relative Validity Indices for Comparing Clustering Approaches},
   volume = {19},
   url = {https://dl.acm.org/doi/10.1145/3748726},
   year = {2025}
}

@article{Dau2019,
   doi = {10.1109/JAS.2019.1911747},
   author = {Hoang Anh Dau and Anthony Bagnall and Kaveh Kamgar and Chin-Chia Michael Yeh and Yan Zhu and Shaghayegh Gharghabi and Chotirat Ann Ratanamahatana and Eamonn Keogh and Hoang Anh Dau and Anthony Bagnall and Kaveh Kamgar and Chin-Chia Michael Yeh and Yan Zhu and Shaghayegh Gharghabi and Chotirat Ann Ratanamahatana and Eamonn Keogh},
   issn = {2329-9266},
   issue = {6},
   journal = {IEEE/CAA Journal of Automatica Sinica, 2019, Vol. 6, Issue 6, Pages: 1293-1305},
   month = {11},
   pages = {1293-1305},
   publisher = {IEEE/CAA Journal of Automatica Sinica},
   title = {The UCR Time Series Archive},
   volume = {6},
   url = {https://www.ieee-jas.net/en/article/doi/10.1109/JAS.2019.1911747},
   year = {2019}
}

@article{Arbelaitz2013,
   author = {Olatz Arbelaitz and Ibai Gurrutxaga and Javier Muguerza and Jesús M. Pérez and Iñigo Perona},
   doi = {10.1016/J.PATCOG.2012.07.021},
   issn = {0031-3203},
   issue = {1},
   journal = {Pattern Recognition},
   month = {1},
   pages = {243-256},
   publisher = {Pergamon},
   title = {An extensive comparative study of cluster validity indices},
   url = {https://www.sciencedirect.com/science/article/pii/S003132031200338X},
   volume = {46},
   year = {2013},
}

@article{Rousseeuw1987,
   author = {Peter J. Rousseeuw},
   doi = {10.1016/0377-0427(87)90125-7},
   issn = {0377-0427},
   issue = {C},
   journal = {Journal of Computational and Applied Mathematics},
   month = {11},
   pages = {53-65},
   publisher = {North-Holland},
   title = {Silhouettes: A graphical aid to the interpretation and validation of cluster analysis},
   url = {https://www.sciencedirect.com/science/article/pii/0377042787901257},
   volume = {20},
   year = {1987},
}

@article{Foerstner2003,
   author = {Wolfgang Förstner and Boudewijn Moonen},
   doi = {10.1007/978-3-662-05296-9_31},
   journal = {Geodesy-The Challenge of the 3rd Millennium},
   pages = {299-309},
   publisher = {Springer, Berlin, Heidelberg},
   title = {A Metric for Covariance Matrices},
   url = {https://link.springer.com/chapter/10.1007/978-3-662-05296-9_31},
   year = {2003},
}

@article{Ergezer2018,
   author = {Hamza Ergezer and Kemal Leblebicioğlu},
   doi = {10.1007/S10115-017-1098-1},
   issn = {02193116},
   issue = {3},
   journal = {Knowledge and Information Systems},
   month = {6},
   pages = {695-718},
   publisher = {Springer London},
   title = {Time series classification with feature covariance matrices},
   volume = {55},
   url = {https://link.springer.com/article/10.1007/s10115-017-1098-1},
   year = {2018},
}

@article{Looney2020,
   author = {Justine O. May and Stephen W. Looney},
   doi = {10.37421/2155-6180.2020.11.440},
   issn = {2155-6180},
   issue = {6},
   journal = {Journal of Biometrics \& Biostatistics},
   month = {4},
   pages = {1-7},
   publisher = {Hilaris SRL},
   title = {Sample Size Charts for Spearman and Kendall Coefficients},
   volume = {11},
   year = {2020},
}

@article{Davies1979,
   author = {David L. Davies and Donald W. Bouldin},
   doi = {10.1109/TPAMI.1979.4766909},
   issn = {01628828},
   issue = {2},
   journal = {IEEE Transactions on Pattern Analysis and Machine Intelligence},
   pages = {224-227},
   pmid = {21868852},
   title = {A Cluster Separation Measure},
   volume = {PAMI-1},
   url = {https://ieeexplore.ieee.org/document/4766909},
   year = {1979},
}

@article{Calinski1974,
   author = {T. Caliñski and J. Harabasz},
   doi = {10.1080/03610927408827101},
   issn = {00903272},
   issue = {1},
   journal = {Communications in Statistics},
   pages = {1-27},
   title = {A Dendrite Method Foe Cluster Analysis},
   volume = {3},
   year = {1974},
}

@article{Pakhira2004,
   author = {Malay K. Pakhira and Sanghamitra Bandyopadhyay and Ujjwal Maulik},
   doi = {10.1016/J.PATCOG.2003.06.005},
   issn = {0031-3203},
   issue = {3},
   journal = {Pattern Recognition},
   month = {3},
   pages = {487-501},
   publisher = {Pergamon},
   title = {Validity index for crisp and fuzzy clusters},
   volume = {37},
   year = {2004},
}

@article{Iglesias2013,
   author = {Félix Iglesias and Wolfgang Kastner},
   doi = {10.3390/EN6020579},
   isbn = {4315880118391},
   issn = {1996-1073},
   issue = {2},
   journal = {Energies 2013, Vol. 6, Pages 579-597},
   month = {1},
   pages = {579-597},
   publisher = {Multidisciplinary Digital Publishing Institute},
   title = {Analysis of Similarity Measures in Times Series Clustering for the Discovery of Building Energy Patterns},
   volume = {6},
   url = {https://www.mdpi.com/1996-1073/6/2/579/htm},
   year = {2013}
}

@article{Chandereng2020,
   author = {Thevaa Chandereng and Anthony Gitter},
   doi = {10.1186/S12859-019-3324-1},
   issn = {14712105},
   issue = {1},
   journal = {BMC Bioinformatics},
   month = {1},
   pmid = {31948388},
   publisher = {BioMed Central},
   title = {Lag penalized weighted correlation for time series clustering},
   volume = {21},
   url = {https://www.ncbi.nlm.nih.gov/pmc/articles/PMC6966853/},
   year = {2020}
}

@inproceedings{Guyon2009,
  title={Clustering: Science or Art?},
  author={Guyon, Isabelle and von Luxburg, Ulrike and Williamson, Robert C.},
  booktitle={NIPS 2009 Workshop on Clustering: Science or art? Towards principled approaches},
  year={2009},
  month={December},
  address={Vancouver, Canada},
  url={https://stanford.edu/~rezab/nips2009workshop/opinions/opinion-artorscience.pdf},
  note={Position paper}
}

@misc{degen2025csts,
      title={CSTS: A Benchmark for the Discovery of Correlation Structures in Time Series Clustering}, 
      author={Isabella Degen and Zahraa S Abdallah and Henry W J Reeve and Kate Robson Brown},
      year={2025},
      eprint={2505.14596},
      archivePrefix={arXiv},
      primaryClass={cs.LG},
      url={https://arxiv.org/abs/2505.14596}, 
}

@article{Nosek2018,
   author = {Brian A. Nosek and Charles R. Ebersole and Alexander C. DeHaven and David T. Mellor},
   doi = {10.1073/PNAS.1708274114},
   issn = {10916490},
   issue = {11},
   journal = {Proceedings of the National Academy of Sciences},
   month = {3},
   pages = {2600-2606},
   pmid = {29531091},
   publisher = {National Academy of Sciences},
   title = {The preregistration revolution},
   volume = {115},
   url = {https://www.pnas.org/doi/abs/10.1073/pnas.1708274114},
   year = {2018}
}

@article{Shannon1948,
   author = {C. E. Shannon},
   doi = {10.1002/J.1538-7305.1948.TB01338.X},
   issn = {15387305},
   issue = {3},
   journal = {Bell System Technical Journal},
   pages = {379-423},
   title = {A Mathematical Theory of Communication},
   volume = {27},
   year = {1948},
   url = {https://ieeexplore.ieee.org/document/6773024},
}

@article{Cover1967NN,
   author = {T. M. Cover and P. E. Hart},
   doi = {10.1109/TIT.1967.1053964},
   issn = {15579654},
   issue = {1},
   journal = {IEEE Transactions on Information Theory},
   pages = {21-27},
   title = {Nearest Neighbor Pattern Classification},
   volume = {13},
   year = {1967},
   url = {https://ieeexplore.ieee.org/document/1053964}
}

@article{Grecki20241NN,
   author = {Tomasz Górecki and Maciej Łuczak and Paweł Piasecki},
   doi = {10.1016/J.JOCS.2024.102235},
   issn = {1877-7503},
   journal = {Journal of Computational Science},
   month = {3},
   pages = {102235},
   publisher = {Elsevier},
   title = {An exhaustive comparison of distance measures in the classification of time series with 1NN method},
   volume = {76},
   year = {2024},
   url = {https://www.sciencedirect.com/science/article/pii/S1877750324000280}
}

@article{Christen2023-F1,
   author = {Peter Christen and David J. Hand and Nishadi Kirielle},
   doi = {10.1145/3606367},
   issn = {15577341},
   issue = {3},
   journal = {ACM Computing Surveys},
   month = {3},
   publisher = {Association for Computing Machinery},
   title = {A Review of the F-Measure: Its History, Properties, Criticism, and Alternatives},
   volume = {56},
   url = {https://dl.acm.org/doi/10.1145/3606367},
   year = {2023}
}

@article{Wilcoxon1945,
   author = {Frank Wilcoxon},
   doi = {10.2307/3001968},
   issn = {00994987},
   issue = {6},
   journal = {Biometrics Bulletin},
   month = {12},
   pages = {80},
   publisher = {JSTOR},
   title = {Individual Comparisons by Ranking Methods},
   volume = {1},
   year = {1945},
   url = {https://www.jstor.org/stable/3001968}
}

@article{Puth2015,
   author = {Marie Therese Puth and Markus Neuhäuser and Graeme D. Ruxton},
   doi = {10.1016/J.ANBEHAV.2015.01.010},
   issn = {0003-3472},
   journal = {Animal Behaviour},
   month = {4},
   pages = {77-84},
   publisher = {Academic Press},
   title = {Effective use of Spearman's and Kendall's correlation coefficients for association between two measured traits},
   volume = {102},
   year = {2015}
}

@misc{uci_ml_repo,
    author = {Markelle Kelly and Rachel Longjohn and Kolby Nottingham},
    title = {The UCI Machine Learning Repository},
    url = {https://archive.ics.uci.edu},
    year = {2013},
    note = {University of California, Irvine, School of Information and Computer Sciences}
}

@misc{Salaudeen2025,
  author = {Olawale Salaudeen and Anka Reuel and Ahmed Ahmed and Suhana Bedi and Zachary Robertson and Sudharsan Sundar and Ben Domingue and Angelina Wang and Sanmi Koyejo},
  title = {Measurement to Meaning: A Validity-Centered Framework for AI Evaluation},
  year = {2025},
  month = may,
  eprint = {2505.10573},
  archivePrefix = {arXiv},
  primaryClass = {cs.LG},
    note = {Presented at ICLR 2025 Workshop on Human-AI Coevolution and NeurIPS 2025 LLM Evaluation Workshop}
}

@article{Cronbach1955,
   author = {Lee J. Cronbach and Paul E. Meehl},
   doi = {10.1037/H0040957},
   issn = {00332909},
   issue = {4},
   journal = {Psychological Bulletin},
   month = {7},
   pages = {281-302},
   pmid = {13245896},
   title = {Construct validity in psychological tests},
   volume = {52},
   year = {1955}
}

@article{Hassan2024,
   author = {Bryar A. Hassan and Noor Bahjat Tayfor and Alla A. Hassan and Aram M. Ahmed and Tarik A. Rashid and Naz N. Abdalla},
   doi = {10.1016/J.NEUCOM.2024.128198},
   issn = {0925-2312},
   journal = {Neurocomputing},
   month = {10},
   pages = {128198},
   publisher = {Elsevier},
   title = {From A-to-Z review of clustering validation indices},
   volume = {601},
   year = {2024}
}

@article{Gagolewski2021,
   author = {Marek Gagolewski and Maciej Bartoszuk and Anna Cena},
   doi = {10.1016/J.INS.2021.10.004},
   issn = {0020-0255},
   journal = {Information Sciences},
   month = {12},
   pages = {620-636},
   publisher = {Elsevier},
   title = {Are cluster validity measures (in) valid?},
   volume = {581},
   year = {2021}
}

@article{Campbell1959,
   author = {Donald T. Campbell and Donald W. Fiske},
   doi = {10.1037/H0046016},
   issn = {00332909},
   issue = {2},
   journal = {Psychological bulletin},
   month = {3},
   pages = {81-105},
   pmid = {13634291},
   title = {Convergent and discriminant validation by the multitrait-multimethod matrix.},
   volume = {56},
   year = {1959}
}

@book{ValidityStandards2014,
  author    = {{American Educational Research Association} and {American Psychological Association} and {National Council on Measurement in Education}},
  title     = {Standards for Educational and Psychological Testing},
  year      = {2014},
  publisher = {American Educational Research Association},
  address   = {Washington, DC},
  isbn      = {978-0-935302-35-6}
}

@article{Messick1994,
   author = {Samuel Messick},
   doi = {10.1002/J.2333-8504.1994.TB01618.X},
   issn = {2330-8516},
   issue = {2},
   journal = {ETS Research Report Series},
   month = {12},
   publisher = {Wiley},
   title = {Validity of Psychological Assessment: Validation of Inferences from Persons' Responses and Performances as Scientific Inquiry into Score Meaning. Research Report RR-94-45.},
   volume = {1994},
   year = {1994}
}
\clearpage
\appendix
\numberwithin{equation}{section}
\numberwithin{table}{section}
\numberwithin{figure}{section}
\section{The Canonical Patterns}\label{app:canonical-patterns}
\vspace{1em}

\begin{table}[!ht]  
\small  
\caption{Overview of all canonical patterns and their relaxed versions. Indicating which patterns can be modelled exactly as they are and which need adjustment.}
\label{tab:A1}
\begin{tabular*}{\columnwidth}{@{\extracolsep{\fill}}l cccc}
\toprule
\textbf{Id} & 
\textbf{Canonical Pattern} & 
\textbf{Relaxed Pattern} & 
\textbf{Ideal} & 
\textbf{Modelled} \\
\midrule
0 & (0, 0, 0) & (0, 0, 0) & Yes & Yes \\
1 & (0, 0, 1) & (0, 0, 1) & Yes & Yes \\
2 & (0, 0, -1) & (0, 0, -1) & Yes & Yes \\
3 & (0, 1, 0) & (0, 1, 0) & Yes & Yes \\
4 & (0, 1, 1) & (0.0, 0.71, 0.7) & No & Yes \\
5 & (0, 1, -1) & (0, 0.71, -0.7) & No & Yes \\
6 & (0, -1, 0) & (0, -1, 0) & Yes & Yes \\
7 & (0, -1, 1) & (0, -0.71, 0.7) & No & Yes \\
8 & (0, -1, -1) & (0, -0.71, -0.7) & No & Yes \\
9 & (1, 0, 0) & (1, 0, 0) & Yes & Yes \\
10 & (1, 0, 1) & (0.71, 0, 0.7) & No & Yes \\
11 & (1, 0, -1) & (0.71, 0, -0.7) & No & Yes \\
12 & (1, 1, 0) & (0.71, 0.7, 0) & No & Yes \\
13 & (1, 1, 1) & (1, 1, 1) & Yes & Yes \\
\midrule
14 & (1, 1, -1) & - & No & No \\
\midrule
15 & (1, -1, 0) & (0.71, -0.7, 0) & No & Yes \\
\midrule
16 & (1, -1, 1) & - & No & No \\
\midrule
17 & (1, -1, -1) & (1, -1, -1) & Yes & Yes \\
18 & (-1, 0, 0) & (-1, 0, 0) & Yes & Yes \\
19 & (-1, 0, 1) & (-0.71, 0, 0.7) & No & Yes \\
20 & (-1, 0, -1) & (-0.71, 0, -0.7) & No & Yes \\
21 & (-1, 1, 0) & (-0.71, 0.7, 0) & No & Yes \\
\midrule
22 & (-1, 1, 1) & - & No & No \\
\midrule
23 & (-1, 1, -1) & (-1, 1, -1) & Yes & Yes \\
24 & (-1, -1, 0) & (-0.71, -0.7, 0) & No & Yes \\
25 & (-1, -1, 1) & (-1, -1, 1) & Yes & Yes \\
\midrule
26 & (-1, -1, -1) & - & No & No \\
\bottomrule
\end{tabular*}
\end{table} 

\FloatBarrier

\section{Distance Functions}\label{app:distance-functions}
Here we give the definitions for the distance functions used in this paper. In our case a distance function $d: \subsetOfSymmetricUnitDiagonalMatrices \times \subsetOfSymmetricUnitDiagonalMatrices \to \R_{\geq 0}$ maps two matrices to a real number to quantify how dissimilar empirical correlation matrices $\correlationMatrix$ are from relaxed canonical patterns $\relaxedPattern$ or how dissimilar two correlation matrices in general are.

\subsection{Lp Norm}
The Lp norm distance (also called Minkowski) between any two correlation matrices is defined as  
\begin{equation}
        \distanceMeasure[L_p] := \left(\sum_{i=1}^{\nVariates}\sum_{j>i}^{\nVariates}(\correlationMatrixElement_{ij}-\canonicalPatternElement_{ij})^p\right)^\frac{1}{p}
        \label{eq:lp_norm}.
\end{equation}
Note that we only use the upper half of the correlation matrices, making this a vector distance.

\subsection{Lp Norm with reference vector}
Let $d$ be an Lp norm which by definition is rotation insensitive resulting in $d_{L_p}(\correlationMatrix[x],\relaxedPattern)=d_{L_p}(\correlationMatrix[y],\relaxedPattern)$ if the vectors of the upper half of the correlation matrix $\mathbf{a}^U,\mathbf{p}^U$ in $\mathbb{R}^{\nVariatePairs}$ are equidistant from a hyperplane $H$ and lie on opposite sides of it, with the line connecting them being perpendicular to $H$ and $|\mathbf{a}^U|=|\mathbf{p}^U|$ with $|\cdot|$ being a vector norm. This might impede similarity differentiation between certain correlation matrices. We attempt to address this by introducing a reference vector $\mathbf{v}_\text{ref} = \frac{(1, 1, ..., 1)^\nVariatePairs}{\sqrt{\nVariatePairs}}$.
The Lp Norm with reference vector is then defined as
\begin{equation}
        \distanceMeasure[\text{ref}_p] := d_{L_p}(\correlationMatrix, \relaxedPattern)(d_{L_p}(\mathbf{a}^U,\mathbf{v}_\text{ref})+d_{L_p}(\mathbf{p}^U,\mathbf{v}_\text{ref}))
        \label{eq:lp_norm_with_ref}
\end{equation}

\subsection{Lp Norm with dot transformation}
This addresses the same rotation insensitivity problem described in the aforementioned Lp Norm with reference vector approach. Here, instead of modifying the distance calculation, we augment the correlation vectors with orientation information. 

Let $\mathbf{a}^U \in \mathbb{R}^{\nVariatePairs}$ be the upper triangular vector of a correlation matrix. We define an augmented vector $\tilde{\mathbf{a}} \in \mathbb{R}^{\nVariatePairs+1}$ as:

\begin{equation}
\tilde{\mathbf{a}} := \begin{pmatrix} \mathbf{a}^U \\ s \end{pmatrix}
\end{equation}

where the orientation scalar $s := \frac{\mathbf{a}^U \cdot \mathbf{v}_\text{ref} + 1}{2}$ with reference vector $\mathbf{v}_\text{ref} := \frac{(1, 1, ..., 1)^T}{\sqrt{\nVariatePairs}}$. The scalar $s \in [0,1]$ encodes the alignment between the correlation vector and the reference vector, preventing the mirror vectors from having identical distances.

The Lp norm with dot transformation is then defined as

\begin{equation}
        \distanceMeasure[\text{dot}_p] := d_{L_p}(\tilde{\mathbf{a}}, \tilde{\mathbf{p}}),
        \label{eq:lp_norm_dot_transform}
\end{equation}

where $\tilde{\mathbf{p}} \in \mathbb{R}^{\nVariatePairs+1}$ is the similarly augmented upper triangular vector of the correlation matrix of the relaxed canonical pattern $\relaxedPattern$.

\subsection{Förstner Distance}\label{sec:foerstner-distance}
The Förstner metric is defined as \cite{Foerstner2003}
\begin{equation}
   \distanceMeasure[F]:= \sqrt{\sum_{i=1}^{\nVariates} \ln^2 \lambda_i(\correlationMatrix, \relaxedPattern)},
    \label{eq:unstable-foerstner-distance}
\end{equation}

where $\lambda_i$ are the $\nVariates$ generalised eigenvalues of $\correlationMatrix x=\lambda\relaxedPattern[y]x$ solving $det(\correlationMatrix - \lambda\relaxedPattern) = 0$. The Förstner metric (also known as the affine-invariant Riemannian metric) is specifically designed to work as a geodesic distance on the manifold of positive definite matrices, treating the space of correlation/covariance matrices as a Riemannian manifold.

For the relaxed canonical patterns $\relaxedPattern$ this distance function leads to frequent nan and inf results due to the valid singular nature of many of the relaxed canonical patterns:
\begin{enumerate}
    \item $\lambda_i=0$: when $\correlationMatrix$ is singular
    \item $\lambda_i=\infty$ when $\relaxedPattern$ is singular 
    \item $\lambda_i<0$ when both correlation matrices $\correlationMatrix, \relaxedPattern$ are singular but due to numerical imprecision in the eigenvalue calculation, the generalised eigenvalues can become negative instead of $\infty$ (see Appendix \ref{sec:app1}) 
\end{enumerate}

We fix issue 1. by calculating $ln^2 (\lambda_i+1)$ instead and issues 2. and 3. by adding a regularisation term $\epsilon\mathbf{I}$ where $\epsilon=1e-10$ to all correlation matrices. The numerically stable version of the Förstner metric is defined as
\begin{equation}
    \distanceMeasure[F,\epsilon]:= \sqrt{\sum_{i=1}^{\nVariates} \ln^2 (\lambda_i(\correlationMatrix+\epsilon\mathbf{I}, \relaxedPattern+\epsilon\mathbf{I})+1)}.
    \label{eq:stable-foerstner-distance}
\end{equation}

\subsection{Log Frobenius Distance}
The Log Frobenius distance is defined as \cite{Ergezer2018}
\begin{equation}
        \distanceMeasure[\log F] := \|\ln(\correlationMatrix)-\ln(\relaxedPattern)\|_F,
        \label{eq:unstable-log-cov-frobenius-distance}
\end{equation}
where $F$ is the Frobenius norm. The Log Frobenius distance maps points on the Riemannian manifold of positive definite matrices to the tangent space (ln) and then measures the Euclidean distance in that tangent space, resulting in simpler computation than the Förstner distance that respects the curved geometry. To avoid issues with singular correlation matrices $\correlationMatrix, \relaxedPattern$ that lead to the determinate being $0$ making $ln$ undefined, we add a small regularisation term $\epsilon\mathbf{I}$ where $\epsilon=1e-10$.
The stable version of the Log Frobenius distance is defined as
\begin{equation}\label{eq:log-cov-frobenius-distance}
        \distanceMeasure[\log F, \epsilon] := \|\ln(\correlationMatrix + \epsilon\mathbf{I})-\ln(\relaxedPattern + \epsilon\mathbf{I})\|_F.
\end{equation}

\FloatBarrier

\section{Marcro F1 Score}\label{app:macro-f1-score}
This section provides the detailed mathematical formulation of the macro F1.

For each class (in our case the relaxed canonical patterns $\relaxedPattern$) the $F_1$ score is defined as following \cite{Christen2023-F1}:
\begin{equation}
    \begin{aligned}
        \text{Precision} &= \frac{t_p}{t_p + f_p}\\
        \text{Recall} &= \frac{t_p}{t_p + f_n}\\
        F_1 &= 2{\frac{\text{Precision}\times\text{Recall}}{\text{Precision}+\text{Recall}}}
        \label{eq:per_class_f1}
    \end{aligned}
\end{equation}

From this the macro averaged $F_1$ score is defined as:
\begin{equation}
    F_{1\text{macro}} = \frac{1}{\nRelaxedPatterns}\sum_{\patternIndex=1}^{\nRelaxedPatterns} F_{1\patternIndex}
    \label{eq:macro_f1}
\end{equation}
where $\nRelaxedPatterns$ is the number of relaxed canonical patterns.

\FloatBarrier

\section{Numerical Instability of Eigenvalue Calculations}\label{sec:app1}
\vspace*{12pt}
In section \ref{sec:foerstner-distance} we described how the generalised eigenvalue problem $\mathbf{A}x=\lambda\mathbf{B}x$ for $\mathbf{A},\mathbf{B}$ in $\mathbb{R}^{n\times n}$ can lead to the surprising effect of $\lambda_i<0$ due to numerical instabilities of Scipy's eigenvalue calculation. Here we show a concrete example of this surprising effect.

Let $\mathbf{A}=\big[\begin{smallmatrix}
  1 & -0.02 & - 0.02\\
  -0.02 & 1 & 1 \\
  -0.02 & 1 & 1
\end{smallmatrix}\big]$ and $\mathbf{B}=\big[\begin{smallmatrix}
  1 & -0.01 & - 0.01\\
  -0.01 & 1 & 1 \\
  -0.01 & 1 & 1
\end{smallmatrix}\big]$ be two similar, singular correlation matrices (row 2 and 3 for both matrices are identical). Theoretically given their singularity both matrices should have one eigenvalue of 0 leading to two inf eigenvalue in the generalised eigenvalue problem. However, the eigenvalues become $\lambda_\mathbf{A}=[9.99200639e-01+0.j, 2.00079936+0.j, 4.11810443e-32+0.j]$ and $\lambda_\mathbf{B}=[9.99800040e-01+0.j, 2.00019996+0.j, 4.11857080e-32+0.j]$ calculated using \texttt{scipy.linalg.eigvals}. Note that instead of $0$ the third eigenvalue for each matrix is very small. The eigenvalues for the generalised problem become $\lambda_{A,B}=[0.98989899,  1.00990099, -0.42480251]$. Theoretically, we expect that when $\mathbf{B}$ is singular two of the generalised eigenvalues become inf, instead we get a a negative eigenvalue that carries a complete different meaning!

This issue can be fixed by adding $\epsilon\mathbf{I}$ to the correlation matrices. For $\epsilon=1e-10$ the generalised eigenvalues for the above matrices become $\lambda_{A+\epsilon\mathbf{I},B+\epsilon\mathbf{I}}=[0.98989899, 1.00990099, 1.00000117]$ which correctly express that $\mathbf{A}$ and $\mathbf{B}$ are indeed very similar matrices. The eigenvalues for $\mathbf{A+\epsilon\mathbf{I}}$ and $\mathbf{B+\epsilon\mathbf{I}}$ become $\lambda_{\mathbf{A}+\epsilon\mathbf{I}}=[9.99200639e-01+0.j, 2.00079936+0.j, 9.99999217e-11+0.j]$ and $\lambda_{\mathbf{B}+\epsilon\mathbf{I}}=[9.99800040e-01+0.j, 2.00019996+0.j, 9.99999217e-11+0.j]$.

The regularisation of $\epsilon\mathbf{I}$ also fixes issues of correct inf generalised eigenvalues from singular matrices $\mathbf{A},\mathbf{B}$. For an example of this, let $\mathbf{A}=\big[\begin{smallmatrix}
  1 & -0.02 & - 0.02\\
  -0.02 & 1 & 1 \\
  -0.02 & 1 & 1
\end{smallmatrix}\big]$ and $\mathbf{B}=\big[\begin{smallmatrix}
  1 & -0.01 & - 0.01\\
  -0.01 & 1 & 1 \\
  -0.01 & 1 & 1
\end{smallmatrix}\big]$ be two similar, singular correlation matrices (row 2 and 3 for both matrices are identical). For these matrixes the eigenvalues are as following $\lambda_\mathbf{A}=[2.+0.j, 0.+0.j, 1.+0.j]$ and $\lambda_\mathbf{B}=[2.00019996+0.j, 4.32869044e-16+0.j, 9.99800040e-01+0.j]$ calculated using \texttt{scipy.linalg.eigvals} and the generalised eigenvalues are as we would expect from singular matrices $\lambda_{A,B}=[0.98874682, inf, inf]$. With regularisation the eigenvalues are $\lambda_{\mathbf{A}+\epsilon\mathbf{I}}=[2.00000000+0.j, 1.00000008e-10+0.j, 1.00000000+0.j]$ and $\lambda_{\mathbf{B}+\epsilon\mathbf{I}}=[2.00019996+0.j, 1.00000339e-10+0.j, 9.99800040e-01+0.j]$ and the generalised eigenvalues are $\lambda_{A+\epsilon\mathbf{I},B+\epsilon\mathbf{I}}=[1.01010101, 1.00000036, 0.99009901]$ which again correctly express that $\mathbf{A}$ and $\mathbf{B}$ are indeed very similar matrices. We can see that while the regularisation moves the matrices slightly out of being valid correlation matrices (diagonal no longer 1), it enables a much more numerically stable comparison that is closer to an intuitive understanding and avoids problems in distance calculations that uses eigenvalues.

We have also attempted to use \texttt{scipy.linalg.eigvalsh} instead which creates very similar results for the regularised matrices but fails to solve the first problem as it wrongly identifies the singular matrix $\mathbf{B}$ as being not positive semi-definite. We have reported this issue to Scipy see \href{https://github.com/scipy/scipy/issues/21951}{https://github.com/scipy/scipy/issues/21951}.

\FloatBarrier

\section{Additional Results for Distance Function Validity Assessment}\label{sec:additional-df-results}

\subsection{Validity Evaluations}

Table~\ref{tab:construct-mean-sd-normal-100} gives the mean and standard deviation for the 5 structural construct validity evaluations ($\distanceFunctionCorrxPatterny{x}{x}$, $\avgLevelSetDistance[i]$, $\avgRateOfIncreaseLevelSet{i}{j}$, $\entropy$, $\avgLevelSetEntropy$) and predictive criterion validity evaluation ($F_1$) for the normal data conditions without sparsification. Tables~\ref{tab:discriminant-mean-sd-raw-100} and~\ref{tab:discriminant-mean-sd-downsampled-100} show the same result for the raw data without correlation patterns (everything is canonical pattern 0), respectively, downsampled data where correlation patterns are distorted establishing Discriminant validity. Tables~\ref{tab:external-mean-sd-normal-70} to~\ref{tab:external-mean-sd-non-normal-10} show results under different sparsified and non-normal distribution data conditions to establish external validity.

\begin{table}[ht!]
\caption{Mean and standard deviation for the 5 structural construct validity evaluations and predictive criterion validity evaluation for the normal data conditions with no sparsification.}
\label{tab:construct-mean-sd-normal-100}
\centering
\small
\setlength{\tabcolsep}{2pt}
\begin{tabular*}{\columnwidth}{@{\extracolsep{\fill}}l c c c c c c}
\toprule
& $\distanceFunctionCorrxPatterny{x}{x}$ & $\avgLevelSetDistance[]$ & $\avgRateOfIncreaseLevelSet{i}{j}$ & \textbf{$\entropy$} & $\avgLevelSetEntropy$ & $F_1$ \\
\midrule
$\dLp[1]$ & 0.07 (SD 0.0)* & 1.0 (SD 0.0)* & 0.86 (SD 0.0)* & 4.15 (SD 0.03)* & 1.6 (SD 0.04)* & 1.0 (SD 0.0)* \\
$\dLp[2]$ & 0.05 (SD 0.0)* & 1.0 (SD 0.0)* & 0.51 (SD 0.0) & 4.2 (SD 0.03)* & 1.93 (SD 0.04)* & 1.0 (SD 0.0)* \\
$\dLp[3]$ & 0.05 (SD 0.0)* & 1.0 (SD 0.0)* & 0.43 (SD 0.0) & 4.44 (SD 0.03)* & 1.94 (SD 0.04)* & 1.0 (SD 0.0)* \\
$\dLp[5]$ & 0.05 (SD 0.0)* & 1.0 (SD 0.0)* & 0.39 (SD 0.0) & 4.27 (SD 0.02)* & 1.98 (SD 0.03)* & 1.0 (SD 0.0)* \\
$\dLp[\infty]$ & 0.05 (SD 0.0)* & 1.0 (SD 0.0)* & 0.33 (SD 0.0) & 3.6 (SD 0.03) & 1.88 (SD 0.04)* & 1.0 (SD 0.0)* \\
$\dRefp[1]$ & 0.29 (SD 0.01) & 1.0 (SD 0.0)* & 4.33 (SD 0.01)* & 4.92 (SD 0.02)* & 2.94 (SD 0.03)* & 1.0 (SD 0.0)* \\
$\dRefp[2]$ & 0.14 (SD 0.01) & 1.0 (SD 0.0)* & 1.65 (SD 0.01)* & 4.72 (SD 0.01)* & 2.86 (SD 0.03)* & 1.0 (SD 0.0)* \\
$\dRefp[3]$ & 0.12 (SD 0.01) & 1.0 (SD 0.0)* & 1.25 (SD 0.01)* & 4.88 (SD 0.01)* & 3.03 (SD 0.02) & 1.0 (SD 0.0)* \\
$\dRefp[5]$ & 0.12 (SD 0.0) & 1.0 (SD 0.0)* & 1.02 (SD 0.01)* & 4.83 (SD 0.02)* & 2.98 (SD 0.03)* & 1.0 (SD 0.0)* \\
$\dRefp[\infty]$ & 0.11 (SD 0.0) & 1.0 (SD 0.0)* & 0.8 (SD 0.0)* & 4.61 (SD 0.03)* & 2.92 (SD 0.04)* & 1.0 (SD 0.0)* \\
$\dDotp[1]$ & 0.09 (SD 0.0)* & 1.0 (SD 0.0)* & 0.99 (SD 0.0)* & 4.65 (SD 0.02)* & 2.54 (SD 0.02)* & 1.0 (SD 0.0)* \\
$\dDotp[2]$ & 0.06 (SD 0.0)* & 1.0 (SD 0.0)* & 0.53 (SD 0.0) & 4.71 (SD 0.02)* & 2.61 (SD 0.03)* & 1.0 (SD 0.0)* \\
$\dDotp[\infty]$ & 0.05 (SD 0.0)* & 1.0 (SD 0.0)* & 0.33 (SD 0.0) & 3.6 (SD 0.03) & 1.88 (SD 0.04)* & 1.0 (SD 0.0)* \\
$\dlogF$& 3.01 (SD 0.56) & 0.0 (SD 0.0) & 11.65 (SD 0.77)* & 4.45 (SD 0.04)* & 2.04 (SD 0.08)* & 0.89 (SD 0.03) \\
$\dFoer$& 3.65 (SD 0.53) & 0.0 (SD 0.0) & 5.62 (SD 0.54)* & 3.67 (SD 0.03) & 2.04 (SD 0.03)* & 0.37 (SD 0.02) \\
\bottomrule
\multicolumn{7}{l}{$^*$ Passed validity threshold.} \\
\end{tabular*}
\end{table}

\begin{table}[ht!]
\caption{Mean and standard deviation for the 5 structural construct validity evaluations and predictive criterion validity evaluation for raw data without canonical pattern other than pattern 0 (no correlations). This establishes deteriorated behaviour under discriminant validity conditions where canonical patterns are absent.}
\label{tab:discriminant-mean-sd-raw-100}
\centering
\small
\setlength{\tabcolsep}{2pt}
\begin{tabular*}{\columnwidth}{@{\extracolsep{\fill}}l c c c c c c}
\toprule
& $\distanceFunctionCorrxPatterny{x}{x}$ & $\avgLevelSetDistance[]$ & $\avgRateOfIncreaseLevelSet{i}{j}$ & \textbf{$\entropy$} & $\avgLevelSetEntropy$ & $F_1$ \\
\midrule
$\dLp[1]$ & 1.53 (SD 0.02) & 0.0 (SD 0.0) & 0.22 (SD 0.0) & 2.64 (SD 0.06) & 2.46 (SD 0.06)* & 0.0 (SD 0.0) \\
$\dLp[2]$ & 1.08 (SD 0.01) & 0.0 (SD 0.0) & 0.1 (SD 0.0) & 2.0 (SD 0.09) & 1.95 (SD 0.09)* & 0.0 (SD 0.0) \\
$\dLp[3]$ & 0.98 (SD 0.01) & 0.0 (SD 0.0) & 0.07 (SD 0.0) & 2.78 (SD 0.06) & 2.58 (SD 0.07)* & 0.0 (SD 0.0) \\
$\dLp[5]$ & 0.9 (SD 0.01) & 0.0 (SD 0.0) & 0.05 (SD 0.0) & 2.91 (SD 0.09) & 2.7 (SD 0.08)* & 0.0 (SD 0.0) \\
$\dLp[\infty]$ & 0.81 (SD 0.01) & 0.0 (SD 0.0) & 0.04 (SD 0.01) & 2.76 (SD 0.06) & 2.64 (SD 0.07)* & 0.0 (SD 0.0) \\
$\dRefp[1]$ & 6.27 (SD 0.11) & 0.0 (SD 0.0) & 1.11 (SD 0.02)* & 3.54 (SD 0.06) & 3.39 (SD 0.08) & 0.0 (SD 0.0) \\
$\dRefp[2]$ & 2.7 (SD 0.03) & 0.0 (SD 0.0) & 0.33 (SD 0.01) & 3.47 (SD 0.06) & 3.3 (SD 0.05) & 0.0 (SD 0.0) \\
$\dRefp[3]$ & 2.1 (SD 0.02) & 0.0 (SD 0.0) & 0.21 (SD 0.01) & 3.71 (SD 0.06) & 3.5 (SD 0.07) & 0.0 (SD 0.0) \\
$\dRefp[5]$ & 1.76 (SD 0.02) & 0.0 (SD 0.0) & 0.14 (SD 0.01) & 3.74 (SD 0.07) & 3.53 (SD 0.09) & 0.0 (SD 0.0) \\
$\dRefp[\infty]$ & 1.4 (SD 0.01) & 0.0 (SD 0.0) & 0.08 (SD 0.01) & 3.54 (SD 0.06) & 3.38 (SD 0.07) & 0.0 (SD 0.0) \\
$\dDotp[1]$ & 1.79 (SD 0.02) & 0.0 (SD 0.0) & 0.24 (SD 0.01) & 3.12 (SD 0.04) & 2.99 (SD 0.05)* & 0.0 (SD 0.0) \\
$\dDotp[2]$ & 1.13 (SD 0.01) & 0.0 (SD 0.0) & 0.1 (SD 0.0) & 2.83 (SD 0.05) & 2.76 (SD 0.07)* & 0.0 (SD 0.0) \\
$\dDotp[\infty]$ & 0.81 (SD 0.01) & 0.0 (SD 0.0) & 0.04 (SD 0.01) & 2.76 (SD 0.06) & 2.64 (SD 0.07)* & 0.0 (SD 0.0) \\
$\dlogF$& 14.74 (SD 0.31) & 0.0 (SD 0.0) & 2.66 (SD 0.55)* & 2.12 (SD 0.03) & 1.96 (SD 0.02)* & 0.0 (SD 0.0) \\
$\dFoer$& 14.8 (SD 0.31) & 0.0 (SD 0.0) & 2.65 (SD 0.55)* & 1.63 (SD 0.0) & 1.47 (SD 0.01)* & 0.0 (SD 0.0) \\
\bottomrule
\multicolumn{7}{l}{$^*$ Passed validity threshold.} \\
\end{tabular*}
\end{table}

\begin{table}[ht!]
\caption{Mean and standard deviation for the 5 structural construct validity evaluations and predictive criterion validity evaluation for downsampled data where pattern MAE increases to 0.13. This establishes deteriorated behaviour for discriminant validity.}
\label{tab:discriminant-mean-sd-downsampled-100}
\centering
\small
\setlength{\tabcolsep}{2pt}
\begin{tabular*}{\columnwidth}{@{\extracolsep{\fill}}l c c c c c c}
\toprule
& $\distanceFunctionCorrxPatterny{x}{x}$ & $\avgLevelSetDistance[]$ & $\avgRateOfIncreaseLevelSet{i}{j}$ & \textbf{$\entropy$} & $\avgLevelSetEntropy$ & $F_1$ \\
\midrule
$\dLp[1]$ & 0.39 (SD 0.07) & 1.0 (SD 0.0)* & 0.72 (SD 0.03)* & 5.29 (SD 0.03)* & 3.68 (SD 0.18) & 0.91 (SD 0.06) \\
$\dLp[2]$ & 0.27 (SD 0.05) & 1.0 (SD 0.0)* & 0.42 (SD 0.02) & 5.25 (SD 0.02)* & 3.8 (SD 0.16) & 0.91 (SD 0.06) \\
$\dLp[3]$ & 0.25 (SD 0.04) & 1.0 (SD 0.0)* & 0.36 (SD 0.02) & 5.25 (SD 0.02)* & 3.83 (SD 0.15) & 0.91 (SD 0.06) \\
$\dLp[5]$ & 0.24 (SD 0.04) & 1.0 (SD 0.0)* & 0.32 (SD 0.02) & 5.26 (SD 0.01)* & 3.9 (SD 0.15) & 0.91 (SD 0.06) \\
$\dLp[\infty]$ & 0.23 (SD 0.04) & 1.0 (SD 0.0)* & 0.28 (SD 0.01) & 5.21 (SD 0.07)* & 3.92 (SD 0.17) & 0.91 (SD 0.06) \\
$\dRefp[1]$ & 1.81 (SD 0.29) & 1.0 (SD 0.0)* & 3.35 (SD 0.23)* & 5.06 (SD 0.03)* & 4.04 (SD 0.07) & 0.87 (SD 0.1) \\
$\dRefp[2]$ & 0.8 (SD 0.13) & 1.0 (SD 0.0)* & 1.27 (SD 0.09)* & 4.99 (SD 0.03)* & 4.05 (SD 0.08) & 0.87 (SD 0.09) \\
$\dRefp[3]$ & 0.65 (SD 0.11) & 1.0 (SD 0.0)* & 0.95 (SD 0.07)* & 5.01 (SD 0.03)* & 4.1 (SD 0.08) & 0.87 (SD 0.1) \\
$\dRefp[5]$ & 0.58 (SD 0.1) & 1.0 (SD 0.0)* & 0.78 (SD 0.06)* & 5.07 (SD 0.03)* & 4.18 (SD 0.08) & 0.86 (SD 0.11) \\
$\dRefp[\infty]$ & 0.53 (SD 0.09) & 1.0 (SD 0.0)* & 0.63 (SD 0.04) & 5.2 (SD 0.03)* & 4.27 (SD 0.09) & 0.84 (SD 0.12) \\
$\dDotp[1]$ & 0.47 (SD 0.08) & 1.0 (SD 0.0)* & 0.82 (SD 0.04)* & 5.19 (SD 0.02)* & 3.83 (SD 0.13) & 0.9 (SD 0.07) \\
$\dDotp[2]$ & 0.28 (SD 0.05) & 1.0 (SD 0.0)* & 0.44 (SD 0.02) & 5.23 (SD 0.02)* & 3.86 (SD 0.13) & 0.91 (SD 0.06) \\
$\dDotp[\infty]$ & 0.23 (SD 0.04) & 1.0 (SD 0.0)* & 0.28 (SD 0.01) & 5.21 (SD 0.07)* & 3.92 (SD 0.17) & 0.91 (SD 0.06) \\
$\dlogF$& 12.42 (SD 0.57) & 0.0 (SD 0.0) & 3.31 (SD 0.74)* & 4.22 (SD 0.13)* & 3.14 (SD 0.07) & 0.02 (SD 0.02) \\
$\dFoer$& 12.57 (SD 0.56) & 0.0 (SD 0.0) & 2.96 (SD 0.59)* & 3.74 (SD 0.08) & 2.72 (SD 0.03)* & 0.01 (SD 0.01) \\
\bottomrule
\multicolumn{7}{l}{$^*$ Passed validity threshold.} \\
\end{tabular*}
\end{table}

\begin{table}[ht!]
\caption{Mean and standard deviation for the 5 structural construct validity evaluations and predictive criterion validity evaluation for normal data with 70\% of observations (sparsification). This establishes external validity across different data conditions.}
\label{tab:external-mean-sd-normal-70}
\centering
\small
\setlength{\tabcolsep}{2pt}
\begin{tabular*}{\columnwidth}{@{\extracolsep{\fill}}l c c c c c c}
\toprule
& $\distanceFunctionCorrxPatterny{x}{x}$ & $\avgLevelSetDistance[]$ & $\avgRateOfIncreaseLevelSet{i}{j}$ & \textbf{$\entropy$} & $\avgLevelSetEntropy$ & $F_1$ \\
\midrule
$\dLp[1]$ & 0.07 (SD 0.0)* & 1.0 (SD 0.0)* & 0.86 (SD 0.0)* & 4.2 (SD 0.04)* & 1.65 (SD 0.05)* & 1.0 (SD 0.0)* \\
$\dLp[2]$ & 0.06 (SD 0.0)* & 1.0 (SD 0.0)* & 0.51 (SD 0.0) & 4.23 (SD 0.03)* & 1.98 (SD 0.04)* & 1.0 (SD 0.0)* \\
$\dLp[3]$ & 0.05 (SD 0.0)* & 1.0 (SD 0.0)* & 0.43 (SD 0.0) & 4.48 (SD 0.03)* & 2.0 (SD 0.04)* & 1.0 (SD 0.0)* \\
$\dLp[5]$ & 0.05 (SD 0.0)* & 1.0 (SD 0.0)* & 0.39 (SD 0.0) & 4.31 (SD 0.03)* & 2.02 (SD 0.04)* & 1.0 (SD 0.0)* \\
$\dLp[\infty]$ & 0.05 (SD 0.0)* & 1.0 (SD 0.0)* & 0.33 (SD 0.0) & 3.64 (SD 0.04) & 1.93 (SD 0.05)* & 1.0 (SD 0.0)* \\
$\dRefp[1]$ & 0.3 (SD 0.02) & 1.0 (SD 0.0)* & 4.33 (SD 0.01)* & 4.93 (SD 0.02)* & 2.97 (SD 0.03)* & 1.0 (SD 0.0)* \\
$\dRefp[2]$ & 0.15 (SD 0.01) & 1.0 (SD 0.0)* & 1.65 (SD 0.01)* & 4.73 (SD 0.01)* & 2.88 (SD 0.03)* & 1.0 (SD 0.0)* \\
$\dRefp[3]$ & 0.13 (SD 0.01) & 1.0 (SD 0.0)* & 1.25 (SD 0.01)* & 4.89 (SD 0.01)* & 3.06 (SD 0.02) & 1.0 (SD 0.0)* \\
$\dRefp[5]$ & 0.12 (SD 0.01) & 1.0 (SD 0.0)* & 1.02 (SD 0.01)* & 4.85 (SD 0.02)* & 3.01 (SD 0.04) & 1.0 (SD 0.0)* \\
$\dRefp[\infty]$ & 0.11 (SD 0.01) & 1.0 (SD 0.0)* & 0.8 (SD 0.01)* & 4.64 (SD 0.04)* & 2.96 (SD 0.05)* & 1.0 (SD 0.0)* \\
$\dDotp[1]$ & 0.09 (SD 0.0)* & 1.0 (SD 0.0)* & 0.98 (SD 0.0)* & 4.67 (SD 0.02)* & 2.57 (SD 0.03)* & 1.0 (SD 0.0)* \\
$\dDotp[2]$ & 0.06 (SD 0.0)* & 1.0 (SD 0.0)* & 0.53 (SD 0.0) & 4.73 (SD 0.02)* & 2.64 (SD 0.03)* & 1.0 (SD 0.0)* \\
$\dDotp[\infty]$ & 0.05 (SD 0.0)* & 1.0 (SD 0.0)* & 0.33 (SD 0.0) & 3.64 (SD 0.04) & 1.93 (SD 0.05)* & 1.0 (SD 0.0)* \\
$\dlogF$& 3.03 (SD 0.61) & 0.0 (SD 0.0) & 11.65 (SD 0.7)* & 4.47 (SD 0.05)* & 2.06 (SD 0.08)* & 0.89 (SD 0.04) \\
$\dFoer$& 3.67 (SD 0.57) & 0.0 (SD 0.0) & 5.62 (SD 0.48)* & 3.67 (SD 0.03) & 2.05 (SD 0.03)* & 0.37 (SD 0.02) \\
\bottomrule
\multicolumn{7}{l}{$^*$ Passed validity threshold.} \\
\end{tabular*}
\end{table}

\begin{table}[ht!]
\caption{Mean and standard deviation for the 5 structural construct validity evaluations and predictive criterion validity evaluation for normal data with 10\% of observations (sparsification). This establishes external validity across different data conditions.}
\label{tab:external-mean-sd-normal-10}
\centering
\small
\setlength{\tabcolsep}{2pt}
\begin{tabular*}{\columnwidth}{@{\extracolsep{\fill}}l c c c c c c}
\toprule
& $\distanceFunctionCorrxPatterny{x}{x}$ & $\avgLevelSetDistance[]$ & $\avgRateOfIncreaseLevelSet{i}{j}$ & \textbf{$\entropy$} & $\avgLevelSetEntropy$ & $F_1$ \\
\midrule
$\dLp[1]$ & 0.1 (SD 0.01)* & 1.0 (SD 0.0)* & 0.86 (SD 0.0)* & 4.52 (SD 0.05)* & 2.04 (SD 0.07)* & 1.0 (SD 0.0)* \\
$\dLp[2]$ & 0.07 (SD 0.0)* & 1.0 (SD 0.0)* & 0.51 (SD 0.0) & 4.52 (SD 0.05)* & 2.38 (SD 0.06)* & 1.0 (SD 0.0)* \\
$\dLp[3]$ & 0.07 (SD 0.0)* & 1.0 (SD 0.0)* & 0.43 (SD 0.0) & 4.75 (SD 0.03)* & 2.4 (SD 0.06)* & 1.0 (SD 0.0)* \\
$\dLp[5]$ & 0.06 (SD 0.0)* & 1.0 (SD 0.0)* & 0.38 (SD 0.0) & 4.58 (SD 0.04)* & 2.41 (SD 0.07)* & 1.0 (SD 0.0)* \\
$\dLp[\infty]$ & 0.06 (SD 0.0)* & 1.0 (SD 0.0)* & 0.33 (SD 0.0) & 4.03 (SD 0.06)* & 2.37 (SD 0.07)* & 1.0 (SD 0.0)* \\
$\dRefp[1]$ & 0.41 (SD 0.03) & 1.0 (SD 0.0)* & 4.3 (SD 0.01)* & 4.98 (SD 0.02)* & 3.15 (SD 0.05) & 1.0 (SD 0.0)* \\
$\dRefp[2]$ & 0.19 (SD 0.01) & 1.0 (SD 0.0)* & 1.64 (SD 0.01)* & 4.82 (SD 0.02)* & 3.11 (SD 0.05) & 1.0 (SD 0.0)* \\
$\dRefp[3]$ & 0.16 (SD 0.01) & 1.0 (SD 0.0)* & 1.24 (SD 0.01)* & 4.94 (SD 0.01)* & 3.25 (SD 0.05) & 1.0 (SD 0.0)* \\
$\dRefp[5]$ & 0.15 (SD 0.01) & 1.0 (SD 0.0)* & 1.02 (SD 0.01)* & 4.95 (SD 0.02)* & 3.27 (SD 0.05) & 1.0 (SD 0.0)* \\
$\dRefp[\infty]$ & 0.14 (SD 0.01) & 1.0 (SD 0.0)* & 0.8 (SD 0.01)* & 4.9 (SD 0.03)* & 3.31 (SD 0.05) & 1.0 (SD 0.0)* \\
$\dDotp[1]$ & 0.12 (SD 0.01) & 1.0 (SD 0.0)* & 0.98 (SD 0.0)* & 4.83 (SD 0.03)* & 2.82 (SD 0.04)* & 1.0 (SD 0.0)* \\
$\dDotp[2]$ & 0.07 (SD 0.0)* & 1.0 (SD 0.0)* & 0.53 (SD 0.0) & 4.9 (SD 0.03)* & 2.88 (SD 0.04)* & 1.0 (SD 0.0)* \\
$\dDotp[\infty]$ & 0.06 (SD 0.0)* & 1.0 (SD 0.0)* & 0.33 (SD 0.0) & 4.03 (SD 0.06)* & 2.37 (SD 0.07)* & 1.0 (SD 0.0)* \\
$\dlogF$& 3.06 (SD 0.55) & 0.0 (SD 0.0) & 11.66 (SD 0.69)* & 4.55 (SD 0.04)* & 2.23 (SD 0.07)* & 0.89 (SD 0.03) \\
$\dFoer$& 3.69 (SD 0.52) & 0.0 (SD 0.0) & 5.62 (SD 0.5)* & 3.72 (SD 0.03) & 2.14 (SD 0.03)* & 0.37 (SD 0.02) \\
\bottomrule
\multicolumn{7}{l}{$^*$ Passed validity threshold.} \\
\end{tabular*}
\end{table}

\begin{table}[ht!]
\caption{Mean and standard deviation for the 5 structural construct validity evaluations and predictive criterion validity evaluation for non-normal data with 100\% of observations. This establishes external validity across different data conditions.}
\label{tab:external-mean-sd-non-normal-100}
\centering
\small
\setlength{\tabcolsep}{2pt}
\begin{tabular*}{\columnwidth}{@{\extracolsep{\fill}}l c c c c c c}
\toprule
& $\distanceFunctionCorrxPatterny{x}{x}$ & $\avgLevelSetDistance[]$ & $\avgRateOfIncreaseLevelSet{i}{j}$ & \textbf{$\entropy$} & $\avgLevelSetEntropy$ & $F_1$ \\
\midrule
$\dLp[1]$ & 0.07 (SD 0.01)* & 1.0 (SD 0.0)* & 0.86 (SD 0.0)* & 4.13 (SD 0.05)* & 1.6 (SD 0.05)* & 1.0 (SD 0.0)* \\
$\dLp[2]$ & 0.05 (SD 0.0)* & 1.0 (SD 0.0)* & 0.51 (SD 0.0) & 4.2 (SD 0.05)* & 1.96 (SD 0.06)* & 1.0 (SD 0.0)* \\
$\dLp[3]$ & 0.05 (SD 0.0)* & 1.0 (SD 0.0)* & 0.43 (SD 0.0) & 4.47 (SD 0.05)* & 1.99 (SD 0.07)* & 1.0 (SD 0.0)* \\
$\dLp[5]$ & 0.05 (SD 0.0)* & 1.0 (SD 0.0)* & 0.38 (SD 0.0) & 4.29 (SD 0.06)* & 2.0 (SD 0.07)* & 1.0 (SD 0.0)* \\
$\dLp[\infty]$ & 0.05 (SD 0.0)* & 1.0 (SD 0.0)* & 0.33 (SD 0.0) & 3.67 (SD 0.09) & 1.91 (SD 0.09)* & 1.0 (SD 0.0)* \\
$\dRefp[1]$ & 0.3 (SD 0.03) & 1.0 (SD 0.0)* & 4.31 (SD 0.03)* & 4.9 (SD 0.03)* & 2.89 (SD 0.05)* & 1.0 (SD 0.0)* \\
$\dRefp[2]$ & 0.15 (SD 0.01) & 1.0 (SD 0.0)* & 1.65 (SD 0.01)* & 4.72 (SD 0.03)* & 2.87 (SD 0.05)* & 1.0 (SD 0.0)* \\
$\dRefp[3]$ & 0.13 (SD 0.01) & 1.0 (SD 0.0)* & 1.24 (SD 0.01)* & 4.88 (SD 0.02)* & 3.02 (SD 0.03) & 1.0 (SD 0.0)* \\
$\dRefp[5]$ & 0.12 (SD 0.01) & 1.0 (SD 0.0)* & 1.02 (SD 0.01)* & 4.84 (SD 0.03)* & 2.98 (SD 0.04)* & 1.0 (SD 0.0)* \\
$\dRefp[\infty]$ & 0.11 (SD 0.01) & 1.0 (SD 0.0)* & 0.8 (SD 0.01)* & 4.65 (SD 0.07)* & 2.94 (SD 0.08)* & 1.0 (SD 0.0)* \\
$\dDotp[1]$ & 0.09 (SD 0.01)* & 1.0 (SD 0.0)* & 0.98 (SD 0.0)* & 4.65 (SD 0.03)* & 2.56 (SD 0.04)* & 1.0 (SD 0.0)* \\
$\dDotp[2]$ & 0.06 (SD 0.0)* & 1.0 (SD 0.0)* & 0.53 (SD 0.0) & 4.68 (SD 0.03)* & 2.54 (SD 0.04)* & 1.0 (SD 0.0)* \\
$\dDotp[\infty]$ & 0.05 (SD 0.0)* & 1.0 (SD 0.0)* & 0.33 (SD 0.0) & 3.67 (SD 0.09) & 1.91 (SD 0.09)* & 1.0 (SD 0.0)* \\
$\dlogF$& 9.37 (SD 0.58) & 0.0 (SD 0.0) & 4.96 (SD 1.01)* & 4.54 (SD 0.06)* & 2.39 (SD 0.14)* & 0.44 (SD 0.03) \\
$\dFoer$& 9.62 (SD 0.56) & 0.0 (SD 0.0) & 3.74 (SD 0.53)* & 3.86 (SD 0.03) & 2.23 (SD 0.06)* & 0.01 (SD 0.01) \\
\bottomrule
\multicolumn{7}{l}{$^*$ Passed validity threshold.} \\
\end{tabular*}
\end{table}

\begin{table}[ht!]
\caption{Mean and standard deviation for the 5 structural construct validity evaluations and predictive criterion validity evaluation for non-normal data with 10\% of observations (sparsification). This establishes external validity across different data conditions.}
\label{tab:external-mean-sd-non-normal-10}
\centering
\small
\setlength{\tabcolsep}{2pt}
\begin{tabular*}{\columnwidth}{@{\extracolsep{\fill}}l c c c c c c}
\toprule
& $\distanceFunctionCorrxPatterny{x}{x}$ & $\avgLevelSetDistance[]$ & $\avgRateOfIncreaseLevelSet{i}{j}$ & \textbf{$\entropy$} & $\avgLevelSetEntropy$ & $F_1$ \\
\midrule
$\dLp[1]$ & 0.1 (SD 0.01)* & 1.0 (SD 0.0)* & 0.85 (SD 0.0)* & 4.5 (SD 0.07)* & 2.04 (SD 0.08)* & 1.0 (SD 0.0)* \\
$\dLp[2]$ & 0.07 (SD 0.01)* & 1.0 (SD 0.0)* & 0.51 (SD 0.0) & 4.53 (SD 0.06)* & 2.4 (SD 0.06)* & 1.0 (SD 0.0)* \\
$\dLp[3]$ & 0.07 (SD 0.01)* & 1.0 (SD 0.0)* & 0.43 (SD 0.0) & 4.77 (SD 0.03)* & 2.44 (SD 0.06)* & 1.0 (SD 0.0)* \\
$\dLp[5]$ & 0.07 (SD 0.01)* & 1.0 (SD 0.0)* & 0.38 (SD 0.0) & 4.59 (SD 0.05)* & 2.42 (SD 0.08)* & 1.0 (SD 0.0)* \\
$\dLp[\infty]$ & 0.06 (SD 0.0)* & 1.0 (SD 0.0)* & 0.33 (SD 0.0) & 4.09 (SD 0.09)* & 2.39 (SD 0.09)* & 1.0 (SD 0.0)* \\
$\dRefp[1]$ & 0.42 (SD 0.04) & 1.0 (SD 0.0)* & 4.29 (SD 0.03)* & 4.97 (SD 0.03)* & 3.1 (SD 0.06) & 1.0 (SD 0.0)* \\
$\dRefp[2]$ & 0.19 (SD 0.02) & 1.0 (SD 0.0)* & 1.64 (SD 0.01)* & 4.83 (SD 0.03)* & 3.12 (SD 0.06) & 1.0 (SD 0.0)* \\
$\dRefp[3]$ & 0.16 (SD 0.01) & 1.0 (SD 0.0)* & 1.23 (SD 0.01)* & 4.93 (SD 0.02)* & 3.24 (SD 0.05) & 1.0 (SD 0.0)* \\
$\dRefp[5]$ & 0.15 (SD 0.01) & 1.0 (SD 0.0)* & 1.01 (SD 0.01)* & 4.96 (SD 0.02)* & 3.28 (SD 0.06) & 1.0 (SD 0.0)* \\
$\dRefp[\infty]$ & 0.14 (SD 0.01) & 1.0 (SD 0.0)* & 0.79 (SD 0.01)* & 4.92 (SD 0.05)* & 3.32 (SD 0.06) & 1.0 (SD 0.0)* \\
$\dDotp[1]$ & 0.12 (SD 0.01) & 1.0 (SD 0.0)* & 0.98 (SD 0.01)* & 4.83 (SD 0.03)* & 2.82 (SD 0.05)* & 1.0 (SD 0.0)* \\
$\dDotp[2]$ & 0.08 (SD 0.01)* & 1.0 (SD 0.0)* & 0.53 (SD 0.0) & 4.88 (SD 0.03)* & 2.82 (SD 0.04)* & 1.0 (SD 0.0)* \\
$\dDotp[\infty]$ & 0.06 (SD 0.0)* & 1.0 (SD 0.0)* & 0.33 (SD 0.0) & 4.09 (SD 0.09)* & 2.39 (SD 0.09)* & 1.0 (SD 0.0)* \\
$\dlogF$& 8.94 (SD 0.77) & 0.0 (SD 0.0) & 5.78 (SD 1.24)* & 4.61 (SD 0.07)* & 2.6 (SD 0.11)* & 0.46 (SD 0.04) \\
$\dFoer$& 9.2 (SD 0.73) & 0.0 (SD 0.0) & 3.87 (SD 0.51)* & 3.9 (SD 0.04) & 2.31 (SD 0.05)* & 0.02 (SD 0.02) \\
\bottomrule
\multicolumn{7}{l}{$^*$ Passed validity threshold.} \\
\end{tabular*}
\end{table}

\FloatBarrier

\subsection{Mean Average Rank}

\begin{table}[ht!]
\caption{Mean average rank and standard deviations for valid distance functions across data conditions.}
\label{tab:distance-measure-avg-ranks}
\centering
\small
\setlength{\tabcolsep}{3pt}
\begin{tabular*}{\columnwidth}{@{\extracolsep{\fill}}l c c c c c c}
\toprule
& \multicolumn{3}{c}{\textbf{Normal}} & \multicolumn{3}{c}{\textbf{Non-normal}} \\
\cmidrule(lr){2-4} \cmidrule(lr){5-7}
& \textbf{100\%} & \textbf{70\%} & \textbf{10\%} & \textbf{100\%} & \textbf{70\%} & \textbf{10\%} \\
\midrule
$\dLp[1]$ & 2.61 (SD 0.08) & 2.65 (SD 0.05) & 2.58 (SD 0.09) & 2.61 (SD 0.08) & 2.63 (SD 0.08) & 2.62 (SD 0.08) \\
$\dLp[2]$ & 2.68 (SD 0.12) & 2.72 (SD 0.09) & 2.78 (SD 0.11) & 2.66 (SD 0.10) & 2.70 (SD 0.12) & 2.72 (SD 0.09) \\
$\dLp[3]$ & 2.38 (SD 0.12)* & 2.46 (SD 0.12)* & 2.52 (SD 0.11) & 2.46 (SD 0.16)* & 2.53 (SD 0.13) & 2.62 (SD 0.09) \\
$\dLp[5]$ & 2.82 (SD 0.05) & 2.81 (SD 0.06) & 2.76 (SD 0.11) & 2.76 (SD 0.10) & 2.74 (SD 0.10) & 2.68 (SD 0.09) \\
$\dDotp[1]$ & 2.61 (SD 0.08) & 2.66 (SD 0.04) & 2.66 (SD 0.04) & 2.72 (SD 0.15) & 2.77 (SD 0.09) & 2.76 (SD 0.10) \\
$\dDotp[2]$ & 2.61 (SD 0.08) & 2.66 (SD 0.04) & 2.66 (SD 0.04) & 2.49 (SD 0.09) & 2.52 (SD 0.10)* & 2.56 (SD 0.09) \\
\bottomrule
\multicolumn{7}{l}{$^*$ statistically significantly lower ranked than $\dLp[1]$} \\
\end{tabular*}
\end{table}

\FloatBarrier

\subsection{Optimal Distance Functions - Family 1}
The valid distance functions were $\dLp[1]$, $\dLp[2]$, $\dLp[3]$, $\dLp[5]$, $\dDotp[1]$, and $\dDotp[2]$. When establishing preregistered hypotheses in Table~\ref{tab:preregistrated-family-1}, for Non-normal, Complete $\dDotp[2]$ was second best ranked but not significantly different from $\dLp[3]$, while for Non-normal $\dLp[3]$ was second best ranked but not significantly different from $\dDotp[2]$. For all other data variants, the significant differences were the top two ranked distance functions.

\begin{table}[htbp]
\caption{Preregistered confirmatory hypotheses (H) for family 1 to establish whether an optimal distance function exists. $H=x<y$ means distance function $x$ is ranked lower (better) than $y$. Seq is the sequence number in which the confirmatory tests need to be run, e indicates the effect size, and N is the number of pairs with a difference $>0.001$.}
\label{tab:preregistrated-family-1}
\centering
\small
\setlength{\tabcolsep}{3pt}
\begin{tabular*}{\columnwidth}{@{\extracolsep{\fill}}l c c c c c c c c}
\toprule
& \multicolumn{4}{c}{\textbf{Normal}} & \multicolumn{4}{c}{\textbf{Non-normal}}\\
\cmidrule(lr){2-5} \cmidrule(lr){6-9}
& \textbf{Seq} & \textbf{H} & \textbf{e} & \textbf{N} & \textbf{Seq} & \textbf{H} & \textbf{e} & \textbf{N}\\
\midrule
\textbf{Complete} (100\%) & 1. & $L_3$ < $L_1$ & 1.1 & 29 & 3. & $L_3$ < $L_1$ & 1.07 & 20 \\[0.3em]
\textbf{Partial} (70\%) & 2. & $L_3$ < $L_1$ & 1.08 & 25 & 4. & $L_{\text{dot}_2}$ < $L_1$ & 1.07 & 20\\[0.3em]
\textbf{Sparse} (10\%) & 6. & $L_3$ < $L_1$ & 0.46 & 19 & 5. & $L_{\text{dot}_2}$ < $L_3$ & 0.66 & 15\\
\bottomrule
\end{tabular*}
\end{table}

\begin{table}[!ht]
\caption{Confirmatory results for preregistered hypotheses for family 1, see Table~\ref{tab:preregistrated-family-1}. All hypotheses achieved statistical significance with large effect sizes and high power. N is the number of pairs with a difference $>0.001$ out of the $30$ confirmatory subjects.}
\label{tab:family1_statistical_validation_results}
\small
\setlength{\tabcolsep}{3pt}
\begin{tabular*}{\columnwidth}{@{\extracolsep{\fill}}l r l r r r r}
\toprule
\multicolumn{2}{c}{\textbf{Data Variant}} & 
\multicolumn{1}{c}{\textbf{H}} & 
\multicolumn{1}{c}{\textbf{p-value}} & 
\multicolumn{1}{c}{\textbf{Effect size}} & 
\multicolumn{1}{c}{\textbf{N}} &
\multicolumn{1}{c}{\textbf{Power \%}} \\
\midrule
\multirow{3}{*}{$\begin{array}{l}\textbf{Normal}\end{array}$} & 100\% & $L_3 < L_1$ & <0.0001 & 1.1 & 29 & >99.9\\
& 70\% & $L_3 < L_1$ & <0.0001 & 1.08 & 23 & >99.9 \\
& 10\% & \colorbox{black!10}{$L_3 < L_1^{**}$} & - & - & - & -\\
\midrule
\multirow{3}{*}{$\begin{array}{l}\textbf{Non-normal}\end{array}$} & 100\% & $L_3 < L_1$ & <0.0001 & 1.06 & 18 & >99.4 \\
& 70\% & $L_{\text{dot}_2}$ < $L_3$ & <0.0001 & 0.9 & 23 & >99.1 \\
& 10\% & \colorbox{black!10}{$L_{\text{dot}_2} < L_1^{*}$} & 0.5 & 0.19 & 13 & 0.1 \\
\bottomrule
\multicolumn{7}{l}{$^*$ Not statistically significant} \\
\multicolumn{7}{l}{$^{**}$ Not tested} \\
\end{tabular*}
\end{table}

\FloatBarrier

\section{Additional Results for ICVI Validity Assessment}\label{sec:additional-icvi-results}

\subsection{Criterion Validity}

\begin{table}[ht!]
\caption{Mean and standard deviation for the correlation between ICVI and Jaccard. The ICVIs are calculated using the respective distance function.}
\label{tab:icvi-mean-sd-normal-100}
\centering
\small
\setlength{\tabcolsep}{2pt}
\begin{tabular*}{\columnwidth}{@{\extracolsep{\fill}}l c c c c}
\toprule
$d$ & \textbf{SWC} & \textbf{DBI} & \textbf{VRC} & \textbf{PBM} \\
\midrule
\textbf{Normal 100\%} & & & & \\
$\dLp[1]$ & 0.91 (SD 0.01)* & -0.90 (SD 0.10)* & 0.21 (SD 0.03) & 0.21 (SD 0.03) \\
$\dLp[2]$ & 0.91 (SD 0.01)* & -0.90 (SD 0.10)* & 0.21 (SD 0.03) & 0.21 (SD 0.03) \\
$\dLp[3]$ & 0.91 (SD 0.01)* & -0.90 (SD 0.10)* & 0.21 (SD 0.03) & 0.21 (SD 0.03) \\
$\dLp[5]$ & 0.91 (SD 0.01)* & -0.90 (SD 0.10)* & 0.21 (SD 0.03) & 0.21 (SD 0.03) \\
$\dDotp[1]$ & 0.91 (SD 0.01)* & -0.90 (SD 0.10)* & 0.21 (SD 0.03) & 0.21 (SD 0.03) \\
$\dDotp[2]$ & 0.91 (SD 0.01)* & -0.90 (SD 0.10)* & 0.21 (SD 0.03) & 0.21 (SD 0.03) \\
\midrule
\textbf{Normal 70\%} & & & & \\
$\dLp[1]$ & 0.91 (SD 0.01)* & -0.90 (SD 0.12)* & 0.23 (SD 0.03) & 0.24 (SD 0.03) \\
$\dLp[2]$ & 0.91 (SD 0.01)* & -0.90 (SD 0.12)* & 0.23 (SD 0.03) & 0.24 (SD 0.03) \\
$\dLp[3]$ & 0.92 (SD 0.01)* & -0.90 (SD 0.11)* & 0.23 (SD 0.03) & 0.24 (SD 0.03) \\
$\dLp[5]$ & 0.92 (SD 0.01)* & -0.90 (SD 0.11)* & 0.23 (SD 0.03) & 0.24 (SD 0.03) \\
$\dDotp[1]$ & 0.91 (SD 0.01)* & -0.90 (SD 0.12)* & 0.23 (SD 0.03) & 0.23 (SD 0.03) \\
$\dDotp[2]$ & 0.91 (SD 0.01)* & -0.90 (SD 0.12)* & 0.23 (SD 0.03) & 0.24 (SD 0.03) \\
\midrule
\textbf{Normal 10\%} & & & & \\
$\dLp[1]$ & 0.89 (SD 0.02)* & -0.88 (SD 0.17)* & 0.32 (SD 0.05) & 0.34 (SD 0.04) \\
$\dLp[2]$ & 0.89 (SD 0.01)* & -0.88 (SD 0.17)* & 0.33 (SD 0.05) & 0.34 (SD 0.04) \\
$\dLp[3]$ & 0.89 (SD 0.01)* & -0.88 (SD 0.17)* & 0.33 (SD 0.05) & 0.34 (SD 0.04) \\
$\dLp[5]$ & 0.89 (SD 0.01)* & -0.88 (SD 0.17)* & 0.33 (SD 0.05) & 0.34 (SD 0.04) \\
$\dDotp[1]$ & 0.89 (SD 0.01)* & -0.88 (SD 0.17)* & 0.32 (SD 0.05) & 0.33 (SD 0.04) \\
$\dDotp[2]$ & 0.89 (SD 0.01)* & -0.88 (SD 0.17)* & 0.33 (SD 0.05) & 0.34 (SD 0.04) \\
\midrule
\textbf{Non-normal 100\%} & & & & \\
$\dLp[1]$ & 0.91 (SD 0.01)* & -0.90 (SD 0.09)* & 0.21 (SD 0.03) & 0.21 (SD 0.03) \\
$\dLp[2]$ & 0.91 (SD 0.01)* & -0.90 (SD 0.09)* & 0.21 (SD 0.03) & 0.21 (SD 0.03) \\
$\dLp[3]$ & 0.91 (SD 0.01)* & -0.90 (SD 0.09)* & 0.21 (SD 0.03) & 0.21 (SD 0.03) \\
$\dLp[5]$ & 0.91 (SD 0.01)* & -0.90 (SD 0.09)* & 0.21 (SD 0.03) & 0.21 (SD 0.03) \\
$\dDotp[1]$ & 0.91 (SD 0.01)* & -0.90 (SD 0.10)* & 0.21 (SD 0.03) & 0.21 (SD 0.03) \\
$\dDotp[2]$ & 0.91 (SD 0.01)* & -0.90 (SD 0.09)* & 0.21 (SD 0.03) & 0.21 (SD 0.03) \\
\midrule
\textbf{Non-normal 10\%} & & & & \\
$\dLp[1]$ & 0.89 (SD 0.02)* & -0.91 (SD 0.06)* & 0.32 (SD 0.04) & 0.34 (SD 0.04) \\
$\dLp[2]$ & 0.89 (SD 0.01)* & -0.91 (SD 0.07)* & 0.33 (SD 0.05) & 0.34 (SD 0.04) \\
$\dLp[3]$ & 0.89 (SD 0.01)* & -0.91 (SD 0.07)* & 0.33 (SD 0.05) & 0.34 (SD 0.04) \\
$\dLp[5]$ & 0.89 (SD 0.01)* & -0.91 (SD 0.07)* & 0.33 (SD 0.05) & 0.34 (SD 0.04) \\
$\dDotp[1]$ & 0.89 (SD 0.01)* & -0.91 (SD 0.06)* & 0.32 (SD 0.05) & 0.33 (SD 0.04) \\
$\dDotp[2]$ & 0.89 (SD 0.01)* & -0.91 (SD 0.07)* & 0.33 (SD 0.05) & 0.34 (SD 0.04) \\
\bottomrule
\multicolumn{5}{l}{$^*$ Passed validity threshold of $|r|>0.5$.} \\
\end{tabular*}
\end{table}

\FloatBarrier

\subsection{Structural Validity}
\begin{table}[ht!]
\caption{Structural 1: Mean and standard deviation of ICVIs calculated using the valid distance functions for the ground truth clustering in CSTS (best quality clustering).}
\label{tab:construct-test-1}
\centering
\small
\setlength{\tabcolsep}{2pt}
\begin{tabular*}{\columnwidth}{@{\extracolsep{\fill}}l l l c c}
\toprule
$d$ & \textbf{SWC} & \textbf{DBI} & \textbf{VRC} & \textbf{PBM} \\
\midrule
\textbf{Normal 100\%} & & & & \\
$\dLp[1]$ & 0.97 (SD 0.00)* & 0.05 (SD 0.01)* & 13475.54 (SD 3038.84)* & 15.14 (SD 3.05)* \\
$\dLp[2]$ & 0.97 (SD 0.00)* & 0.04 (SD 0.01)* & 12762.34 (SD 2825.69)* & 6.63 (SD 1.33) \\
$\dLp[3]$ & 0.98 (SD 0.00)* & 0.04 (SD 0.01)* & 12487.38 (SD 2748.09)* & 5.28 (SD 1.06) \\
$\dLp[5]$ & 0.98 (SD 0.00)* & 0.04 (SD 0.01)* & 12061.59 (SD 2656.02)* & 4.31 (SD 0.86) \\
$\dDotp[1]$ & 0.97 (SD 0.00)* & 0.05 (SD 0.01)* & 13324.21 (SD 2958.72)* & 25.47 (SD 4.99)* \\
$\dDotp[2]$ & 0.97 (SD 0.00)* & 0.04 (SD 0.01)* & 12693.24 (SD 2783.29)* & 7.71 (SD 1.52) \\
\midrule
\textbf{Normal 70\%} & & & & \\
$\dLp[1]$ & 0.97 (SD 0.00)* & 0.06 (SD 0.01)* & 9313.93 (SD 2351.48)* & 10.56 (SD 2.09)* \\
$\dLp[2]$ & 0.97 (SD 0.00)* & 0.05 (SD 0.01)* & 8849.07 (SD 2300.49)* & 4.62 (SD 0.94) \\
$\dLp[3]$ & 0.97 (SD 0.00)* & 0.05 (SD 0.01)* & 8678.76 (SD 2264.80)* & 3.69 (SD 0.75) \\
$\dLp[5]$ & 0.97 (SD 0.00)* & 0.05 (SD 0.01)* & 8397.54 (SD 2186.52)* & 3.01 (SD 0.61) \\
$\dDotp[1]$ & 0.97 (SD 0.00)* & 0.06 (SD 0.01)* & 9288.92 (SD 2396.67)* & 17.84 (SD 3.60)* \\
$\dDotp[2]$ & 0.97 (SD 0.00)* & 0.05 (SD 0.01)* & 8832.04 (SD 2308.71)* & 5.39 (SD 1.11) \\
\midrule
\textbf{Normal 10\%} & & & & \\
$\dLp[1]$ & 0.91 (SD 0.01)* & 0.16 (SD 0.02) & 1257.43 (SD 388.43)* & 1.48 (SD 0.33) \\
$\dLp[2]$ & 0.91 (SD 0.01)* & 0.14 (SD 0.02)* & 1186.00 (SD 371.46)* & 0.64 (SD 0.14) \\
$\dLp[3]$ & 0.92 (SD 0.01)* & 0.14 (SD 0.02)* & 1162.61 (SD 364.33)* & 0.51 (SD 0.11) \\
$\dLp[5]$ & 0.92 (SD 0.01)* & 0.14 (SD 0.02)* & 1126.41 (SD 352.88)* & 0.41 (SD 0.09) \\
$\dDotp[1]$ & 0.91 (SD 0.01)* & 0.16 (SD 0.02) & 1273.07 (SD 390.40)* & 2.52 (SD 0.55) \\
$\dDotp[2]$ & 0.91 (SD 0.01)* & 0.14 (SD 0.02)* & 1189.52 (SD 370.66)* & 0.74 (SD 0.17) \\
\midrule
\textbf{Non-normal 100\%} & & & & \\
$\dLp[1]$ & 0.97 (SD 0.00)* & 0.05 (SD 0.01)* & 13391.78 (SD 3178.89)* & 15.27 (SD 3.32)* \\
$\dLp[2]$ & 0.97 (SD 0.00)* & 0.04 (SD 0.01)* & 12632.33 (SD 2912.86)* & 6.62 (SD 1.40) \\
$\dLp[3]$ & 0.98 (SD 0.00)* & 0.04 (SD 0.01)* & 12343.12 (SD 2822.84)* & 5.25 (SD 1.11) \\
$\dLp[5]$ & 0.98 (SD 0.00)* & 0.04 (SD 0.01)* & 11902.30 (SD 2719.74)* & 4.27 (SD 0.90) \\
$\dDotp[1]$ & 0.97 (SD 0.00)* & 0.05 (SD 0.01)* & 13250.29 (SD 3064.01)* & 25.68 (SD 5.44)* \\
$\dDotp[2]$ & 0.97 (SD 0.00)* & 0.04 (SD 0.01)* & 12573.30 (SD 2861.58)* & 7.70 (SD 1.61) \\
\midrule
\textbf{Non-normal 10\%} & & & & \\
$\dLp[1]$ & 0.90 (SD 0.01) & 0.16 (SD 0.03) & 1237.54 (SD 389.21)* & 1.44 (SD 0.33) \\
$\dLp[2]$ & 0.91 (SD 0.01)* & 0.15 (SD 0.02) & 1169.73 (SD 373.71)* & 0.62 (SD 0.14) \\
$\dLp[3]$ & 0.91 (SD 0.01)* & 0.14 (SD 0.02)* & 1146.87 (SD 367.16)* & 0.49 (SD 0.11) \\
$\dLp[5]$ & 0.92 (SD 0.01)* & 0.14 (SD 0.02)* & 1110.89 (SD 356.29)* & 0.40 (SD 0.09) \\
$\dDotp[1]$ & 0.90 (SD 0.01) & 0.16 (SD 0.03) & 1253.41 (SD 393.34)* & 2.45 (SD 0.57) \\
$\dDotp[2]$ & 0.91 (SD 0.01)* & 0.15 (SD 0.02) & 1173.81 (SD 373.38)* & 0.73 (SD 0.17) \\
\bottomrule
\multicolumn{5}{l}{$^*$ Passed validity threshold of $\text{SWC}>0.9$,  $\text{DBI}<0.15$, $\text{VRC}>1000$, and $\text{PBM}>10$}.\\
\end{tabular*}
\end{table}

\begin{table}[ht!]
\caption{Structural 2: Mean and standard deviation of ICVIs calculated using the valid distance functions for the clustering in CSTS with the highest MAE (worst quality clustering).}
\label{tab:construct-test-2}
\centering
\small
\setlength{\tabcolsep}{2pt}
\begin{tabular*}{\columnwidth}{@{\extracolsep{\fill}}l c c c c}
\toprule
$d$ & \textbf{SWC} & \textbf{DBI} & \textbf{VRC} & \textbf{PBM} \\
\midrule
\textbf{Normal 100\%} & & & & \\
$\dLp[1]$ & -0.41 (SD 0.04)* & 7.13 (SD 1.14)* & 1.32 (SD 0.20)* & 0.00 (SD 0.00)* \\
$\dLp[2]$ & -0.39 (SD 0.03)* & 6.84 (SD 1.08)* & 1.33 (SD 0.16)* & 0.00 (SD 0.00)* \\
$\dLp[3]$ & -0.39 (SD 0.03)* & 6.80 (SD 1.09)* & 1.35 (SD 0.15)* & 0.00 (SD 0.00)* \\
$\dLp[5]$ & -0.38 (SD 0.03)* & 6.81 (SD 1.12)* & 1.36 (SD 0.14)* & 0.00 (SD 0.00)* \\
$\dDotp[1]$ & -0.41 (SD 0.04)* & 7.09 (SD 1.10)* & 1.33 (SD 0.18)* & 0.00 (SD 0.00)* \\
$\dDotp[2]$ & -0.39 (SD 0.03)* & 6.84 (SD 1.08)* & 1.34 (SD 0.15)* & 0.00 (SD 0.00)* \\
\midrule
\textbf{Normal 70\%} & & & & \\
$\dLp[1]$ & -0.41 (SD 0.04)* & 7.14 (SD 1.13)* & 1.32 (SD 0.20)* & 0.00 (SD 0.00)* \\
$\dLp[2]$ & -0.39 (SD 0.03)* & 6.84 (SD 1.05)* & 1.34 (SD 0.16)* & 0.00 (SD 0.00)* \\
$\dLp[3]$ & -0.39 (SD 0.03)* & 6.79 (SD 1.06)* & 1.35 (SD 0.15)* & 0.00 (SD 0.00)* \\
$\dLp[5]$ & -0.38 (SD 0.03)* & 6.80 (SD 1.09)* & 1.36 (SD 0.14)* & 0.00 (SD 0.00)* \\
$\dDotp[1]$ & -0.41 (SD 0.04)* & 7.09 (SD 1.09)* & 1.33 (SD 0.18)* & 0.00 (SD 0.00)* \\
$\dDotp[2]$ & -0.39 (SD 0.03)* & 6.83 (SD 1.05)* & 1.34 (SD 0.15)* & 0.00 (SD 0.00)* \\
\midrule
\textbf{Normal 10\%} & & & & \\
$\dLp[1]$ & -0.40 (SD 0.04)* & 7.18 (SD 0.94)* & 1.31 (SD 0.19)* & 0.00 (SD 0.00)* \\
$\dLp[2]$ & -0.39 (SD 0.04)* & 6.92 (SD 0.95)* & 1.33 (SD 0.15)* & 0.00 (SD 0.00)* \\
$\dLp[3]$ & -0.38 (SD 0.03)* & 6.90 (SD 0.98)* & 1.35 (SD 0.14)* & 0.00 (SD 0.00)* \\
$\dLp[5]$ & -0.38 (SD 0.03)* & 6.92 (SD 1.02)* & 1.36 (SD 0.14)* & 0.00 (SD 0.00)* \\
$\dDotp[1]$ & -0.40 (SD 0.04)* & 7.18 (SD 0.95)* & 1.32 (SD 0.18)* & 0.00 (SD 0.00)* \\
$\dDotp[2]$ & -0.39 (SD 0.04)* & 6.93 (SD 0.95)* & 1.33 (SD 0.15)* & 0.00 (SD 0.00)* \\
\midrule
\textbf{Non-normal 100\%} & & & & \\
$\dLp[1]$ & -0.41 (SD 0.04)* & 7.12 (SD 1.12)* & 1.32 (SD 0.20)* & 0.00 (SD 0.00)* \\
$\dLp[2]$ & -0.39 (SD 0.03)* & 6.84 (SD 1.09)* & 1.33 (SD 0.16)* & 0.00 (SD 0.00)* \\
$\dLp[3]$ & -0.39 (SD 0.03)* & 6.81 (SD 1.12)* & 1.35 (SD 0.15)* & 0.00 (SD 0.00)* \\
$\dLp[5]$ & -0.39 (SD 0.03)* & 6.82 (SD 1.17)* & 1.36 (SD 0.14)* & 0.00 (SD 0.00)* \\
$\dDotp[1]$ & -0.41 (SD 0.04)* & 7.09 (SD 1.11)* & 1.33 (SD 0.18)* & 0.00 (SD 0.00)* \\
$\dDotp[2]$ & -0.39 (SD 0.03)* & 6.84 (SD 1.10)* & 1.34 (SD 0.15)* & 0.00 (SD 0.00)* \\
\midrule
\textbf{Non-normal 10\%} & & & & \\
$\dLp[1]$ & -0.41 (SD 0.04)* & 7.13 (SD 0.99)* & 1.31 (SD 0.19)* & 0.00 (SD 0.00)* \\
$\dLp[2]$ & -0.39 (SD 0.03)* & 6.86 (SD 0.97)* & 1.33 (SD 0.15)* & 0.00 (SD 0.00)* \\
$\dLp[3]$ & -0.38 (SD 0.03)* & 6.83 (SD 0.98)* & 1.34 (SD 0.14)* & 0.00 (SD 0.00)* \\
$\dLp[5]$ & -0.38 (SD 0.03)* & 6.84 (SD 1.01)* & 1.35 (SD 0.13)* & 0.00 (SD 0.00)* \\
$\dDotp[1]$ & -0.41 (SD 0.04)* & 7.13 (SD 0.99)* & 1.32 (SD 0.18)* & 0.00 (SD 0.00)* \\
$\dDotp[2]$ & -0.39 (SD 0.04)* & 6.87 (SD 0.97)* & 1.33 (SD 0.15)* & 0.00 (SD 0.00)* \\
\bottomrule
\multicolumn{5}{l}{$^*$ Passed validity threshold of $\text{SWC}<0$,  $\text{DBI}>2$, $\text{VRC}<100$, and $\text{PBM}<10$.} \\
\end{tabular*}
\end{table}

\begin{table}[ht!]
\caption{Structural 3: Mean and standard deviation of SWC and DBI indices for reduced cluster counts of 6 and 11 clusters, calculated using valid distance functions.}
\label{tab:construct-test-3-swc-dbi}
\centering
\small
\setlength{\tabcolsep}{2pt}
\begin{tabular*}{\columnwidth}{@{\extracolsep{\fill}}l c c c c}
\toprule
 & \multicolumn{2}{c}{\textbf{SWC}} & \multicolumn{2}{c}{\textbf{DBI}} \\
\cmidrule(lr){2-3} \cmidrule(lr){4-5}
$d$ & 11 clusters & 6 clusters & 11 clusters & 6 clusters \\
\midrule
\textbf{Normal 100\%} & & & & \\
$\dLp[1]$ & 0.97 (SD 0.01)* & 0.98 (SD 0.01)* & 0.04 (SD 0.01)* & 0.04 (SD 0.01)* \\
$\dLp[2]$ & 0.98 (SD 0.01)* & 0.98 (SD 0.01)* & 0.04 (SD 0.01)* & 0.03 (SD 0.01)* \\
$\dLp[3]$ & 0.98 (SD 0.01)* & 0.98 (SD 0.01)* & 0.04 (SD 0.01)* & 0.03 (SD 0.01)* \\
$\dLp[5]$ & 0.98 (SD 0.01)* & 0.98 (SD 0.01)* & 0.04 (SD 0.01)* & 0.03 (SD 0.01)* \\
$\dDotp[1]$ & 0.98 (SD 0.01)* & 0.98 (SD 0.01)* & 0.04 (SD 0.01)* & 0.04 (SD 0.01)* \\
$\dDotp[2]$ & 0.98 (SD 0.01)* & 0.98 (SD 0.01)* & 0.04 (SD 0.01)* & 0.03 (SD 0.01)* \\
\midrule
\textbf{Normal 70\%} & & & & \\
$\dLp[1]$ & 0.97 (SD 0.01)* & 0.97 (SD 0.01)* & 0.05 (SD 0.01)* & 0.04 (SD 0.01)* \\
$\dLp[2]$ & 0.97 (SD 0.01)* & 0.97 (SD 0.01)* & 0.04 (SD 0.01)* & 0.04 (SD 0.01)* \\
$\dLp[3]$ & 0.97 (SD 0.01)* & 0.98 (SD 0.01)* & 0.04 (SD 0.01)* & 0.04 (SD 0.01)* \\
$\dLp[5]$ & 0.97 (SD 0.01)* & 0.98 (SD 0.01)* & 0.04 (SD 0.01)* & 0.04 (SD 0.01)* \\
$\dDotp[1]$ & 0.97 (SD 0.01)* & 0.97 (SD 0.01)* & 0.05 (SD 0.01)* & 0.04 (SD 0.01)* \\
$\dDotp[2]$ & 0.97 (SD 0.01)* & 0.97 (SD 0.01)* & 0.04 (SD 0.01)* & 0.04 (SD 0.01)* \\
\midrule
\textbf{Normal 10\%} & & & & \\
$\dLp[1]$ & 0.92 (SD 0.02)* & 0.93 (SD 0.02)* & 0.13 (SD 0.03)* & 0.11 (SD 0.03)* \\
$\dLp[2]$ & 0.92 (SD 0.02)* & 0.93 (SD 0.02)* & 0.12 (SD 0.03)* & 0.10 (SD 0.02)* \\
$\dLp[3]$ & 0.93 (SD 0.02)* & 0.93 (SD 0.02)* & 0.12 (SD 0.02)* & 0.10 (SD 0.02)* \\
$\dLp[5]$ & 0.93 (SD 0.02)* & 0.93 (SD 0.02)* & 0.11 (SD 0.02)* & 0.10 (SD 0.02)* \\
$\dDotp[1]$ & 0.92 (SD 0.02)* & 0.93 (SD 0.02)* & 0.13 (SD 0.03)* & 0.11 (SD 0.03)* \\
$\dDotp[2]$ & 0.92 (SD 0.02)* & 0.93 (SD 0.02)* & 0.12 (SD 0.03)* & 0.10 (SD 0.02)* \\
\midrule
\textbf{Non-normal 100\%} & & & & \\
$\dLp[1]$ & 0.97 (SD 0.01)* & 0.98 (SD 0.01)* & 0.04 (SD 0.01)* & 0.04 (SD 0.01)* \\
$\dLp[2]$ & 0.98 (SD 0.01)* & 0.98 (SD 0.01)* & 0.04 (SD 0.01)* & 0.03 (SD 0.01)* \\
$\dLp[3]$ & 0.98 (SD 0.01)* & 0.98 (SD 0.01)* & 0.04 (SD 0.01)* & 0.03 (SD 0.01)* \\
$\dLp[5]$ & 0.98 (SD 0.01)* & 0.98 (SD 0.01)* & 0.04 (SD 0.01)* & 0.03 (SD 0.01)* \\
$\dDotp[1]$ & 0.98 (SD 0.01)* & 0.98 (SD 0.01)* & 0.04 (SD 0.01)* & 0.04 (SD 0.01)* \\
$\dDotp[2]$ & 0.98 (SD 0.01)* & 0.98 (SD 0.01)* & 0.04 (SD 0.01)* & 0.03 (SD 0.01)* \\
\midrule
\textbf{Non-normal 10\%} & & & & \\
$\dLp[1]$ & 0.92 (SD 0.02)* & 0.93 (SD 0.02)* & 0.13 (SD 0.03)* & 0.11 (SD 0.03)* \\
$\dLp[2]$ & 0.92 (SD 0.02)* & 0.93 (SD 0.02)* & 0.12 (SD 0.03)* & 0.10 (SD 0.02)* \\
$\dLp[3]$ & 0.93 (SD 0.02)* & 0.93 (SD 0.02)* & 0.12 (SD 0.02)* & 0.10 (SD 0.02)* \\
$\dLp[5]$ & 0.93 (SD 0.02)* & 0.93 (SD 0.02)* & 0.12 (SD 0.02)* & 0.10 (SD 0.02)* \\
$\dDotp[1]$ & 0.92 (SD 0.02)* & 0.93 (SD 0.02)* & 0.13 (SD 0.03)* & 0.11 (SD 0.03)* \\
$\dDotp[2]$ & 0.92 (SD 0.02)* & 0.93 (SD 0.02)* & 0.12 (SD 0.03)* & 0.10 (SD 0.02)* \\
\bottomrule
\multicolumn{5}{l}{$^*$ Passed validity threshold of $\text{SWC}>0.9$ and $\text{DBI}<0.15$.} \\
\end{tabular*}
\end{table}

\begin{table}[ht!]
\caption{Structural 3: Mean and standard deviation of VRC and PBM indices for reduced cluster counts of 6 and 11 clusters, calculated using valid distance functions.}
\label{tab:construct-test-3-vrc-pmb}
\centering
\small
\setlength{\tabcolsep}{2pt}
\begin{tabular*}{\columnwidth}{@{\extracolsep{\fill}}l c c c c}
\toprule
$d$ & \multicolumn{2}{c}{\textbf{VRC}} & \multicolumn{2}{c}{\textbf{PBM}} \\
\cmidrule(lr){2-3} \cmidrule(lr){4-5}
 & 11 clusters & 6 clusters & 11 clusters & 6 clusters \\
\midrule
\textbf{Normal 100\%} & & & & \\
$\dLp[1]$ & 17196.30 (SD 10420.54)* & 20985.09 (SD 16753.32)* & 86.27 (SD 61.30)* & 282.65 (SD 254.99)* \\
$\dLp[2]$ & 15271.93 (SD 8375.50)* & 18768.77 (SD 14388.12)* & 33.97 (SD 23.91)* & 106.98 (SD 96.18)* \\
$\dLp[3]$ & 14859.39 (SD 7894.11)* & 18333.11 (SD 14114.06)* & 26.58 (SD 18.80)* & 82.78 (SD 75.45)* \\
$\dLp[5]$ & 14612.00 (SD 7594.39)* & 18178.36 (SD 14224.07)* & 22.10 (SD 15.37)* & 69.70 (SD 63.04)* \\
$\dDotp[1]$ & 17227.56 (SD 10269.43)* & 19493.51 (SD 14707.92)* & 129.00 (SD 93.07)* & 356.84 (SD 304.95)* \\
$\dDotp[2]$ & 15240.01 (SD 8287.83)* & 18181.85 (SD 13650.63)* & 36.80 (SD 25.30)* & 109.45 (SD 95.51)* \\
\midrule
\textbf{Normal 70\%} & & & & \\
$\dLp[1]$ & 11589.44 (SD 6665.89)* & 13617.85 (SD 9699.38)* & 56.61 (SD 34.42)* & 189.92 (SD 161.09)* \\
$\dLp[2]$ & 10509.94 (SD 5793.45)* & 12145.79 (SD 8057.84)* & 22.49 (SD 13.81)* & 71.07 (SD 59.30)* \\
$\dLp[3]$ & 10270.58 (SD 5541.95)* & 11860.22 (SD 7783.90)* & 17.60 (SD 10.84)* & 54.85 (SD 45.86)* \\
$\dLp[5]$ & 10118.74 (SD 5375.40)* & 11738.74 (SD 7692.56)* & 14.62 (SD 8.80)* & 46.09 (SD 37.52)* \\
$\dDotp[1]$ & 11549.30 (SD 6240.22)* & 12608.27 (SD 8100.56)* & 84.34 (SD 52.71)* & 240.86 (SD 197.23)* \\
$\dDotp[2]$ & 10474.45 (SD 5617.37)* & 11755.38 (SD 7487.11)* & 24.30 (SD 14.43)* & 72.65 (SD 58.14)* \\
\midrule
\textbf{Normal 10\%} & & & & \\
$\dLp[1]$ & 1601.65 (SD 905.04)* & 2029.33 (SD 1193.11)* & 8.47 (SD 5.05) & 28.93 (SD 21.46)* \\
$\dLp[2]$ & 1441.93 (SD 780.04)* & 1827.38 (SD 1048.20)* & 3.33 (SD 1.99) & 10.87 (SD 7.95)* \\
$\dLp[3]$ & 1405.88 (SD 734.72)* & 1791.24 (SD 1036.65)* & 2.59 (SD 1.54) & 8.39 (SD 6.14) \\
$\dLp[5]$ & 1381.76 (SD 698.59)* & 1780.47 (SD 1051.63)* & 2.14 (SD 1.24) & 7.05 (SD 5.05) \\
$\dDotp[1]$ & 1614.32 (SD 865.77)* & 1912.60 (SD 1108.73)* & 12.61 (SD 7.76)* & 37.68 (SD 29.30) \\
$\dDotp[2]$ & 1441.72 (SD 764.50)* & 1785.08 (SD 1022.95)* & 3.59 (SD 2.07) & 11.23 (SD 8.27)* \\
\midrule
\textbf{Non-normal 100\%} & & & & \\
$\dLp[1]$ & 16826.88 (SD 9730.84)* & 21330.96 (SD 17349.78)* & 87.62 (SD 60.79)* & 302.63 (SD 295.31)* \\
$\dLp[2]$ & 14998.71 (SD 7987.20)* & 18824.80 (SD 14464.88)* & 34.13 (SD 23.49)* & 111.87 (SD 107.09)* \\
$\dLp[3]$ & 14596.36 (SD 7496.38)* & 18325.05 (SD 14124.61)* & 26.56 (SD 18.21)* & 85.96 (SD 83.26)* \\
$\dLp[5]$ & 14341.51 (SD 7145.45)* & 18100.46 (SD 14165.92)* & 21.97 (SD 14.66)* & 71.80 (SD 68.86)* \\
$\dDotp[1]$ & 16912.04 (SD 9614.67)* & 19734.62 (SD 15070.40)* & 131.14 (SD 93.41)* & 377.94 (SD 345.97)* \\
$\dDotp[2]$ & 14983.99 (SD 7891.72)* & 18226.76 (SD 13693.02)* & 37.03 (SD 25.07)* & 114.04 (SD 105.28)* \\
\midrule
\textbf{Non-normal 10\%} & & & & \\
$\dLp[1]$ & 1571.66 (SD 893.23)* & 2021.20 (SD 1211.60)* & 8.22 (SD 4.95) & 28.90 (SD 21.80)* \\
$\dLp[2]$ & 1418.67 (SD 771.66)* & 1829.14 (SD 1079.51)* & 3.23 (SD 1.94) & 10.83 (SD 8.01)* \\
$\dLp[3]$ & 1383.10 (SD 727.58)* & 1792.50 (SD 1067.12)* & 2.51 (SD 1.50) & 8.33 (SD 6.15) \\
$\dLp[5]$ & 1358.72 (SD 692.56)* & 1779.54 (SD 1079.72)* & 2.07 (SD 1.20) & 6.97 (SD 5.02) \\
$\dDotp[1]$ & 1584.02 (SD 855.69)* & 1905.71 (SD 1136.85)* & 12.25 (SD 7.66)* & 37.67 (SD 30.15)* \\
$\dDotp[2]$ & 1419.27 (SD 756.71)* & 1788.35 (SD 1057.95)* & 3.49 (SD 2.03) & 11.21 (SD 8.45)* \\
\bottomrule
\multicolumn{5}{l}{$^*$ Passed validity threshold of $\text{VRC}>1000$ and $\text{PBM}>10$.} \\
\end{tabular*}
\end{table}

\begin{table}[ht!]
\caption{Structural 4: Mean and standard deviation of SWC and DBI indices for reduced segment counts of 50 and 25 segments, calculated using valid distance functions.}
\label{tab:construct-test-4-swc-dbi}
\centering
\small
\setlength{\tabcolsep}{2pt}
\begin{tabular*}{\columnwidth}{@{\extracolsep{\fill}}l c c c c}
\toprule
 & \multicolumn{2}{c}{\textbf{SWC}} & \multicolumn{2}{c}{\textbf{DBI}} \\
\cmidrule(lr){2-3} \cmidrule(lr){4-5}
$d$ & 50 segments & 25 segments & 50 segments & 25 segments \\
\midrule
\textbf{Normal 100\%} & & & & \\
$\dLp[1]$ & 0.87 (SD 0.03) & 0.62 (SD 0.08) & 0.04 (SD 0.01)* & 0.03 (SD 0.01)* \\
$\dLp[2]$ & 0.88 (SD 0.03) & 0.62 (SD 0.08) & 0.04 (SD 0.01)* & 0.03 (SD 0.01)* \\
$\dLp[3]$ & 0.88 (SD 0.03) & 0.62 (SD 0.08) & 0.04 (SD 0.01)* & 0.03 (SD 0.01)* \\
$\dLp[5]$ & 0.88 (SD 0.03) & 0.62 (SD 0.08) & 0.04 (SD 0.01)* & 0.03 (SD 0.01)* \\
$\dDotp[1]$ & 0.88 (SD 0.03) & 0.62 (SD 0.08) & 0.04 (SD 0.01)* & 0.03 (SD 0.01)* \\
$\dDotp[2]$ & 0.88 (SD 0.03) & 0.62 (SD 0.08) & 0.04 (SD 0.01)* & 0.03 (SD 0.01)* \\
\midrule
\textbf{Normal 70\%} & & & & \\
$\dLp[1]$ & 0.87 (SD 0.03) & 0.62 (SD 0.08) & 0.05 (SD 0.01)* & 0.03 (SD 0.01)* \\
$\dLp[2]$ & 0.87 (SD 0.03) & 0.62 (SD 0.08) & 0.05 (SD 0.01)* & 0.03 (SD 0.01)* \\
$\dLp[3]$ & 0.87 (SD 0.03) & 0.62 (SD 0.08) & 0.05 (SD 0.01)* & 0.03 (SD 0.01)* \\
$\dLp[5]$ & 0.87 (SD 0.03) & 0.62 (SD 0.09) & 0.05 (SD 0.01)* & 0.03 (SD 0.01)* \\
$\dDotp[1]$ & 0.87 (SD 0.03) & 0.62 (SD 0.08) & 0.05 (SD 0.01)* & 0.03 (SD 0.01)* \\
$\dDotp[2]$ & 0.87 (SD 0.03) & 0.62 (SD 0.08) & 0.05 (SD 0.01)* & 0.03 (SD 0.01)* \\
\midrule
\textbf{Normal 10\%} & & & & \\
$\dLp[1]$ & 0.82 (SD 0.04) & 0.58 (SD 0.09) & 0.13 (SD 0.03)* & 0.09 (SD 0.03)* \\
$\dLp[2]$ & 0.82 (SD 0.04) & 0.59 (SD 0.09) & 0.12 (SD 0.03)* & 0.08 (SD 0.02)* \\
$\dLp[3]$ & 0.82 (SD 0.04) & 0.59 (SD 0.09) & 0.12 (SD 0.03)* & 0.08 (SD 0.02)* \\
$\dLp[5]$ & 0.83 (SD 0.04) & 0.59 (SD 0.09) & 0.12 (SD 0.03)* & 0.08 (SD 0.02)* \\
$\dDotp[1]$ & 0.82 (SD 0.04) & 0.59 (SD 0.09) & 0.13 (SD 0.03)* & 0.08 (SD 0.02)* \\
$\dDotp[2]$ & 0.82 (SD 0.04) & 0.59 (SD 0.09) & 0.12 (SD 0.03)* & 0.08 (SD 0.02)* \\
\midrule
\textbf{Non-normal 100\%} & & & & \\
$\dLp[1]$ & 0.87 (SD 0.03)* & 0.62 (SD 0.08) & 0.04 (SD 0.01)* & 0.03 (SD 0.01)* \\
$\dLp[2]$ & 0.88 (SD 0.03)* & 0.62 (SD 0.08) & 0.04 (SD 0.01)* & 0.03 (SD 0.01)* \\
$\dLp[3]$ & 0.88 (SD 0.03)* & 0.62 (SD 0.08) & 0.04 (SD 0.01)* & 0.03 (SD 0.01)* \\
$\dLp[5]$ & 0.88 (SD 0.03)* & 0.62 (SD 0.08) & 0.04 (SD 0.01)* & 0.03 (SD 0.01)* \\
$\dDotp[1]$ & 0.88 (SD 0.03)* & 0.62 (SD 0.08) & 0.04 (SD 0.01)* & 0.03 (SD 0.01)* \\
$\dDotp[2]$ & 0.88 (SD 0.03)* & 0.62 (SD 0.08) & 0.04 (SD 0.01)* & 0.03 (SD 0.01)* \\
\midrule
\textbf{Non-normal 10\%} & & & & \\
$\dLp[1]$ & 0.81 (SD 0.04)* & 0.58 (SD 0.09) & 0.13 (SD 0.03)* & 0.09 (SD 0.03)* \\
$\dLp[2]$ & 0.82 (SD 0.04)* & 0.59 (SD 0.09) & 0.12 (SD 0.03)* & 0.08 (SD 0.02)* \\
$\dLp[3]$ & 0.82 (SD 0.04)* & 0.59 (SD 0.09) & 0.12 (SD 0.03)* & 0.08 (SD 0.02)* \\
$\dLp[5]$ & 0.83 (SD 0.04)* & 0.59 (SD 0.09) & 0.12 (SD 0.03)* & 0.08 (SD 0.02)* \\
$\dDotp[1]$ & 0.82 (SD 0.04)* & 0.58 (SD 0.09) & 0.13 (SD 0.03)* & 0.08 (SD 0.02)* \\
$\dDotp[2]$ & 0.82 (SD 0.04)* & 0.59 (SD 0.09) & 0.12 (SD 0.03)* & 0.08 (SD 0.02)* \\
\bottomrule
\multicolumn{5}{l}{$^*$ Passed validity threshold of $\text{SWC}>0.9$ and $\text{DBI}<0.15$.} \\
\end{tabular*}
\end{table}

\begin{table}[ht!]
\caption{Structural 4: Mean and standard deviation of VRC and PBM indices for reduced segment counts of 50 and 25 segments, calculated using valid distance functions.}
\label{tab:construct-test-4-vrc-pmb}
\centering
\small
\setlength{\tabcolsep}{2pt}
\begin{tabular*}{\columnwidth}{@{\extracolsep{\fill}}l c c c c}
\toprule
 & \multicolumn{2}{c}{\textbf{VRC}} & \multicolumn{2}{c}{\textbf{PBM}} \\
\cmidrule(lr){2-3} \cmidrule(lr){4-5}
$d$ & 50 segments & 25 segments & 50 segments & 25 segments \\
\midrule
\textbf{Normal 100\%} & & & & \\
$\dLp[1]$ & 7489.43 (SD 3745.31)* & 8059.84 (SD 9371.46)* & 102.75 (SD 47.78)* & 1500.43 (SD 1907.91)* \\
$\dLp[2]$ & 6904.45 (SD 3209.78)* & 7711.91 (SD 9067.50)* & 43.48 (SD 18.84)* & 636.08 (SD 811.35)* \\
$\dLp[3]$ & 6704.85 (SD 3085.84)* & 7571.77 (SD 8825.44)* & 34.38 (SD 14.41)* & 506.85 (SD 647.27)* \\
$\dLp[5]$ & 6483.06 (SD 3030.89)* & 7375.19 (SD 8434.95)* & 28.04 (SD 11.45)* & 418.57 (SD 533.55)* \\
$\dDotp[1]$ & 7201.27 (SD 3329.28)* & 8138.45 (SD 8568.76)* & 168.73 (SD 73.12)* & 2332.59 (SD 2407.27)* \\
$\dDotp[2]$ & 6798.91 (SD 3079.22)* & 7735.79 (SD 8810.18)* & 50.25 (SD 21.33)* & 697.88 (SD 790.77)* \\
\midrule
\textbf{Normal 70\%} & & & & \\
$\dLp[1]$ & 4964.66 (SD 2469.34)* & 6314.07 (SD 9734.78)* & 71.29 (SD 36.27)* & 1135.82 (SD 1697.92)* \\
$\dLp[2]$ & 4615.65 (SD 2207.38)* & 5562.90 (SD 8184.52)* & 30.11 (SD 14.47)* & 451.20 (SD 655.43)* \\
$\dLp[3]$ & 4516.75 (SD 2131.89)* & 5390.85 (SD 7988.10)* & 23.89 (SD 11.22)* & 354.86 (SD 524.22)* \\
$\dLp[5]$ & 4396.91 (SD 2067.78)* & 5288.53 (SD 8006.14)* & 19.58 (SD 9.03)* & 293.81 (SD 449.70)* \\
$\dDotp[1]$ & 4933.25 (SD 2449.63)* & 6370.93 (SD 9839.40)* & 119.45 (SD 60.80)* & 1765.17 (SD 2322.46)* \\
$\dDotp[2]$ & 4595.05 (SD 2200.24)* & 5608.05 (SD 8321.14)* & 35.02 (SD 16.89)* & 496.79 (SD 655.19)* \\
\midrule
\textbf{Normal 10\%} & & & & \\
$\dLp[1]$ & 723.94 (SD 413.92) & 685.38 (SD 590.10) & 10.29 (SD 5.37)* & 125.77 (SD 86.50)* \\
$\dLp[2]$ & 666.69 (SD 373.60) & 628.56 (SD 525.74) & 4.29 (SD 2.18) & 51.08 (SD 33.93)* \\
$\dLp[3]$ & 648.75 (SD 363.57) & 620.74 (SD 516.48) & 3.39 (SD 1.71) & 40.57 (SD 26.76)* \\
$\dLp[5]$ & 628.35 (SD 355.23) & 617.92 (SD 514.13) & 2.77 (SD 1.41) & 33.75 (SD 22.27)* \\
$\dDotp[1]$ & 742.19 (SD 415.11) & 696.80 (SD 536.28) & 17.44 (SD 8.75)* & 205.61 (SD 129.82)* \\
$\dDotp[2]$ & 671.36 (SD 373.22) & 632.73 (SD 512.87) & 5.01 (SD 2.50) & 58.11 (SD 37.19)* \\
\midrule
\textbf{Non-normal 100\%} & & & & \\
$\dLp[1]$ & 7381.54 (SD 3781.20)* & 8049.21 (SD 9303.80)* & 104.33 (SD 49.91)* & 1552.09 (SD 1939.03)* \\
$\dLp[2]$ & 6776.33 (SD 3197.50)* & 7618.02 (SD 9058.43)* & 43.48 (SD 19.12)* & 643.70 (SD 831.24)* \\
$\dLp[3]$ & 6577.41 (SD 3076.44)* & 7465.48 (SD 8874.84)* & 34.21 (SD 14.57)* & 509.12 (SD 664.20)* \\
$\dLp[5]$ & 6358.46 (SD 3031.25)* & 7267.15 (SD 8531.39)* & 27.79 (SD 11.59)* & 417.37 (SD 545.90)* \\
$\dDotp[1]$ & 7090.93 (SD 3348.13)* & 8130.40 (SD 8534.19)* & 170.35 (SD 75.53)* & 2415.96 (SD 2487.10)* \\
$\dDotp[2]$ & 6676.59 (SD 3068.96)* & 7643.68 (SD 8789.66)* & 50.22 (SD 21.58)* & 707.50 (SD 811.49)* \\
\midrule
\textbf{Non-normal 10\%} & & & & \\
$\dLp[1]$ & 701.62 (SD 381.68) & 699.89 (SD 617.66) & 9.89 (SD 4.84) & 125.24 (SD 88.06)* \\
$\dLp[2]$ & 646.97 (SD 347.97) & 640.98 (SD 545.69) & 4.12 (SD 1.95) & 50.69 (SD 34.22)* \\
$\dLp[3]$ & 629.46 (SD 338.50) & 631.79 (SD 534.83) & 3.25 (SD 1.53) & 40.14 (SD 26.88)* \\
$\dLp[5]$ & 608.98 (SD 328.86) & 626.47 (SD 530.03) & 2.65 (SD 1.25) & 33.24 (SD 22.22)* \\
$\dDotp[1]$ & 714.94 (SD 371.95) & 710.35 (SD 565.10) & 16.72 (SD 7.91)* & 203.05 (SD 130.30)* \\
$\dDotp[2]$ & 650.77 (SD 345.09) & 645.70 (SD 534.37) & 4.81 (SD 2.25) & 57.53 (SD 37.24)* \\
\bottomrule
\multicolumn{5}{l}{$^*$ Passed validity threshold of $\text{VRC}>1000$ and $\text{PBM}>10$.} \\
\end{tabular*}
\end{table}

\FloatBarrier

\subsection{Discriminant Validity}
\begin{table}[ht!]
\caption{Mean and standard deviation for discriminant construct validity for the raw and downsampled, not sparsified data variants. The indices are calculated using the respective distance function.}
\label{tab:discriminant-mean-sd-100}
\centering
\small
\setlength{\tabcolsep}{2pt}
\begin{tabular*}{\columnwidth}{@{\extracolsep{\fill}}l c c c c}
\toprule
$d$ & \textbf{SWC} & \textbf{DBI} & \textbf{VRC} & \textbf{PBM} \\
\midrule
\textbf{Raw 100\%} & & & & \\
$\dLp[1]$ & -0.37 (SD 0.02)* & $4.75 \times 10^{11}$ (SD $1.48 \times 10^{12}$)* & 0.22 (SD 0.07)* & 0.0 (SD 0.0)* \\
$\dLp[2]$ & -0.36 (SD 0.02)* & $3.21 \times 10^{11}$ (SD $9.87 \times 10^{11}$)* & 0.23 (SD 0.07)* & 0.0 (SD 0.0)* \\
$\dLp[3]$ & -0.36 (SD 0.02)* & $2.90 \times 10^{11}$ (SD $8.89 \times 10^{11}$)* & 0.24 (SD 0.08)* & 0.0 (SD 0.0)* \\
$\dLp[5]$ & -0.37 (SD 0.02)* & $2.72 \times 10^{11}$ (SD $8.33 \times 10^{11}$)* & 0.24 (SD 0.08)* & 0.0 (SD 0.0)* \\
$\dDotp[1]$ & -0.37 (SD 0.02)* & $5.63 \times 10^{11}$ (SD $1.74 \times 10^{12}$)* & 0.23 (SD 0.07)* & 0.0 (SD 0.0)* \\
$\dDotp[2]$ & -0.36 (SD 0.02)* & $3.35 \times 10^{11}$ (SD $1.03 \times 10^{12}$)* & 0.23 (SD 0.07)* & 0.0 (SD 0.0)* \\
\midrule
\textbf{Downsampled 100\%} & & & & \\
$\dLp[1]$ & 0.61 (SD 0.05) & 0.56 (SD 0.08) & 108 (SD 27) & 0.09 (SD 0.03) \\
$\dLp[2]$ & 0.62 (SD 0.05) & 0.51 (SD 0.07) & 104 (SD 26) & 0.04 (SD 0.01) \\
$\dLp[3]$ & 0.62 (SD 0.05) & 0.50 (SD 0.07) & 102 (SD 26) & 0.03 (SD 0.01) \\
$\dLp[5]$ & 0.63 (SD 0.06) & 0.50 (SD 0.07) & 98 (SD 25) & 0.02 (SD 0.01) \\
$\dDotp[1]$ & 0.60 (SD 0.05) & 0.56 (SD 0.08) & 105 (SD 25) & 0.15 (SD 0.04) \\
$\dDotp[2]$ & 0.61 (SD 0.05) & 0.52 (SD 0.07) & 103 (SD 26) & 0.04 (SD 0.01) \\
\bottomrule
\multicolumn{5}{l}{$^*$ Passed validity threshold for raw data $\text{SWC}<0$, $\text{DBI}>100$, $\text{VRC}<1$, and $\text{PBM}<1$.}\\
\end{tabular*}
\end{table}

\FloatBarrier

\subsection{Optimal Distance Function ICVI - Family 2}

\begin{table}[htbp]
\caption{Preregistered confirmatory hypotheses (H) for family 2 establishing whether the top ranked distance functions lead to significantly more optimal ground truth values for SWC and DBI. $H=x>y$ means distance function $x$ leads to higher ICVI values than $y$. For SWC higher is better, for DBI lower is better. Seq is the sequence number in which the confirmatory tests need to be run, e indicates the effect size, and N is the number of pairs with a difference $>0.0001$.}
\label{tab:preregistrated-family-2}
\centering
\small
\setlength{\tabcolsep}{3pt}
\begin{tabular*}{\columnwidth}{@{\extracolsep{\fill}}l c c c c c c c c}
\toprule
& \multicolumn{4}{c}{\textbf{Normal}} & \multicolumn{4}{c}{\textbf{Non-normal}} \\
\cmidrule(lr){2-5} \cmidrule(lr){6-9}
& \textbf{Seq} & \textbf{H} & \textbf{e} & \textbf{N} & \textbf{Seq} & \textbf{H} & \textbf{e} & \textbf{N}\\
\midrule
\textbf{SWC:} \\[0.2em]
\textbf{Complete} (100\%) & 1. & $\dLp[5]$  > $\dLp[3]$  & 1.1 & 30 & 3. & $\dLp[5]$  > $\dLp[3]$  & 1.1 & 29 \\[0.3em]
\textbf{Partial} (70\%) & 2. & $\dLp[5]$  > $\dLp[3]$  & 1.1 & 30 & 4. & $\dLp[5]$  > $\dLp[3]$  & 1.1 & 29 \\[0.3em]
\textbf{Sparse} (10\%) & 5. & $\dLp[5]$  > $\dLp[3]$  & 1.1 & 30 & 6. & $\dLp[5]$  > $\dLp[3]$  & 1.1 & 30 \\[0.5em]
\midrule
\textbf{DBI:} \\[0.2em]
\textbf{Complete} (100\%) & 11. & $\dLp[5]$ < $\dLp[3]$  & 0.57 & 21 & 9. & $\dLp[5]$  < $\dLp[3]$  & 0.64 & 19 \\[0.3em]
\textbf{Partial} (70\%) & 10. & $\dLp[5]$ < $\dLp[3]$  & 0.61 & 23 & 12. & $\dLp[5]$  < $\dLp[3]$  & 0.49 & 20 \\[0.3em]
\textbf{Sparse} (10\%) & 7. & $\dLp[5]$  < $\dLp[3]$  & 0.81 & 28 & 8. & $\dLp[5]$  < $\dLp[3]$  & 0.8 & 29 \\
\bottomrule
\end{tabular*}
\end{table}

\begin{table}[!ht]
\caption{Confirmatory results for preregistered hypotheses for Family 2, see Table~\ref{tab:preregistrated-family-2}. All hypotheses (H) achieved statistical significance. $H=x>y$ means distance function $x$ leads to higher ICVI values than $y$. For SWC higher is better, for DBI lower is better. N is the number of pairs with a difference $>0.001$ out of the $30$ confirmatory subjects.}
\label{tab:family2_statistical_validation_results}
\small
\setlength{\tabcolsep}{3pt}
\begin{tabular*}{\columnwidth}{@{\extracolsep{\fill}}l r c r r r r}
\toprule
\multicolumn{2}{l}{\textbf{Data Variant}} & 
\multicolumn{1}{c}{\textbf{H}} &
\multicolumn{1}{c}{\textbf{p-value}} & 
\multicolumn{1}{c}{\textbf{Effect size}} & 
\multicolumn{1}{c}{\textbf{N}} &
\multicolumn{1}{c}{\textbf{Power \%}} \\[0.2em]
\midrule
\textbf{SWC:} \\[0.2em]
\multirow{3}{*}{\textbf{Normal}} & 100\% & $\dLp[5]$ > $\dLp[3]$ & <0.0001 & 1.1 & 29 & >99.9\\
& 70\% & $\dLp[5]$ > $\dLp[3]$ & <0.0001 & 1.1 & 30 & >99.9 \\
& 10\% & $\dLp[5]$ > $\dLp[3]$ & <0.0001 & 1.1 & 30 & >99.9\\
\midrule
\multirow{3}{*}{\textbf{Non-normal}} & 100\% & $\dLp[5]$ > $\dLp[3]$ & <0.0001 & 1.1 & 29 & >99.9 \\
& 70\% & $\dLp[5]$ > $\dLp[3]$ & <0.0001 & 1.1 & 30 & >99.9 \\
& 10\% & $\dLp[5]$ > $\dLp[3]$ & <0.0001 & 1.1 & 30 & >99.9 \\[0.2em]
\midrule
\textbf{DBI:} \\[0.2em]
\multirow{3}{*}{\textbf{Normal}} & 100\% & $\dLp[5]$ < $\dLp[3]$ & 0.0002 & 0.75 & 22 & 97\\
& 70\% & $\dLp[5]$ < $\dLp[3]$ & <0.0001 & 0.83 & 24 & 99.2 \\
& 10\% & $\dLp[5]$ < $\dLp[3]$ & <0.0001 & 0.8 & 27 & 99.4\\
\midrule
\multirow{3}{*}{\textbf{Non-normal}} & 100\% & $\dLp[5]$ < $\dLp[3]$ & 0.0002 & 0.78 & 20 & 96.7 \\
& 70\% & $\dLp[5]$ < $\dLp[3]$ & <0.0001 & 0.85 & 25 & 99.6 \\
& 10\% & $\dLp[5]$ < $\dLp[3]$ & <0.0001 & 0.73 & 30 & 99.1 \\
\bottomrule
\end{tabular*}
\end{table}

\FloatBarrier

\subsection{Optimal Distance Function for Correlation - Family 3}
\begin{table}[htbp]
\caption{Preregistered confirmatory hypotheses (H) for Family 3 establishing whether the top ranked distance functions lead to significantly stronger correlations between the ICVI and the Jaccard Index. $H=x>y$ means distance function $x$ leads to stronger correlation than $y$. Seq is the sequence number in which the confirmatory tests need to be run, e indicates the effect size, and N is the number of pairs with a difference $>0.001$. - indicates there was no significant differences between $\dLp[5]$ and $\dLp[3]$.}
\label{tab:preregistrated-family-3}
\centering
\small
\setlength{\tabcolsep}{3pt}
\begin{tabular*}{\columnwidth}{@{\extracolsep{\fill}}l c c c c c c c c}
\toprule
& \multicolumn{4}{c}{\textbf{Normal}} & \multicolumn{4}{c}{\textbf{Non-normal}}\\
\cmidrule(lr){2-5} \cmidrule(lr){6-9}
& \textbf{Seq} & \textbf{H} & \textbf{e} & \textbf{N} & \textbf{Seq} & \textbf{H} & \textbf{e} & \textbf{N}\\
\midrule
\textbf{SWC:} \\[0.2em]
\textbf{Complete} (100\%) & 1. & $\dLp[5]$ > $\dLp[3]$ & 1.09 & 27 & 2. & $\dLp[5]$ > $\dLp[3]$ & 1.09 & 27 \\[0.3em]
\textbf{Partial} (70\%) & 3. & $\dLp[5]$ > $\dLp[3]$ & 1.09 & 28 & 4. & $\dLp[5]$ > $\dLp[3]$ & 1.09 & 27 \\[0.3em]
\textbf{Sparse} (10\%) & 5. & $\dLp[5]$ > $\dLp[3]$ & 1.09 & 27 & 6. & $\dLp[5]$ > $\dLp[3]$ & 1.09 & 26 \\[0.5em]
\midrule
\textbf{DBI:} \\[0.2em]
\textbf{Complete} (100\%) & - & - & - & - & - & - & - & - \\[0.3em]
\textbf{Partial} (70\%) & 7. & $\dLp[5]$ > $\dLp[3]$ & 0.52 & 22 & - & - & - & - \\[0.3em]
\textbf{Sparse} (10\%) & - & - & - & - & - & - & - & - \\
\bottomrule
\end{tabular*}
\end{table}

\begin{table}[!ht]
\caption{Confirmatory results for preregistered hypotheses for family 3, see Table~\ref{tab:preregistrated-family-3}. All but the last three hypotheses achieved statistical significance. Testing stopped after the first failure (DBI, downsampled, 10\%). N is the number of pairs with a difference $>0.001$ out of the $30$ confirmatory subjects. - indicates there was no significant differences between $\dLp[5]$ and $\dLp[3]$.}
\label{tab:family3_statistical_validation_results}
\small
\setlength{\tabcolsep}{3pt}
\begin{tabular*}{\columnwidth}{@{\extracolsep{\fill}}l r c r r r r}
\toprule
\multicolumn{2}{l}{\textbf{Data Variant}} & 
\multicolumn{1}{c}{\textbf{H}} &
\multicolumn{1}{c}{\textbf{p-value}} & 
\multicolumn{1}{c}{\textbf{Effect size}} & 
\multicolumn{1}{c}{\textbf{N}} &
\multicolumn{1}{c}{\textbf{Power \%}} \\[0.2em]
\midrule
\textbf{SWC:} \\[0.2em]
\multirow{3}{*}{\textbf{Normal}} & 100\% & $\dLp[5]$ > $\dLp[3]$ & <0.0001 & 1.09 & 28 & >99.9\\
& 70\% & $\dLp[5]$ > $\dLp[3]$ & <0.0001 & 1.1 & 29 & >99.9\\
& 10\% & $\dLp[5]$ > $\dLp[3]$ & <0.0001 & 1.09 & 26 & >99.9\\
\midrule
\multirow{3}{*}{\textbf{Non-normal}} & 100\% & $\dLp[5]$ > $\dLp[3]$ & <0.0001 & 1.09 & 28 & >99.9 \\
& 70\% & $\dLp[5]$ > $\dLp[3]$ & <0.0001 & 1.1 & 29 & >99.9 \\
& 10\% & $\dLp[5]$ > $\dLp[3]$ & <0.0001 & 1.08 & 25 & >99.9 \\[0.2em]
\midrule
\textbf{DBI:} \\[0.2em]
\multirow{3}{*}{\textbf{Normal}} & 100\% & - & - & - & - & -\\
& 70\% & $\dLp[5]$ > $\dLp[3]$ & 0.004 & 0.62 & 18 & 83.4 \\
& 10\% & - & - & - & - & -\\
\midrule
\multirow{3}{*}{\textbf{Non-normal}} & 100\% & - & - & - & - & - \\
& 70\% & - & - & - & - & -\\
& 10\% & - & - & - & - & -\\
\bottomrule
\end{tabular*}
\end{table}

\FloatBarrier

\subsection{Optimal ICVI - Family 4}

\begin{table}[htbp]
\caption{Preregistered confirmatory hypotheses (H) for family 4 comparing whether SWC or DBI is stronger linearly correlated to the Jaccard index using the L5 distance function for all data variants. $H=x>y$ means ICVIs $x$ is stronger correlated with Jaccard than ICVIs $y$. Seq is the sequence number in which the confirmatory tests need to be run, e indicates the effect size, and N is the number of pairs with a difference $>0.001$. - indicates there was no significant differences between SWC and DBI.}
\label{tab:preregistrated-family-4}
\centering
\small
\setlength{\tabcolsep}{3pt}
\begin{tabular*}{\columnwidth}{@{\extracolsep{\fill}}l c c c c c c c c}
\toprule
& \multicolumn{4}{c}{\textbf{Normal}} & \multicolumn{4}{c}{\textbf{Non-normal}} \\
\cmidrule(lr){2-5} \cmidrule(lr){6-9}
& \textbf{Seq} & \textbf{H} & \textbf{e} & \textbf{N} & \textbf{Seq} & \textbf{H} & \textbf{e} & \textbf{N}\\
\midrule
\textbf{Complete} (100\%) & - & - & - & - & - & - & - & - \\[0.3em]
\textbf{Partial} (70\%) & - & - & - & - & - & - & - & - \\[0.3em]
\textbf{Sparse} (10\%) & 2. & DBI > SWC & 0.5 & 30 & 1. & DBI > SWC & 0.6 & 30\\
\bottomrule
\end{tabular*}
\end{table}

\begin{table}[!ht]
\caption{Confirmatory results for preregistered hypotheses for family 4, see Table~\ref{tab:preregistrated-family-4}. All hypotheses achieved statistical significance. N is the number of pairs with a difference $>0.001$ out of the $30$ confirmatory subjects.}
\label{tab:family4_statistical_validation_results}
\small
\setlength{\tabcolsep}{3pt}
\begin{tabular*}{\columnwidth}{@{\extracolsep{\fill}}l r c r r r r}
\toprule
\multicolumn{2}{l}{\textbf{Data Variant}} & 
\multicolumn{1}{c}{\textbf{H}} &
\multicolumn{1}{c}{\textbf{p-value}} & 
\multicolumn{1}{c}{\textbf{Effect size}} & 
\multicolumn{1}{c}{\textbf{N}} &
\multicolumn{1}{c}{\textbf{Power \%}} \\[0.2em]
\midrule
\multirow{3}{*}{\textbf{Normal}} & 100\% & - & - & - & - & - \\
& 70\% & - & - & - & - & - \\
& 10\% & DBI > SWC & 0.0002 & 0.64 & 30 & 97 \\
\midrule
\multirow{3}{*}{\textbf{Non-normal}} & 100\% & - & - & - & - & - \\
& 70\% & - & - & - & - & - \\
& 10\% & DBI > SWC & 0.0002 & 0.64 & 30 & 97 \\
\bottomrule
\end{tabular*}
\end{table}

\FloatBarrier
\subsection{Cluster Bias - Family 5}

\begin{table}[htbp]
\caption{Preregistered confirmatory hypotheses (H) for family 5 comparing the impact of cluster count reduction from 23 to 11, 11 to 6, and 23 to 6 on raw values for SWC and DBI using the L5 distance function for all data variants. $H=x>y$ means ICVI $x$ values are higher than ICVI $y$ values. Higher values are better for SWC, while lower values are better for DBI. Seq is the sequence number in which the confirmatory tests need to be run, e indicates the effect size, and N is the number of pairs with a difference $>0.0001$. - indicates there was no significant difference.}
\label{tab:preregistrated-family-5}
\centering
\small
\setlength{\tabcolsep}{3pt}
\begin{tabular*}{\columnwidth}{@{\extracolsep{\fill}}l c c c c c c c c}
\toprule
& \multicolumn{4}{c}{\textbf{Normal}} & \multicolumn{4}{c}{\textbf{Non-normal}} \\
\cmidrule(lr){2-5} \cmidrule(lr){6-9}
& \textbf{Seq} & \textbf{H} & \textbf{e} & \textbf{N} & \textbf{Seq} & \textbf{H} & \textbf{e} & \textbf{N}\\
\midrule
\textbf{SWC:} \\[0.2em]
\multirow{3}{*}{\textbf{Complete} (100\%)} & 17. & $\text{SWC}_{6} > \text{SWC}_{23}$ & 0.48 & 26 & 14. & $\text{SWC}_{6} > \text{SWC}_{23}$ & 0.51 & 28 \\
& 23. & $\text{SWC}_{11} > \text{SWC}_{23}$ & 0.42 & 26 & 24. & $\text{SWC}_{11} > \text{SWC}_{23}$ & 0.41 & 26 \\
& - & $\text{SWC}_{6}\, \text{vs} \, \text{SWC}_{11}$ & - & - & - & $\text{SWC}_{6}\, \text{vs} \, \text{SWC}_{11}$ & - & - \\[0.3em]
\multirow{3}{*}{\textbf{Partial} (70\%)} & 16. & $\text{SWC}_{6} > \text{SWC}_{23}$ & 0.5 & 26 & 12. & $\text{SWC}_{6} > \text{SWC}_{23}$ & 0.56 & 26 \\
& - & $\text{SWC}_{11}\, \text{vs} \, \text{SWC}_{23}$ & - & - & - & $\text{SWC}_{11}\, \text{vs} \, \text{SWC}_{23}$ & - & - \\
& - & $\text{SWC}_{6}\, \text{vs} \, \text{SWC}_{11}$ & - & - & - & $\text{SWC}_{6}\, \text{vs} \, \text{SWC}_{11}$ & - & - \\[0.3em]
\multirow{3}{*}{\textbf{Sparse} (10\%)} & 9. & $\text{SWC}_{6} > \text{SWC}_{23}$ & 0.73 & 30 & 7. & $\text{SWC}_{6} > \text{SWC}_{23}$ & 0.8 & 30 \\
& 18. & $\text{SWC}_{11} > \text{SWC}_{23}$ & 0.64 & 29 & 11. & $\text{SWC}_{11} > \text{SWC}_{23}$ & 0.7 & 29 \\
& - & $\text{SWC}_{6}\, \text{vs} \, \text{SWC}_{11}$ & - & - & - & $\text{SWC}_{6}\, \text{vs} \, \text{SWC}_{11}$ & - & - \\[0.5em]
\midrule
\textbf{DBI:} \\[0.2em]
\multirow{3}{*}{\textbf{Complete} (100\%)} & 5. & $\text{DBI}_{23} > \text{DBI}_{6}$ & 1.04 & 28 & 3. & $\text{DBI}_{23} > \text{DBI}_{6}$ & 1.09 & 28 \\
& 6. & $\text{DBI}_{23} > \text{DBI}_{11}$ & 0.98 & 30 & 4. & $\text{DBI}_{23} > \text{DBI}_{11}$ & 0.97 & 30 \\
& 25. & $\text{DBI}_{11} > \text{DBI}_{6}$ & 0.37 & 30 & 26. & $\text{DBI}_{11} > \text{DBI}_{6}$ & 0.37 & 30 \\[0.3em]
\multirow{3}{*}{\textbf{Partial} (70\%)} & 8. & $\text{DBI}_{23} > \text{DBI}_{6}$ & 1.02 & 29 & 10. & $\text{DBI}_{23} > \text{DBI}_{6}$ & 1.02 & 30 \\
& 13. & $\text{DBI}_{23} > \text{DBI}_{11}$ & 0.95 & 28 & 15. & $\text{DBI}_{23} > \text{DBI}_{11}$ & 0.93 & 28 \\
& 21. & $\text{DBI}_{11} > \text{DBI}_{6}$ & 0.46 & 29 & 22. & $\text{DBI}_{11} > \text{DBI}_{6}$ & 0.46 & 29 \\[0.3em]
\multirow{3}{*}{\textbf{Sparse} (10\%)} & 1. & $\text{DBI}_{23} > \text{DBI}_{6}$ & 1.1 & 30 & 2. & $\text{DBI}_{23} > \text{DBI}_{6}$ & 1.1 & 29 \\
& 19. & $\text{DBI}_{23} > \text{DBI}_{11}$ & 1.01 & 30 & 20. & $\text{DBI}_{23} > \text{DBI}_{11}$ & 1.04 & 30 \\
& 25. & $\text{DBI}_{11} > \text{DBI}_{6}$ & 0.47 & 29 & 26. & $\text{DBI}_{11} > \text{DBI}_{6}$ & 0.47 & 29 \\
\bottomrule
\end{tabular*}
\end{table}

\begin{table}[!ht]
\caption{Confirmatory results for preregistered hypotheses for family 5, see Table~\ref{tab:preregistrated-family-5}. Testing stopped after the first failure (DBI$_{11}$ > DBI$_6$, partial, 70\%). N is the number of pairs with a difference $>0.001$ out of the $30$ confirmatory subjects.}
\label{tab:family5_statistical_validation_results}
\small
\setlength{\tabcolsep}{3pt}
\begin{tabular*}{\columnwidth}{@{\extracolsep{\fill}}l r c r r r r}
\toprule
\multicolumn{2}{l}{\textbf{Data Variant}} & 
\multicolumn{1}{c}{\textbf{H}} &
\multicolumn{1}{c}{\textbf{p-value}} & 
\multicolumn{1}{c}{\textbf{Effect size}} & 
\multicolumn{1}{c}{\textbf{N}} &
\multicolumn{1}{c}{\textbf{Power \%}} \\[0.2em]
\midrule
\textbf{SWC:} \\[0.2em]
\multirow{5}{*}{\textbf{Normal}} & 100\% & SWC$_6$ > SWC$_{23}$ & 0.017 & 0.39 & 29 & 68.1\\
& & \colorbox{black!10}{SWC$_{11}$ > SWC$_{23}$$^{**}$} & - & - & - & -\\[0.3em]
& 70\% & SWC$_6$ > SWC$_{23}$ & 0.033 & 0.34 & 30 & 57.5\\[0.3em]
& 10\% & SWC$_6$ > SWC$_{23}$ & 0.001 & 0.56 & 30 & 92.2\\
& & SWC$_{11}$ > SWC$_{23}$ & 0.0008 & 0.61 & 27 & 93.6\\
\midrule
\multirow{5}{*}{\textbf{Non-normal}} & 100\% & SWC$_6$ > SWC$_{23}$ & 0.01 & 0.44 & 28 & 74.9 \\
& & \colorbox{black!10}{SWC$_{11}$ > SWC$_{23}$$^{**}$} & - & - & - & - \\[0.3em]
& 70\% & SWC$_6$ > SWC$_{23}$ & 0.016 & 0.39 & 30 & 68.8 \\[0.3em]
& 10\% & SWC$_6$ > SWC$_{23}$ & 0.0006 & 0.59 & 30 & 94.4 \\
& & SWC$_{11}$ > SWC$_{23}$ & 0.0003 & 0.66 & 28 & 96.6 \\[0.2em]
\midrule
\textbf{DBI:} \\[0.2em]
\multirow{8}{*}{\textbf{Normal}} & 100\% & DBI$_{23}$ > DBI$_6$ & <0.0001 & 0.96 & 29 & >99.9\\
& & DBI$_{23}$ > DBI$_{11}$ & <0.0001 & 0.89 & 28 & >99.9\\
& & \colorbox{black!10}{DBI$_{11}$ > DBI$_6$$^{**}$} & - & - & - & -\\[0.3em]
& 70\% & DBI$_{23}$ > DBI$_6$ & <0.0001 & 1.06 & 29 & >99.9\\
& & DBI$_{23}$ > DBI$_{11}$ & <0.0001 & 0.78 & 29 & >99.9\\
& & \colorbox{black!10}{DBI$_{11}$ > DBI$_6$$^*$} & \textit{0.081} & \textit{0.26} & \textit{29} & \textit{40.3}\\[0.3em]
& 10\% & DBI$_{23}$ > DBI$_6$ & <0.0001 & 1.05 & 30 & >99.9\\
& & DBI$_{23}$ > DBI$_{11}$ & <0.0001 & 1.02 & 30 & >99.9\\
\midrule
\multirow{8}{*}{\textbf{Non-normal}} & 100\% & DBI$_{23}$ > DBI$_6$ & <0.0001 & 0.97 & 29 & >99.9\\
& & DBI$_{23}$ > DBI$_{11}$ & <0.0001 & 0.92 & 28 & >99.9\\
& & \colorbox{black!10}{DBI$_{11}$ > DBI$_6$$^{**}$} & - & - & - & -\\[0.3em]
& 70\% & DBI$_{23}$ > DBI$_6$ & <0.0001 & 1.06 & 29 & >99.9\\
& & DBI$_{23}$ > DBI$_{11}$ & <0.0001 & 0.8 & 29 & >99.9\\
& & \colorbox{black!10}{DBI$_{11}$ > DBI$_6$$^{**}$} & - & - & - & -\\[0.3em]
& 10\% & DBI$_{23}$ > DBI$_6$ & <0.0001 & 1.07 & 29 & >99.9\\
& & DBI$_{23}$ > DBI$_{11}$ & <0.0001 & 1.04 & 30 & >99.9\\
\bottomrule
\multicolumn{7}{l}{$^*$ Not statistically significant} \\
\multicolumn{7}{l}{$^{**}$ Not tested} \\
\end{tabular*}
\end{table}

\FloatBarrier 
\subsection{Segmentation Bias - Family 6}

\begin{table}[htbp]
\caption{Preregistered confirmatory hypotheses (H) for family 6 comparing the impact of segment count reduction from 100 to 50 and 50 to 25 on raw values for SWC and DBI using the L5 distance function for all data variants. $H=x>y$ means ICVI $x$ values are higher than ICVI $y$ values. Higher values are better for SWC, while lower values are better for DBI. Seq is the sequence number in which the confirmatory tests need to be run, e indicates the effect size, and N is the number of pairs with a difference $>0.0001$.}
\label{tab:preregistrated-family-6}
\centering
\small
\setlength{\tabcolsep}{3pt}
\begin{tabular*}{\columnwidth}{@{\extracolsep{\fill}}l c c c c c c c c}
\toprule
& \multicolumn{4}{c}{\textbf{Normal}} & \multicolumn{4}{c}{\textbf{Non-normal}} \\
\cmidrule(lr){2-5} \cmidrule(lr){6-9}
& \textbf{Seq} & \textbf{H} & \textbf{e} & \textbf{N} & \textbf{Seq} & \textbf{H} & \textbf{e} & \textbf{N}\\
\midrule
\textbf{SWC:} \\[0.2em]
\multirow{2}{*}{\textbf{Complete} (100\%)} & 1. & $\text{SWC}_{100} > \text{SWC}_{50}$ & 1.10 & 30 & 2. & $\text{SWC}_{100} > \text{SWC}_{50}$ & 1.10 & 30 \\
& 3. & $\text{SWC}_{50} > \text{SWC}_{25}$ & 1.10 & 30 & 4. & $\text{SWC}_{50} > \text{SWC}_{25}$ & 1.10 & 30 \\[0.3em]
\multirow{2}{*}{\textbf{Partial} (70\%)} & 5. & $\text{SWC}_{100} > \text{SWC}_{50}$ & 1.10 & 30 & 6. & $\text{SWC}_{100} > \text{SWC}_{50}$ & 1.10 & 30 \\
& 7. & $\text{SWC}_{50} > \text{SWC}_{25}$ & 1.10 & 30 & 8. & $\text{SWC}_{50} > \text{SWC}_{25}$ & 1.10 & 30 \\[0.3em]
\multirow{2}{*}{\textbf{Sparse} (10\%)} & 9. & $\text{SWC}_{100} > \text{SWC}_{50}$ & 1.10 & 30 & 10. & $\text{SWC}_{100} > \text{SWC}_{50}$ & 1.10 & 30 \\
& 11. & $\text{SWC}_{50} > \text{SWC}_{25}$ & 1.10 & 30 & 12. & $\text{SWC}_{50} > \text{SWC}_{25}$ & 1.10 & 30 \\[0.5em]
\midrule
\textbf{DBI:} \\[0.2em]
\multirow{2}{*}{\textbf{Complete} (100\%)} & 20. & $\text{DBI}_{100} > \text{DBI}_{50}$ & 0.62 & 30 & 19. & $\text{DBI}_{100} > \text{DBI}_{50}$ & 0.66 & 30 \\
& 17. & $\text{DBI}_{50} > \text{DBI}_{25}$ & 0.84 & 30 & 18. & $\text{DBI}_{50} > \text{DBI}_{25}$ & 0.84 & 30 \\[0.3em]
\multirow{2}{*}{\textbf{Partial} (70\%)} & 24. & $\text{DBI}_{100} > \text{DBI}_{50}$ & 0.52 & 29 & 22. & $\text{DBI}_{100} > \text{DBI}_{50}$ & 0.54 & 29 \\
& 15. & $\text{DBI}_{50} > \text{DBI}_{25}$ & 0.94 & 30 & 16. & $\text{DBI}_{50} > \text{DBI}_{25}$ & 0.94 & 30 \\[0.3em]
\multirow{2}{*}{\textbf{Sparse} (10\%)} & 23. & $\text{DBI}_{100} > \text{DBI}_{50}$ & 0.54 & 30 & 21. & $\text{DBI}_{100} > \text{DBI}_{50}$ & 0.59 & 30 \\
& 13. & $\text{DBI}_{50} > \text{DBI}_{25}$ & 1.02 & 29 & 14. & $\text{DBI}_{50} > \text{DBI}_{25}$ & 1.02 & 29 \\
\bottomrule
\end{tabular*}
\end{table}

\begin{table}[!ht]
\caption{Confirmatory results for preregistered hypotheses for family 6, see Table~\ref{tab:preregistrated-family-6}. Testing stopped after the first failure (DBI$_{100}$ > DBI$_{50}$, sparse, non-normal). N is the number of pairs with a difference $>0.001$ out of the $30$ confirmatory subjects.}
\label{tab:family6_statistical_validation_results}
\small
\setlength{\tabcolsep}{3pt}
\begin{tabular*}{\columnwidth}{@{\extracolsep{\fill}}l r c r r r r}
\toprule
\multicolumn{2}{l}{\textbf{Data Variant}} & 
\multicolumn{1}{c}{\textbf{H}} &
\multicolumn{1}{c}{\textbf{p-value}} & 
\multicolumn{1}{c}{\textbf{Effect size}} & 
\multicolumn{1}{c}{\textbf{N}} &
\multicolumn{1}{c}{\textbf{Power \%}} \\[0.2em]
\midrule
\textbf{SWC:} \\[0.2em]
\multirow{6}{*}{\textbf{Normal}} & 100\% & SWC$_{100}$ > SWC$_{50}$ & <0.0001 & 1.1 & 30 & >99.9\\
& & SWC$_{50}$ > SWC$_{25}$ & <0.0001 & 1.1 & 30 & >99.9\\[0.3em]
& 70\% & SWC$_{100}$ > SWC$_{50}$ & <0.0001 & 1.1 & 30 & >99.9\\
& & SWC$_{50}$ > SWC$_{25}$ & <0.0001 & 1.1 & 30 & >99.9\\[0.3em]
& 10\% & SWC$_{100}$ > SWC$_{50}$ & <0.0001 & 1.1 & 30 & >99.9\\
& & SWC$_{50}$ > SWC$_{25}$ & <0.0001 & 1.1 & 30 & >99.9\\
\midrule
\multirow{6}{*}{\textbf{Non-normal}} & 100\% & SWC$_{100}$ > SWC$_{50}$ & <0.0001 & 1.1 & 30 & >99.9 \\
& & SWC$_{50}$ > SWC$_{25}$ & <0.0001 & 1.1 & 30 & >99.9 \\[0.3em]
& 70\% & SWC$_{100}$ > SWC$_{50}$ & <0.0001 & 1.1 & 30 & >99.9 \\
& & SWC$_{50}$ > SWC$_{25}$ & <0.0001 & 1.1 & 30 & >99.9 \\[0.3em]
& 10\% & SWC$_{100}$ > SWC$_{50}$ & <0.0001 & 1.1 & 30 & >99.9 \\
& & SWC$_{50}$ > SWC$_{25}$ & <0.0001 & 1.1 & 30 & >99.9 \\
\midrule
\textbf{DBI:} \\[0.2em]
\multirow{6}{*}{\textbf{Normal}} & 100\% & DBI$_{100}$ > DBI$_{50}$ & 0.024 & 0.36 & 30 & 63.2\\[0.3em]
&  & DBI$_{50}$ > DBI$_{25}$ & <0.0001 & 0.90 & 29 & >99.9\\
& 70\%  & \colorbox{black!10}{DBI$_{100}$ > DBI$_{50}$$^{**}$} & - & - & - & -\\[0.3em]
& & DBI$_{50}$ > DBI$_{25}$ & <0.0001 & 0.93 & 29 & >99.9\\
& 10\% & \colorbox{black!10}{DBI$_{100}$ > DBI$_{50}$$^{**}$} & - & - & - & -\\
& & DBI$_{50}$ > DBI$_{25}$ & <0.0001 & 0.84 & 30 & >99.9\\
\midrule
\multirow{6}{*}{\textbf{Non-normal}} & 100\% & DBI$_{100}$ > DBI$_{50}$ & 0.015 & 0.40 & 29 & 69.7\\
&  & DBI$_{50}$ > DBI$_{25}$ & <0.0001 & 0.90 & 29 & >99.9\\[0.3em]
& 70\% & \colorbox{black!10}{DBI$_{100}$ > DBI$_{50}$$^{**}$} & - & - & - & -\\
&  & DBI$_{50}$ > DBI$_{25}$ & <0.0001 & 0.93 & 29 & >99.9\\[0.3em]
& 10\% & \colorbox{black!10}{DBI$_{100}$ > DBI$_{50}$$^*$} & \textit{0.149} & \textit{0.19} & \textit{30} & \textit{27.3}\\
&  & DBI$_{50}$ > DBI$_{25}$ & <0.0001 & 0.84 & 30 & >99.9\\
\bottomrule
\multicolumn{7}{l}{$^*$ Not statistically significant} \\
\multicolumn{7}{l}{$^{**}$ Not tested} \\
\end{tabular*}
\end{table}
\end{document}